\documentclass[twoside]{article}

\usepackage[accepted]{aistats2024arxiv}
\usepackage[round]{natbib}

\usepackage{ifthen}
\newboolean{arxiv-version}
\setboolean{arxiv-version}{false}
\usepackage[suppress]{color-edits}  
\addauthor{as}{orange}     
\addauthor{ak}{red}    
\addauthor{lw}{blue}   

\usepackage{hyperref}       
\usepackage{url}            
\usepackage{booktabs}       
\usepackage{amsfonts}       
\usepackage{nicefrac}       
\usepackage{microtype}      
\usepackage{xcolor}         

\usepackage{xspace}

\usepackage{algorithm}
\usepackage{algorithmic}
\usepackage{mathtools}
\usepackage{amsmath}
\usepackage{amsthm}
\usepackage{amssymb}
\numberwithin{equation}{section}
\usepackage{subcaption}
\usepackage{caption}
\usepackage{wrapfig}
\usepackage{Definitions}

\newcommand{\supdensity}{\delta_{\textnormal{sup}}}
\newcommand{\infdensity}{\delta_{\textnormal{inf}}}
\newcommand{\given}{\,|\,}

\newcommand{\dif}{\mathop{}\!\mathrm{d}}
\newcommand{\EH}{H_e} 

\newcommand{\xhdr}[1]{\vspace{1mm} \noindent{\bf #1}}
\newcommand{\R} {\ensuremath{\mathbb{R}}} 
\newcommand{\N} {\ensuremath{\mathbb{N}}} 
\newcommand{\Var}{\ensuremath{\text{Var}}}
\newcommand{\PL}{\textnormal{PL}}
\newcommand{\IPW}{\textnormal{IPW}}
\renewcommand{\refeq}[1]{Eq.~(\ref{#1})}

\newcommand{\RelImp}{\texttt{RelImp}\xspace}

\begin{document}

%
\runningtitle{Oracle-Efficient Pessimism}

%
\runningauthor{Lequn Wang, Akshay Krishnamurthy, Aleksandrs Slivkins}

\renewcommand*{\thefootnote}{\fnsymbol{footnote}}
\setcounter{footnote}{0}

\twocolumn[

\aistatstitle{Oracle-Efficient Pessimism:\\Offline Policy Optimization In Contextual Bandits\footnotemark}

\aistatsauthor{ Lequn Wang \And Akshay Krishnamurthy \And  Aleksandrs Slivkins}

\aistatsaddress{ Netflix \\ lequnw@netflix.com  \And  Microsoft Research NYC \\ akshaykr@microsoft.com \And Microsoft Research NYC \\ slivkins@microsoft.com}
]

\begin{abstract}
We consider offline policy optimization (OPO) in contextual bandits, where one is given a fixed dataset of logged interactions. While pessimistic regularizers are typically used to mitigate distribution shift, prior implementations thereof are \asedit{either specialized or computationally inefficient}.
We present the first \emph{\asedit{general} oracle-efficient} algorithm for pessimistic \asedit{OPO}:
it reduces to supervised learning, leading to broad applicability. We obtain
statistical guarantees analogous to those for prior pessimistic approaches.
We instantiate our approach for both discrete and continuous actions and perform experiments in both settings, showing advantage over
unregularized \asedit{OPO} across a wide range of
configurations.

\end{abstract}

\ifthenelse{\boolean{arxiv-version}}{
\section{Introduction}
}
{
\section{INTRODUCTION}
\footnotetext{
Initial version: June 2023. This version (Oct'23): revised presentation, expanded discussion of related work.\newline
\indent Most work done while LW was a graduate student at Cornell and an intern at Microsoft Research NYC.}
\renewcommand{\thefootnote}{\arabic{footnote}}
\setcounter{footnote}{0}
}
\label{sec:intro}
\asedit{Offline Policy Optimization (OPO)} is a fundamental variant of reinforcement learning (RL) where one  optimizes a decision-making policy using previously collected data.
(It is also called \emph{Offline RL}.) OPO is particularly useful
\asdelete{in high-stakes applications}
when new experimentation via online RL  is costly, dangerous, \asedit{or would take too long}. A central challenge in \asedit{OPO} is \emph{distribution shift},
\asedit{when the logging policy and the learned policy induce different data distributions,
possibly leading to high uncertainty on the learned policy and poor overall performance.}
This challenge is typically mitigated via \emph{pessimism}: optimizing a regularized objective that \asedit{evaluates each policy via a ``pessimistic" confidence bound on its loss, thereby penalizing policies with high empirical variance.} While this approach is well-understood from statistical perspective, computationally efficient implementations \asedit{remain elusive}. In this paper, we address this issue for \emph{contextual bandits}, a practically important special case of RL and a research area in its own right.



\asedit{Thus, we study  pessimistic OPO in contextual bandits.%
\footnote{\asedit{\Ie OPO with pessimism, as described above. Without further mention, we focus on regularizers that penalize empirical variance, rather than policy complexity. While the latter regularizers may also be construed as ``pessimistic", they target overfitting rather than distribution shift.}}
We develop a new algorithm for this problem.} Our algorithm is \emph{oracle-efficient}, making a single call to an (arbitrary) computational oracle for supervised learning. The algorithm efficiently forms an artificial problem instance of  supervised learning which incorporates pessimism and is passed to the oracle. This reduction to supervised learning allows us to handle any ``oracle-supported" policy class (\ie a policy class that an oracle can optimize over), and offers flexibility to employ various oracle implementations developed in prior work.
We obtain similar statistical guarantees
as prior (computationally inefficient) implementations of pessimism.
\asedit{On a high level, we obtain the first computationally efficient algorithm with statistical guarantees for an arbitrary oracle-supported policy class.}%

Our approach carries over to contextual bandits with continuous actions, an important, well-studied scenario motivated by optimizing prices and continuous system  parameters. Distribution shift and computational tractability are particularly challenging in this scenario. This is due to the complexity of the action space and the large number of hyper-parameters, respectively.

We conduct an extensive empirical study for both discrete and continuous actions. We instantiate our approach across a range of configurations, both for the experimental environment and for the algorithm itself. We find that our approach is broadly superior to the vanilla policy optimization,
while it is much more widely applicable---due to oracle efficiency---than the implementations of pessimism using prior techniques.

All proofs and experiment details are in the appendix.

\subsection{Related work}
\label{sec:related}
OPO is extensively studied in contextual bandits
\citep[starting from, \eg][]{beygelzimer2009offset,bottou2013counterfactual,dudik2014doubly,athey2016recursive},
and in RL more generally~\citep{levine2020offline}.


OPO methods typically build on estimators for \emph{offline policy evaluation}~\citep{langford2008exploration,dudik2014doubly,farajtabar2018more}, with inverse probability weighting (IPW) estimator~\citep{horvitz1952generalization} as the canonical example. It is well known that IPW can have large variance, and a number of variations of this estimator, such as clipping~\citep{strehl2010learning,wang2017optimal,su2019cab}, self-normalization~\citep{swaminathan2015self}, and shrinkage~\citep{su2020doubly}, have been developed to mitigate variance in exchange for introducing bias. These estimators do not alleviate distribution shift
and are complementary to our approach. As such, we focus on vanilla IPW for our theoretical treatment.

The oracle-efficiency framework has been prominent in contextual bandits since \citet{Langford-nips07}, with much of this work focusing on supervised learning oracles~\citep[e.g.,][]{Langford-nips07,dudik2011efficient,agarwal2014taming}.
It has been adopted across a range of other problems including structured prediction~\citep{daume2009search,ross2011reduction}, active learning~\citep{dasgupta2007general},
and online learning~\citep{haghtalab2022oracle,block2022smoothed} and in many cases has lead to highly practical algorithms.

Contextual bandits with continuous actions have been studied since \citet{Pal-Bandits-aistats10,contextualMAB-colt11}, amidst many related papers on \emph{non-}contextual bandits, usually under Lipschitz assumptions~\citep[c.f.,][Ch 4.4]{slivkins-MABbook}. Our work builds on the ``smoothing'' approach from \citet{krishnamurthy2020contextual}, see also
\citep{majzoubi2020efficient,zhu2022contextual,zhu2021making}.
OPO with continuous actions has also been studied in \citet{kallus2018policy,chernozhukov2019semi},
but focusing on issues other than pessimism.

\ascomment{new}\xhdr{Pessimistic OPO in contextual bandits}
was introduced in \citep{swaminathan2015batch} via the Empirical Bernstein (EB) regularizer. This approach allows for arbitrary policy classes, but is computationally inefficient, limiting applicability.

The vast follow-up work is either
computationally inefficient
\citep[\eg][]{jin2022policy},
or lacks statistical guarantees
\citep[\eg][]{fujimoto2019off,kumar2020conservative,yu2020mopo,trabucco2021conservative},
or is substantially restricted in scope. The latter work
posits \emph{realizability} -- a particular loss model, \eg linear
\citep[\eg][]{liu2020provably,li2022pessimism,rashidinejad2022optimal,uehara2021pessimistic},
or focuses on a ``tabular" problem%
\footnote{\Ie optimizes over the class of all policies, implicitly assuming a small number of contexts.}
\citep[\eg][]{kidambi2020morel,rashidinejad2021bridging},
or only applies to specific policy classes
\citep[\eg][]{london2019bayesian,nguyen2021offline,sakhi2023pac,aouali2023exponential}.

In contrast to all this follow-up work, our method is computationally efficient and general in scope, not relying on realizability or small number of contexts  and handling any ``oracle-supported" policy class. Further, none of this follow-up work handles continuous actions.

In simultaneous work,%
\footnote{According to resp. publication dates on \texttt{arxiv.org}.}
\citet{aouali2023exponential} use a regularizer similar to ours, but with a specialized scope (mixtures of linear mixed-logit policies) and a different (PAC-Bayesian) perspective in the analysis.






\newcommand{\mycite}[1]{\citep{#1}}
\newcommand{\mycitet}[1]{\citet{#1}}

\ifthenelse{\boolean{arxiv-version}}{
\section{Preliminaries: Offline Policy Optimization (OPO)}
}
{
\section{PRELIMINARIES}
}
\label{sec:prelims}
\noindent{\bf Offline Policy Optimization (OPO).}
In contextual bandits, an agent interacts with an environment with a context space $\Xcal$ and an action space $\Acal$. In each round $i$, the agent observes a context $x_i\in\Xcal$, chooses an action $a_i\in\Acal$, and observes a loss $\ell_i(a_i)\in[0,1]$ (and nothing else). The pair $(x_i,\ell_i)$, where $\ell_i$ is a \emph{loss function} $\Acal\to[0,1]$, is drawn independently from some fixed (but unknown) distribution $\Dscr$.

As a form of inductive bias, a policy class $\Pi$ is given, where each policy $\pi \in \Pi$ is a randomized mapping from contexts to actions. $\pi(\cdot \mid x)$ specifies the distribution over actions given context $x$.
We define the risk for policy $\pi$ and the optimal policy $\pi^\star$, respectively, as
\begin{align*}
    R\rbr{\pi} & \coloneqq \EE_{(x, \ell) \sim \Dscr,\; a \sim \pi\rbr{\cdot \given x}}\sbr{\ell(a)}, \\ \pi^\star & \in {\textstyle \argmax_{\pi \in \Pi}}\; R(\pi).
\end{align*}

In \asedit{OPO}, one seeks a policy $\pi\in\Pi$ with low \emph{excess risk} $R(\pi) - R(\pi^\star)$.
The input is a dataset
    $\Scal = \cbr{(x_i, a_i, \ell_i(a_i))}_{i\in[N]}$
collected over $N$ rounds of contextual bandits by a known \emph{logging policy} $\mu: \Xcal \to \Delta(\Acal)$. Here, each action $a_i$ is drawn independently from distribution $\mu(\cdot \given x_i)$ specified by the logging policy.

We posit access to a \emph{computational oracle}: an algorithm for some hard but well-studied problem. Indeed, OPO tends to be NP-hard even with full feedback: given datapoints $(x_i,a_i,\ell_i)$, $i\in[N]$, where $\ell_i:\Acal\to[0,1]$ is the entire loss function. However, this is precisely \emph{cost-sensitive classification} (CSC), a classical and well-studied problem in supervised learning. Therefore, we posit an oracle which exactly solves CSC for a particular action set $\Acal$ and policy class $\Pi$.%
\footnote{Sometimes the oracle needs to handle losses that range on $\mathbb{R}_+$, so we allow this without further mention. For continuous actions, we use a standard CSC oracle that handles a finite action space, see Section~\ref{sec:cont-PL}.}
We allow only a small number of oracle calls; such algorithms are called \emph{oracle-efficient} (and our algorithms only call the oracle once). This is a standard approach in prior work on contextual bandits and OPO.

\xhdr{Naive solution: IPW.}
The prevailing approach to OPO from the statistical perspective is to construct an estimator
    $\widehat{R}:\Pi\to\R_+$
for the policy risk $R(\cdot)$ based on the dataset $\Scal$ and minimize $\widehat{R}(\pi)$ over the policy class $\Pi$. The standard estimator is the inverse probability weighting (IPW) estimator. Define:
\begin{align*}
\widehat{\ell}_i(\pi) & \coloneqq \frac{\pi(a_i \mid x_i)}{\mu(a_i \mid x_i)}\ell_i(a_i), \quad  \widehat{R}_{\IPW}(\pi) \coloneqq \frac{1}{N} \sum_{i=1}^N \widehat{\ell}_i(\pi),
\end{align*}
and set $\widehat{\pi}_{\IPW}  \in \argmin_{\pi \in \Pi} \widehat{R}_{\IPW} (\pi)$.
The IPW estimator is unbiased and (for finitely many actions) asymptotically consistent whenever the support of the logging policy $\mu$ is the entire action space (for any context).
To formalize this:
\begin{assumption}
\label{assump:support}
$\mu\rbr{a \given x} > 0$ for any context $x \in \Xcal$ and action $a \in \Acal$.
\end{assumption}

Further, the IPW-based approach is oracle-efficient: optimizing $\widehat{R}_{\IPW}(\cdot)$ is equivalent to calling the oracle with the loss vectors $a \mapsto \ell_i(a_i)/\mu(a_i \mid x_i) \cdot \one\{a=a_i\}$.

The \emph{variance} of the IPW estimator, specifically
    $V(\pi) := \Var\sbr{\widehat{\ell}_i(\pi)}$,
will be crucial in what follows. We call it the \emph{IPW-variance} of policy $\pi$. Denote
$V(\Pi) \coloneqq \sup_{\pi\in \Pi} V(\pi)$. Also important is the worst-case density ratios,
$\delta_{\sup}(\pi,\mu) \coloneqq \sup_{x\in\Xcal,a\in\Acal} \pi(a \mid x)/\mu(a \mid x)$ for a particular policy $\pi$ and
$\delta_{\sup}(\Pi,\mu) := \sup_{\pi\in\Pi} \delta_{\sup}(\pi,\mu)$.

\xhdr{Known issue: distribution shift.}
OPO with the IPW estimator can have poor finite-sample behavior, particularly when the support of the logging policy $\mu$ is highly non-uniform across contexts (e.g., see \citet{jin2021pessimism} and references therein).
Indeed, the standard (and essentially best) bound for IPW is that for any $\alpha \in (0, 1)$, with probability at least $1 - \alpha$,
\begin{multline}
\label{eq:oracle-inequality-no-pessimism}
R(\widehat{\pi}_{\IPW}) - R(\pi^\star)
    \lesssim \sqrt{V(\Pi)\cdot\tfrac{1}{N}\cdot\ln(|\Pi|/\alpha)}
    + \\ \supdensity(\Pi,\mu)\cdot \tfrac{1}{N}\cdot \ln \rbr{|\Pi|/\alpha},
\end{multline}
\noindent where $\lesssim$ ignores constant factors. The undesirable behavior of IPW manifests in the dependence on the worst-case IPW-variance $V(\Pi)$. In more detail, if $\pi^\star$ (or, more generally, any high-quality policy) has good coverage under the logging policy $\mu$, its
IPW-variance $V(\pi^\star)$
would be small, and we would hope that the excess risk of $\widehat{\pi}_{\IPW}$ would be correspondingly small. Unfortunately, this is not the case for IPW-based policy optimization; a low quality policy with large variance can significantly degrade the finite sample performance.

\xhdr{Known fix: pessimism.}
\emph{Pessimism} mitigates the effects of distribution shift in OPO.
One now minimizes an \emph{upper confidence bound} (UCB) on policy risk, penalizing policies with high uncertainty. Formally, one minimizes a pessimistic estimator
    $\widehat{R}:\Pi\times [0,1]\to\R_+$
which satisfies
    $\Pr\sbr{ \forall \pi \in \Pi: \widehat{R}(\pi,\alpha)\geq R(\pi) }\geq 1-\alpha$,
where $\alpha\in(0,1)$ is a parameter.%
\footnote{For implementation, it may be convenient to modify $\widehat{R}(\pi)$ so as to drop any additive ``constants'' that do not depend on $\pi$ (as this preserves the minimizer).}
This yields policy $\widehat{\pi}\in\Pi$ which w.h.p. satisfies
\begin{align}\label{eq:intro-pessimism}
R(\widehat{\pi})
        \leq \widehat{R}(\widehat{\pi},\alpha)
        \leq \min_{\pi \in \Pi} \widehat{R}(\pi, \alpha).
\end{align}

\asedit{Thus, $R(\widehat{\pi})$ is compared to the best policy risk guaranteed by the data, under this pessimistic estimator. Here, we interpret $\widehat{R}(\pi, \alpha)$ as the ``guaranteed policy risk", which tends to be smaller for policies of similar risk but better coverage in the data. Guarantees of this form are sometimes called \emph{best-effort guarantees} \citep[e.g.,][]{xie2021bellman,jin2021pessimism}.}

\begin{remark}\label{rem:pessimism}
\asedit{The key advantage of ``pessimistic" guarantee \eqref{eq:intro-pessimism} is that the estimator only needs to be ``sharp'' on \emph{some} good policy $\pi$, regardless of how well it can estimate other policies. This suffices to guarantee that the learned policy $\widehat{\pi}$ has low risk. In particular, if the logging policy $\mu$ has good support for $\pi^\star$, the best policy, then we can expect $\widehat{\pi}$ to perform well, even if other policies are poorly supported.}
\end{remark}

\asedit{To characterize the \emph{quality} of best-effort guarantees, one provides a data-independent upper bound for $\widehat{R}(\pi, \alpha) - R(\pi)$, and therefore for $R(\widehat{\pi})-R(\pi)$.}


\xhdr{Prior implementations of pessimism in OPO.}
\mycitet{swaminathan2015batch}
obtain an upper confidence bound (UCB) on the policy risk
using the empirical Bernstein (EB) inequality~\mycite{maurer2009empirical}.
Letting
    $\widehat{V}(\pi)
        \coloneqq \tfrac{1}{N(N-1)}\,\sum_{1\leq i < j \leq N} \rbr{\widehat{\ell}_i(\pi) - \widehat{\ell}_j(\pi)}^2$
be the sample variance, they minimize \asedit{the UCB on the policy risk},
\begin{equation}
\label{eq:EB-objective}
    \widehat{R}_{\IPW}(\pi) + \sqrt{\widehat{V}(\pi)\cdot\tfrac{1}{N}\cdot \ln (|\Pi|/\alpha)}.
\end{equation}
\asedit{This is advantageous as per Remark~\ref{rem:pessimism}. The following data-independent guarantee holds:}
letting $\widehat{\pi}_{\textnormal{IPW+EB}}$ be the learned policy, $\forall \pi \in \Pi$,
\begin{multline}
\label{eq:oracle-inequality-EB}
R(\widehat{\pi}_{\textnormal{IPW+EB}}) - R(\pi)
         \lesssim
             \sqrt{V(\pi)\cdot \tfrac{1}{N}\cdot \ln (|\Pi|/\alpha)}  + \\
            \supdensity(\Pi,\mu)\cdot\tfrac{1}{N-1}\cdot \ln(|\Pi|/\alpha).
\end{multline}
\asedit{It is also a best-effort guarantee (since it is obtained by upper-bounding \eqref{eq:EB-objective}). The technical advantage} over Eq.~\eqref{eq:oracle-inequality-no-pessimism} is that the worst-case IPW-variance $V(\Pi)$ is replaced with policy-specific $V(\pi)$. This method outperforms the vanilla IPW approach in experiments.

However, this approach
\asdelete{(as well as some other pessimistic approaches in prior work on OPO)}
suffers from \textbf{computational inefficiency}.
Particularly, the EB-based objective in Eq.~\eqref{eq:EB-objective}: (a) does not decompose across data points so it is not amenable to streaming or stochastic optimization methods,\footnote{However, \mycitet{swaminathan2015batch} proposed an approach to optimize this objective using stochastic gradient descent by iteratively (across epochs) optimizing an upper bound on the objective.}
(b) yields a non-convex landscape with a differentiable policy class, and (c) does not support non-differentiable policy classes except in highly specialized cases~\mycite{london2023boosted}. Note that non-differentiable policy classes are employed by a variety of methods, \eg those that train a regression model $f: \Xcal \times \Acal \to \RR$ and induce the policy $\pi_f: x \mapsto \argmin_a f(x,a)$.

\ifthenelse{\boolean{arxiv-version}}{
\section{Pseudo-loss with discrete actions}
}
{
\section{PSEUDO-LOSS}
}
\label{sec:disc-PL}
We introduce a new regularizer, dubbed \emph{pseudo-loss} (PL), and show that it provides pessimism-style guarantees for OPO while admitting an oracle-efficient implementation.
We focus on discrete actions here, i.e., when $|\Acal| < \infty$.%
\footnote{Given context $x$, each policy $\pi$ produces a probability mass function (p.m.f.) $\pi(\cdot \given x)$ over the actions. We call such policies mass-based.} Pseudo-loss is defined as follows:

\begin{definition}
\label{def:PL-confidence-term-discrete}
Given a policy $\pi$, pseudo-loss $\widehat{\PL}(\pi)$ and its expectation are
\begin{align*}
    \widehat{\PL} (\pi) &:= \frac{1}{N} \sum_{i=1}^N \sum_{a \in \Acal} \frac{\pi(a \given x_i)}{\mu(a \given x_i)} \\
    \PL(\pi)
        &:= \EE\sbr{\widehat{\PL} (\pi)}
        = \EE_{x \sim \Dscr}\sbr{\sum_{a \in \Acal} \frac{\pi(a \given x)}{\mu(a \given x)}}.
\end{align*}
\end{definition}

This is well-defined by Assumption~\ref{assump:support} (which we assume throughout without further mention).

We optimize this objective, parameterized by $\beta>0$:
\begin{equation}
\label{eq:objective-pl-discrete}
    \widehat{\pi}_{\textnormal{IPW+PL}, \beta} \in \argmin_{\pi \in \Pi} \widehat{R}_{\textnormal{IPW}}(\pi) + \beta\cdot \widehat{\PL}(\pi).
\end{equation}

\begin{remark}
Our regularizer is inspired by a technique in the EXP3.P algorithm \mycite{bandits-exp3}. \asedit{This algorithm works in a very different scenario: high-probability regret bounds in online adversarial bandits. In particular, it is not concerned with contexts, offline optimization, pessimism, or oracles. Our analysis is technically different (because of the different scenario), and, in some sense, stronger:} e.g., we remove the dependence on the range of $\beta$ in \refeq{eq:objective-pl-discrete}.
\end{remark}

\subsection{PL implements pessimism}

Consider the objective in \refeq{eq:objective-pl-discrete} plus some term $\Psi_\beta$ that does not depend on policy $\pi$ (so the optimization stays the same). We prove that the modified objective is an upper confidence bound on the policy risk.

We need some notation to define $\Psi_\beta$. Denote the supremum and infimum of the probability mass function (p.m.f.) induced by policy $\pi$ as, resp.,
    $\supdensity\rbr{\pi} \coloneqq \sup \pi(a \given x)$
and
    $\infdensity \rbr{\pi} \coloneqq \inf \pi(a \given x)$,
where both extrema are over contexts $x \in \Xcal$ and actions $a \in \Acal$. Write
$\supdensity\rbr{\Pi} = \sup_{\pi\in\Pi}\,\supdensity\rbr{\pi}$. Denote
\begin{align*}
\Delta(\Pi,\mu)
    :=  \max\rbr{\sqrt{\supdensity(\Pi) / \infdensity(\mu)},\; \supdensity(\Pi,\mu)},
\end{align*}
where $\supdensity(\Pi,\mu)$ was defined in Section~\ref{sec:prelims}. Thus:

\begin{lemma}
\label{lemma:IPW+PL-best-effort}
Fix $\alpha \in (0,1)$. Let
\begin{align}\label{eq:Psi}
\Psi_\beta
    &= \frac{O(\supdensity(\Pi)/\beta + \Delta(\Pi,\mu))\cdot \ln(|\Pi|/\alpha)}{N}.
\end{align}
With probability at least $ 1- \alpha$,
the following holds for all policies $\pi\in\Pi$ and $\beta > 0$:
\begin{align}\label{eq:lemma:IPW+PL-best-effort}
R(\pi) &\leq \widehat{R}_{\textnormal{IPW}}(\pi) +
    \beta \cdot \widehat{\PL}(\pi) + \Psi_\beta.
\end{align}
\end{lemma}

\subsection{Oracle-efficient implementation}

\begin{proposition}
\label{prop:IPW+PL-compatible-classification-oracle-discrete}
The optimization in \refeq{eq:objective-pl-discrete}
can be solved by a single call to any CSC oracle for policy class $\Pi$,
with modified loss vectors
\begin{align*}
a \mapsto \ell_i(a_i)/\mu(a_i \given x_i) \cdot \one\{a=a_i\} + \beta/\mu(a \given x_i).
\end{align*}
\end{proposition}

In practice, we treat $\beta$ as a hyper-parameter, following prior implementations of EB in~\citet{swaminathan2015batch},%
\footnote{\asedit{There, the bound in
Eq.~\eqref{eq:EB-objective} is not used directly,
because it may be too loose or the exact complexity of the policy class $\Pi$ may be unknown.}
Instead, a hyper-parameter similar to $\beta$ is introduced.}
with the goal of selecting a near-optimal value during a subsequent policy selection step.

\subsection{Performance guarantees}

\asedit{Lemma~\ref{lemma:IPW+PL-best-effort} immediately implies a best-effort guarantee via \eqref{eq:intro-pessimism}. Further, we obtain best-effort guarantees that are data-independent.%
\footnote{\asedit{The guarantees in Theorem~\ref{theo:IPW+PL-oracle-inequality-discrete} are also ``best-effort guarantees", as they are obtained by upper-bounding the right-hand side in \eqref{eq:lemma:IPW+PL-best-effort} and minimizing over all policies.}}
This is advantageous as per Remark~\ref{rem:pessimism} and similar to EB.}


\begin{theorem}
\label{theo:IPW+PL-oracle-inequality-discrete}
Fix $\alpha \in (0, 1)$. 
With probability at least $1 - \alpha$, for any $\beta > 0$ we have
\begin{align} \label{eq:theo:IPW+PL-oracle-inequality-discrete}
R(\widehat{\pi}_{\textnormal{IPW+PL},\beta})
     \leq \min_{\pi \in \Pi}\cbr{R(\pi) + O(\Phi)},
\end{align}
where $\Phi$ equals $\beta\cdot \widehat{\PL}(\pi)$ plus $\Psi_\beta$ from \refeq{eq:Psi}.

Further, with probability at least $1-\alpha$, for some $\beta^\star>0$ 
\asedit{\refeq{eq:theo:IPW+PL-oracle-inequality-discrete} holds with $\beta=\beta^*$ and $\Phi$ given by}
\begin{align}
\label{eq:discrete-beta-star}
\sqrt{\frac{\supdensity(\Pi)\cdot \PL(\pi)\cdot\ln\frac{|\Pi|}{\alpha}}{N}} +
        \frac{ \Delta(\Pi,\mu)\cdot \ln\frac{|\Pi|}{\alpha}}{N}.
\end{align}
\end{theorem}



\asedit{The key advantage over EB is the oracle-efficient implementation. However, our guarantee is worse than that of EB, since each terms in \refeq{eq:discrete-beta-star}
is lower-bounded by the respective term in \refeq{eq:oracle-inequality-EB}. (This is because
    $V(\pi) \leq \supdensity(\Pi)\cdot \PL(\pi)$
for any policy $\pi$, see Prop.~\ref{prop:PL-upper-bound-of-V}.)}

\ifthenelse{\boolean{arxiv-version}}{
\section{Empirical evaluation with discrete actions}
}
{
\section{EMPIRICAL EVALUATION}
}
\label{sec:disc-emp}
\noindent\textbf{Scope.} \asedit{We compare pseudo-loss (PL) to other ``general" approaches for pessimistic OPO that accommodate an arbitrary oracle: Empirical Bernstein and ``no pessimism''. We consider two representative oracles, based, resp., on policy gradient and linear regression. The full scope of our experiments is explained below. 

To keep our scope manageable, we do not compare PL to the numerous ``specialized" approaches for pessimistic OPO (see Related Work). While we believe such comparisons are somewhat unfair to the general method such as ours, we leave open the possibility that some of these specialized approaches are superior for their respective policy classes. Likewise, we did not consider deep learning oracles.%
\footnote{Aside from implementation complexity, replicating our experiments in a similarly systematic way would require much more compute power than we had access to.}}


\ifthenelse{\boolean{arxiv-version}}
{
\begin{table}[t]
\parbox{.6\linewidth}{
\centering
\caption{Options for the experimental conditions. }
\label{tb:discrete-setting-options}
\begin{tabular}{ll}
\toprule
Item    & Options    \\
\midrule
dataset size      &   \{$0.01$, $0.1$, $1$\}$\times 1$M \\
cost-type                 & \{real-valued, binary-valued\} \\
\# actions               & \{\# classes, 5 $\times$ \# classes\} \\
logging policy & \{$\mu_{\textnormal{good}, \epsilon = 0.1}, \mu_{\textnormal{good}, \epsilon = 0.01}, \mu_{\textnormal{bad}, \epsilon=0.1}$\}\\
\bottomrule
\end{tabular}
}
\parbox{.3\linewidth}{
\centering
\caption{Options for the algorithms. }
\label{tb:discrete-algorithm-options}
\begin{tabular}{ll}
\toprule
Item    & Options    \\
\midrule
regularizer             &   \{PL, EB, None\}\\
estimator               &   \{IPW, DR\}  \\
CSC oracle   &   \{PG, LR\}   \\
EB oracle               &   \{PG\} \\
\bottomrule
\end{tabular}
}
\end{table}
}
{
\begin{table*}[t]
\parbox{.58\linewidth}{
\centering
\caption{Experimental environment.}
\label{tb:discrete-setting-options}
\begin{tabular}{ll}
\toprule
Item    & Options    \\
\midrule
dataset size      &   \{$0.01$, $0.1$, $1$\}$\times 1$M \\
cost-type                 & \{real-valued, binary-valued\} \\
\#actions               & \{\#classes, 5 $\times$ \#classes\} \\
logging policy & \{$\mu_{\textnormal{good}, \epsilon = 0.1}, \mu_{\textnormal{good}, \epsilon = 0.01}, \mu_{\textnormal{bad}, \epsilon=0.1}$\}\\
\bottomrule
\end{tabular}
}
\hfill
\parbox{.35\linewidth}{
\centering
\caption{Options for the algorithms. }
\label{tb:discrete-algorithm-options}
\begin{tabular}{ll}
\toprule
Item    & Options    \\
\midrule
regularizer             &   \{PL, EB, None\}\\
estimator               &   \{IPW, DR\}  \\
CSC oracle   &   \{PG, LR\}   \\
EB oracle               &   \{PG\} \\
\bottomrule
\end{tabular}
}
\end{table*}
}

\xhdr{Experimental setup.}
We simulate offline contextual bandit instances from full-information classification datasets.%
\footnote{\asedit{This is a standard approach for contextual bandit experiments, \eg see  ~\citet{beygelzimer2009offset,dudik2014doubly,wang2017optimal,su2019cab,su2020doubly}.}}
This semi-synthetic setup gives us the ground-truth for evaluation and allows to precisely vary experimental conditions. We use four datasets from OpenML, \asedit{with 1M datapoints, 14-36 real-valued features, and 6-26 classes (see Table~\ref{tb:discrete-data-detail} in Appendix~\ref{sec:additional-exp}). 

For the experimental environment, we vary the following factors:} dataset size, cost-type (binary- vs. real-valued), number of actions, and logging policy (see Table~\ref{tb:discrete-setting-options}). \asedit{We try all  $2\times2\times2\times3=24$ possible environments.} In particular, we use the technique of~\citet{foster2018practical} to vary the cost-type and the number of actions. We use 3 logging policies, denoted
    $\mu_{\textnormal{good}, \epsilon}$ (resp., $\mu_{\textnormal{bad}, \epsilon}$),
by training good (resp., bad) policies and mixing with $\epsilon$-probability uniform exploration.

For methods, we primarily compare three options for pessimistic regularizer: pseudo-loss (PL), Empirical Bernstein (EB), and ``no regularizer'' (None). \asedit{Further, there are two choices for the risk estimator:} inverse probability weighting (IPW) and a doubly robust estimator (DR)~\mycite{dudik2014doubly}, which is also compatible with CSC oracles. \asedit{Finally, we consider two different underlying optimizers for the CSC oracle:} policy gradient (PG) using a linear+softmax policy architecture and a linear regression approach (LR) where we fit a linear model to the loss for each action and define the policy to be greedy with respect to the predicted losses. (However, EB cannot accommodate the linear regression approach since the policy architecture is not differentiable.) See Table~\ref{tb:discrete-algorithm-options} for an overview. \asedit{Thus, we have 4 possible (estimator, CSC oracle) configurations for PL and None, and only 2 for EB.}


We tune hyper-parameters for each method using a policy selection rule based on the EB bound in Eq.~\eqref{eq:EB-objective} using a 50/50 split of the data for policy optimization and selection respectively.\footnote{Policy selection is another instance of OPO with a enumerably-small policy class, with one policy for each hyper-parameter setting of the OPO method. Since the class is small, computational efficiency is less of a concern and we can use the statistically tighter EB bound.} See Appendix~\ref{sec:additional-exp} for hyper-parameters for each method. All results are based on 50 replicates with mean and standard errors reported.

\begin{figure*}[!t]
\centering
\includegraphics[width=.46\textwidth]{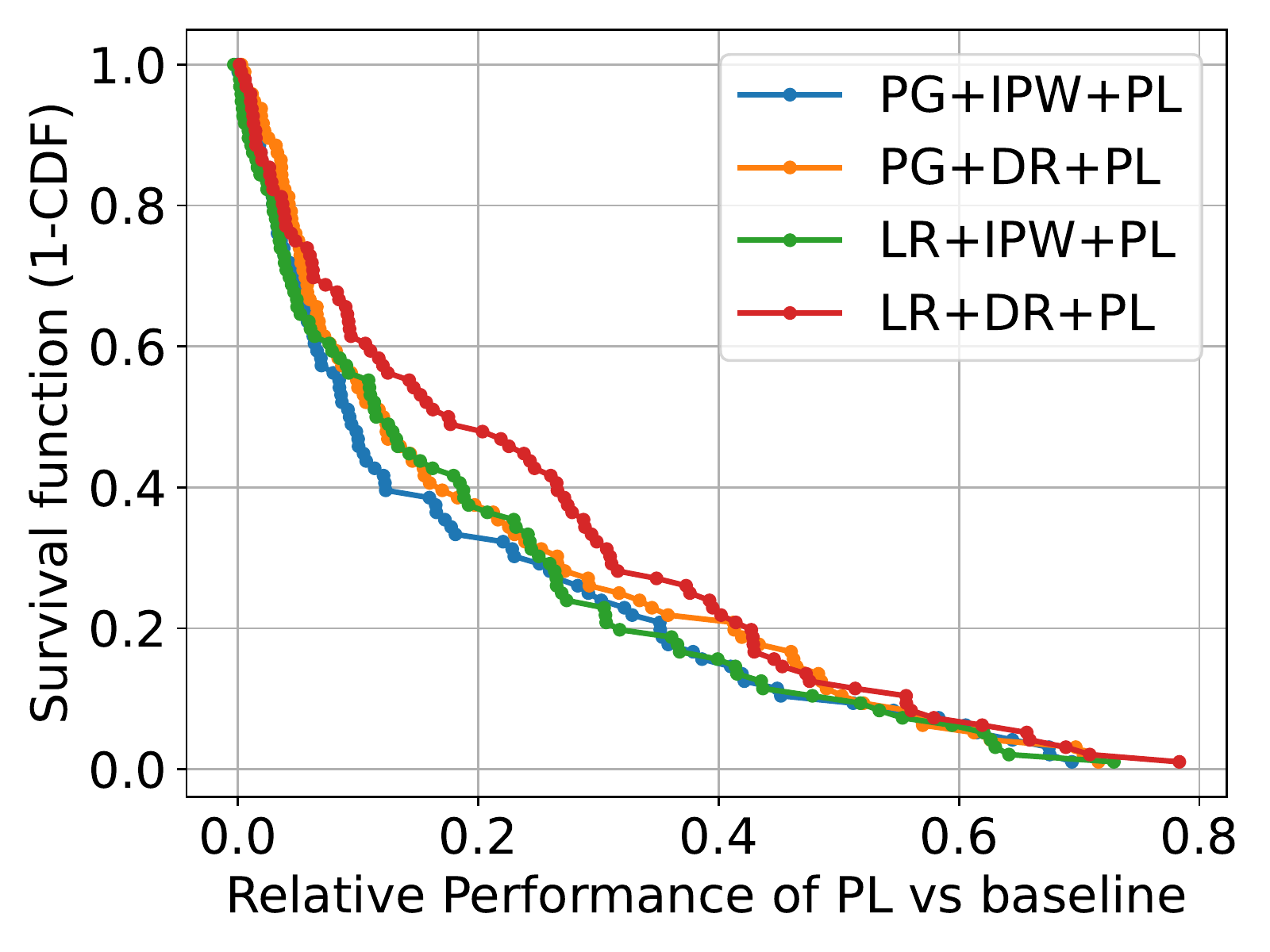}
\includegraphics[width=.46\textwidth]{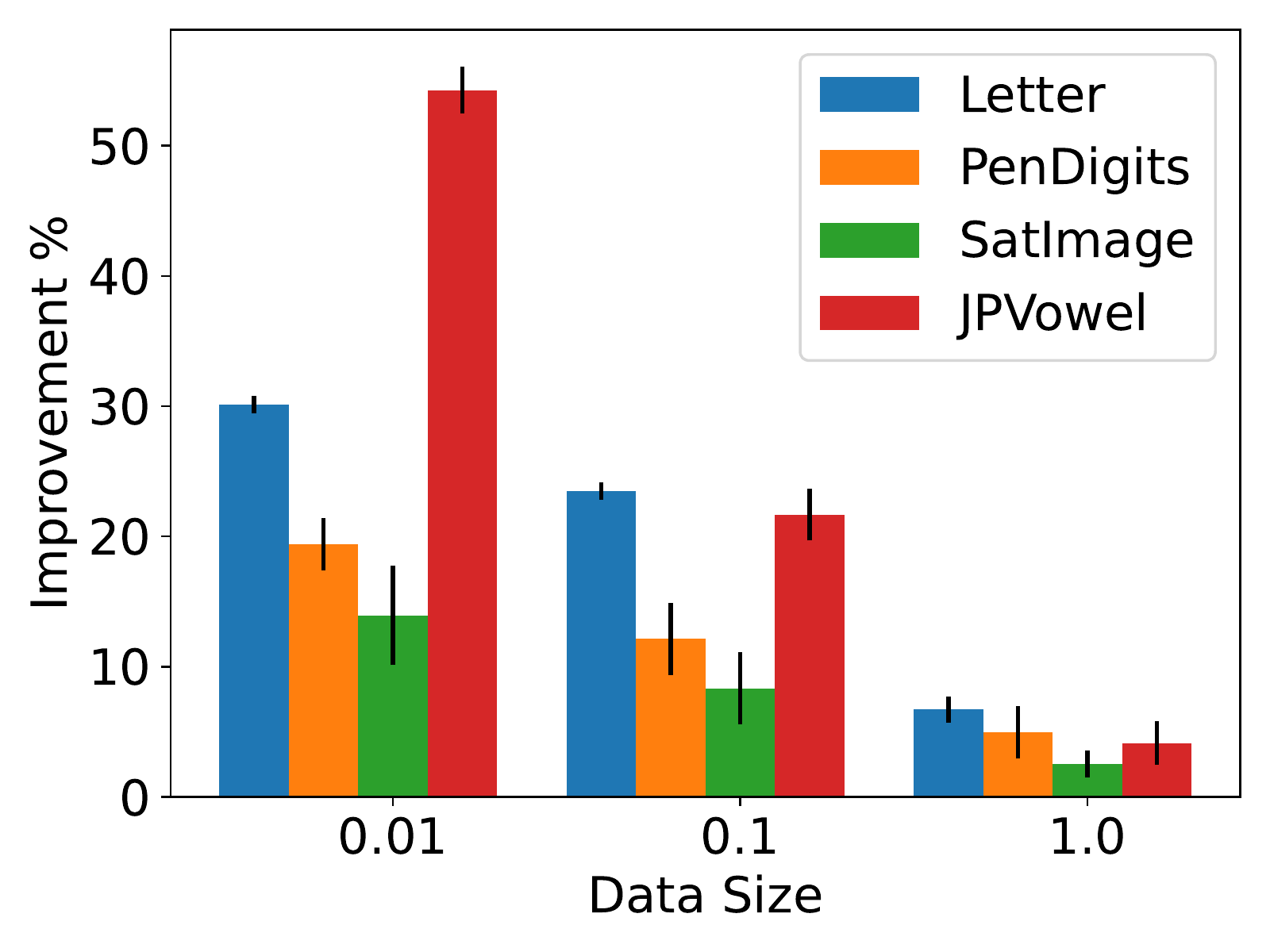}
\captionof{figure}{Relative improvement (\RelImp) for PL against the baseline with no pessimism: mean over all runs.\\
\textbf{Left:} $1-\text{CDF}$, the empirical cumulative density function of \RelImp across all (dataset, environment) pairs. Each curve corresponds to (CSC oracle, risk estimator) pair. \\
\textbf{Right:} \RelImp (mean $\pm$ $2$ standard errors) for a particular (dataset, environment) pair and the best-performing (CSC oracle, risk estimator) pair. Each bar corresponds to a (dataset, dataset size) pair.\\
\textbf{Details:} 50 runs; 24 environments, as per Table~\ref{tb:discrete-setting-options}; 4 datasets, as per Table~\ref{tb:discrete-data-detail}.  \\
The environment on the right is: real-valued cost, $\mu_{\textnormal{good},\epsilon=0.1}$, and \#actions = \#classes.}
\label{fig:main_discrete}
\end{figure*}

\xhdr{Results.} Across a wide range of experimental conditions, we find that using the PL regularizer consistently outperforms vanilla OPO with no pessimism. These results are visualized in Figure~\ref{fig:main_discrete} \asedit{(left), where we plot the relative performance improvement (\RelImp) of PL over the baseline with no pessimism, averaged over all runs. Specifically, each curve corresponds to a particular (CSC oracle, risk estimator) pair, and represents the empirical CDF of \RelImp across all $4\times 24$ (dataset, environment) pairs.}
The median \RelImp is $11.7\%$. In our experiments, PL was, essentially, \emph{never worse} than the no-pessimism baseline.%
\footnote{More precisely, PL prevails on $99.5\%$ of the conditions.}
We note that PL is especially helpful when the sample size is small relative to the number of actions, which is consistent with the theory that pessimism is particularly helpful in the non-asymptotic regime.

We also compare with Empirical Bernstein (EB) approach of \citet{swaminathan2015batch}. 
We'd expect that EB outperforms PL statistically (due to Prop.~\ref{prop:PL-upper-bound-of-V}) and indeed it does: \asedit{EB offers median \RelImp of $19.1\%$}
for the configurations where it is applicable.
However, EB cannot be instantiated with the linear regression (LR) optimizer. LR, when coupled with PL, sometimes yields the best overall performance, so that when selecting the best algorithm configuration for each regularizer PL beats EB on $26\%$ of the settings. We note that our implementation of the PG optimizer for EB is an order of magnitude slower than PL.

\asedit{We present detailed visualizations in Appendix~\ref{sec:additional-exp}, isolating each algorithm and environment configuration. Representative visualizations are displayed in Figure~\ref{fig:main_discrete} (right)
and Table~\ref{tb:exp-d-real-eps0.1-size0.1} (just for two datasets out of four). 
We also report computation times.}

\begin{table}[ht]
\caption{Performance of different OPO methods: mean $\pm$ two standard errors over 50 runs. Bold numbers represent the best performance within each (CSC oracle, estimator) pair. Boxed numbers represent the best across all algorithmic configurations. \\ This experiment:  
real-valued cost, logging policy $\mu_{\textnormal{good}, \epsilon=0.1}$, data size $\times0.1$, \# actions = \# classes.}

\centering
\begin{tabular}{lll}\toprule
Risk $\times$ 100	& Letter	& PenDigits		\\
\midrule
PG+IPW+PL	& \textbf{31.6±0.1}	& 22.0±0.3		\\
PG+IPW+EB	& 32.4±0.1	& \textbf{20.4±0.4}	\\
PG+IPW	& 45.5±0.8	& 26.4±0.8		\\
\midrule
PG+DR+PL	& \fbox{\textbf{31.5±0.1}}	& 18.5±0.3		\\
PG+DR+EB	& 37.0±0.4	& \fbox{\textbf{15.7±0.3}}		\\
PG+DR	& 43.1±0.6	& 21.3±0.6		\\
\midrule
LR+IPW+PL	& \textbf{31.8±0.0}	& \textbf{23.1±0.3}		\\
LR+IPW	& 42.5±0.4	& 28.4±0.7		\\
\midrule
LR+DR+PL	& \textbf{31.8±0.1}	& \textbf{22.9±0.2}		\\
LR+DR	& 42.0±0.3	& 27.2±0.6		\\
\bottomrule
\end{tabular}
\label{tb:exp-d-real-eps0.1-size0.1}
\end{table}


\asedit{\xhdr{Suggested guidance for practitioners:}}

    \emph{Pessimism should always be employed for offline policy optimization. If running time is not a concern and the policy architecture supports it, use empirical Bernstein; otherwise, use the PL estimator.}

\ifthenelse{\boolean{arxiv-version}}{
\section{Pseudo-loss with continuous actions}
}
{
\section{CONTINUOUS ACTIONS}
}
\label{sec:cont-PL}
We extend our approach to continuous actions, namely the one-dimensional action space $\Acal = [0,1]$. We posit that each policy $\pi\in \Pi$ produces a probability density function (p.d.f.) $\pi(\cdot \given x)$ over the action space $\Acal$ for each context $x$. We call such policies \emph{density-based}.

\begin{definition}
\label{def:PL_confidence_term-continuous}
For a density-based policy $\pi$, pseudo-loss $\widehat{\PL} (\pi)$ and its expectation are
\begin{align*}
    \widehat{\PL} (\pi) & = \frac{1}{N} \sum_{i=1}^N \int_{a \in \Acal} \frac{\pi(a \given x_i)}{\mu(a \given x_i)} \dif{a} \\ \PL(\pi)
        & := \EE\sbr{\widehat{\PL}}
        = \EE_{x \sim \Dscr}\sbr{\int_{a \in \Acal} \frac{\pi(a \given x)}{\mu(a \given x)} \dif{a}}.
\end{align*}
\end{definition}


\asedit{Much of our analysis seamlessly carries over to density-based policies. Specifically, Lemma~\ref{lemma:IPW+PL-confidence-discrete} and Theorem~\ref{theo:IPW+PL-oracle-inequality-discrete} (as well as Propositions
\ref{prop:PL-upper-bound-of-V} and~\ref{prop:PL-and-expected-PL} in the appendix)} all hold for density-based policies if $\supdensity(\pi)$ and $\infdensity(\pi)$ denote the supremum and infimum of the p.d.f induced by $\pi$. All these results are proved similarly to the discrete-action case, except the sums over $\Acal$ are replaced with integrals over $[0,1]$. Consequently, these proofs are omitted.

Next, we need to transform the learning problem. This is because (a) CSC algorithms typically can only handle finitely many actions, and (b) the variance of IPW estimator might be infinite when we consider deterministic policies~\mycite{kallus2018policy}, making learning impossible. Thus, inspired by prior work on contextual bandits with continuous actions~\mycite{krishnamurthy2020contextual,zhu2022contextual}, we transform the OPO problem with the original (density-based) policy class $\Pi$ to a CSC problem with a mass-based policy class (denoted $\widetilde{\Pi}_K$) such that each policy in $\Pi$ is a \emph{smoothed version} of some policy in $\widetilde{\Pi}_K$.

Formalizing this requires some care. We start with a class of mass-based policies over $K$ actions, denoted  $\widetilde{\Pi}_K$, for each $K\in\N$. We interpret these $K$ actions as \emph{surrogate actions}:
\[ \widetilde{\Acal}_K \coloneqq \cbr{\widetilde{a}_i}_{i\in[K]} = \cbr{\tfrac{2i-1}{2K}}_{i\in[K]} \subset \Acal.\]
Next, form  density-based policies
\begin{align*}
    \Pi_{K, H} \coloneqq \{\;\textnormal{Smooth}_H(\widetilde{\pi}) : \widetilde{\pi} \in \widetilde{\Pi}_K\;\},
\end{align*}
where the smoothed policy
    $\textnormal{Smooth}_H(\widetilde{\pi})$
selects an action $a$ given a context $x$ through the following process:
\begin{multline*}
\widetilde{a} \sim \widetilde{\pi}\rbr{\cdot \given x},\; \text{then }
a \sim \\ \textnormal{Uniform}\rbr{\sbr{\max\rbr{0, \widetilde{a} - H / 2}, \min\rbr{1, \widetilde{a} + H / 2}}}.
\end{multline*}
We call $H$ the \emph{bandwidth} of smoothing.%
\footnote{Such ``smoothing" was introduced in the online setting in~\citet{krishnamurthy2020contextual}.}

We can optimize the PL-based objective in Eq.~\eqref{eq:objective-pl-discrete} over $\Pi = \Pi_{K, H}$
by calling a CSC oracle over $\widetilde{\Pi}_{K}$.

\begin{proposition}
\label{prop:IPW+PL-compatible-classification-oracle-continuous}
Fix $K,H\in\N$. Consider the density-based policy class $\Pi = \Pi_{K, H}$ as constructed above. Then the objective in \refeq{eq:objective-pl-discrete} can be optimized via a single call to a CSC oracle for the mass-based policy class $\widetilde{\Pi}_{K}$, with suitably modified loss functions (details in Appendix~\ref{sec:proof}).
\end{proposition}

We characterize generalization performance of
pseudo-loss and smoothed policy class $\Pi = \Pi_{K, H}$. Note that $\supdensity\rbr{\pi}\leq 2 / H$ for any policy $\pi \in \Pi$.

\begin{corollary}
\label{coro:IPW+PL-oracle-inequality-for-smoothed-policies}
Fix $\alpha \in (0, 1)$. 
With probability at least $1 - \alpha$, for any $\beta > 0$ we have
\asedit{\refeq{eq:theo:IPW+PL-oracle-inequality-discrete} holds with}
\begin{align*}
\Phi = \beta\cdot \widehat{\PL}(\pi)
    + \frac{
        \rbr{1/\beta +1/\infdensity(\mu)}
        \cdot\ln\rbr{|\Pi| / \alpha}}{NH}.
\end{align*}
Further, with probability at least $1-\alpha$, for some $\beta^\star>0$
\refeq{eq:theo:IPW+PL-oracle-inequality-discrete} holds with $\beta = \beta^*$ and
\begin{align*}
\Phi = \sqrt{\frac{\PL(\pi)\cdot\ln(|\Pi|/\alpha)}{NH}} +
         \frac{ \ln(|\Pi|/\alpha)}{NH\infdensity(\mu)}.
\end{align*}
\end{corollary}

We also discuss how to construct the range of hyper-parameters $H$ and $K$ in Appendix~\ref{sec:hyper-parameter-continuous-action}.

\ifthenelse{\boolean{arxiv-version}}{
\section{Empirical evaluation with continuous actions}
}
{
\subsection{Empirical Evaluation}
}
\label{sec:cont-emp}
We conduct an empirical study in the continuous-action setting, with \asedit{the same scope as in Section~\ref{sec:disc-emp}}.

\xhdr{Experimental setup.}
We follow~\mycite{bietti2021contextual,majzoubi2020efficient,zhu2022contextual} to simulate continuous-action contextual bandit instances from 5 OpenML regression datasets~\mycite{OpenML2013}, \asedit{with 160K-5M datapoints and 9-32 features (see Table~\ref{tb:coninuous-data-detail} in Appendix~\ref{sec:additional-exp} for details).} We convert a regression example $(x,y)$, $y\in\R$ to a contextual bandit example by defining the loss as $\ell(a) = |a - y|$. 

For the experimental environment, we vary two factors: the dataset size and logging policy. The logging policies $\mu_\epsilon$ are obtained by training a regression model on the original dataset, smoothing the prediction with bandwidth $0.1$ and mixing with the uniform-at-random policy with proportion $\epsilon \in \{0.1, 0.01\}$.


\begin{table}[h]
\centering
\caption{Experimental environment: continuous actions}
\label{tb:continuous-setting-options}
\begin{tabular}{ll}
\toprule
Item    & Options    \\
\midrule
dataset size      &   \{$0.01$, $0.1$, $1$\}$\times \texttt{ActualSize}$  \\
logging policy & $\mu_{\epsilon}$, $\epsilon \in \{0.1, 0.01\}$\\
\bottomrule
\end{tabular}
\end{table}

\begin{figure*}[ht]
\centering
\includegraphics[width=.46\textwidth]{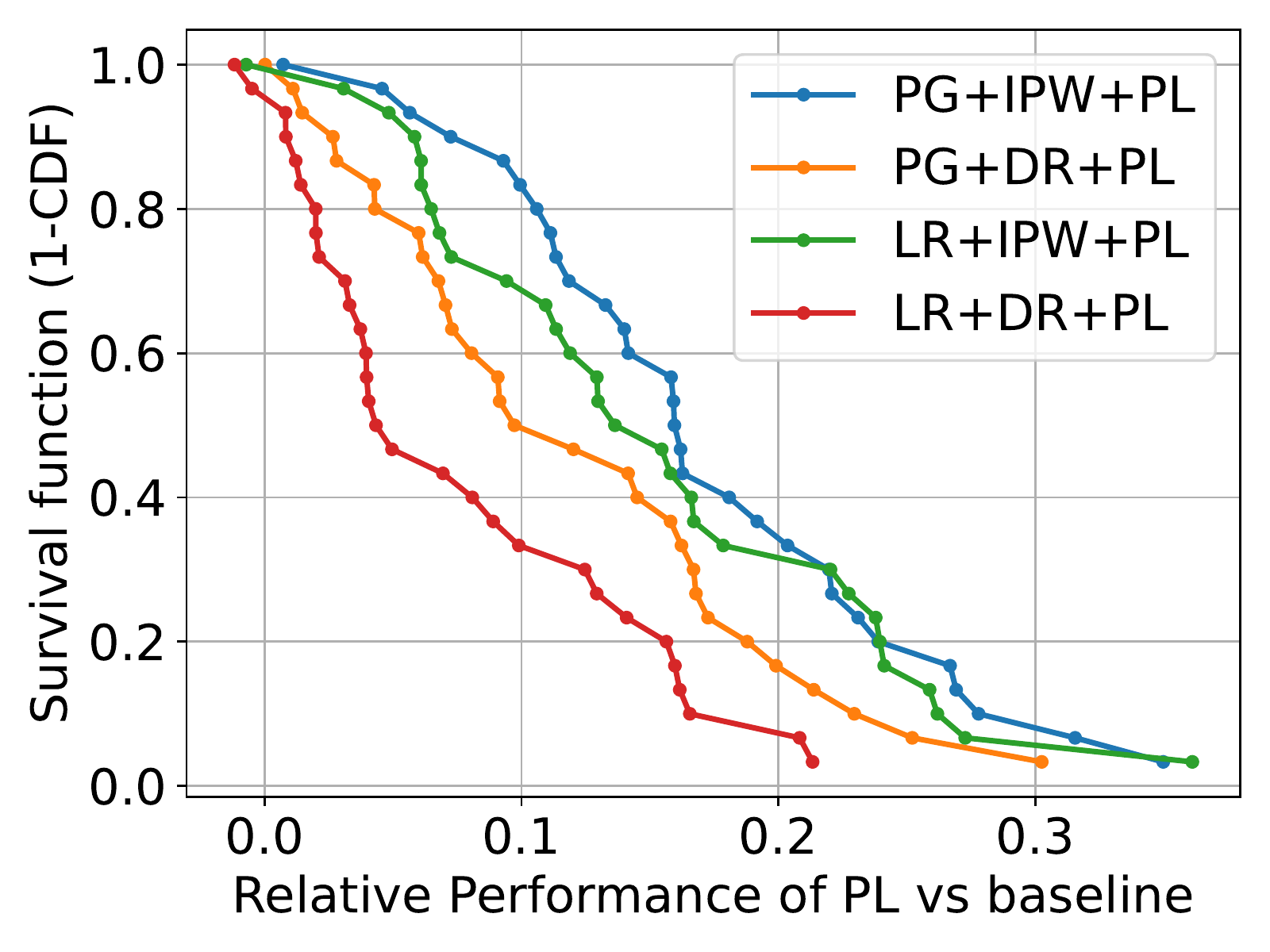}
\includegraphics[width=.46\textwidth]{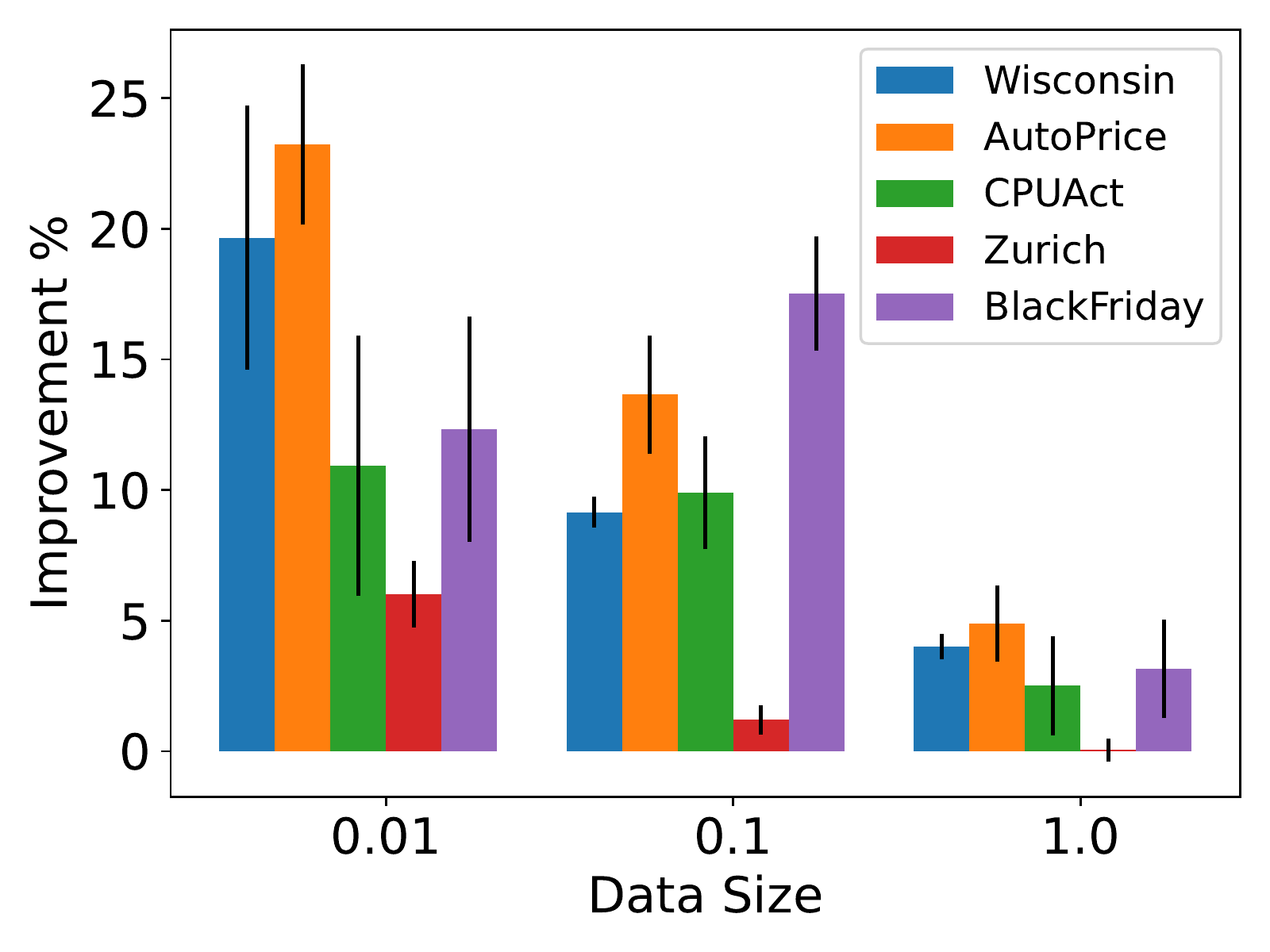}
\captionof{figure}{Same semantics as in Figure~\ref{fig:main_discrete}. Details: 10 runs,
6 environments (Table~\ref{tb:continuous-setting-options}) and 5 datasets (Table~\ref{tb:coninuous-data-detail}.) The environment on the right has  logging policy $\mu_{\epsilon=0.1}$.}
\label{fig:main_continuous}
\end{figure*}

The algorithmic configurations are the same as those for the discrete-action setting (the 10 options in Table~\ref{tb:discrete-algorithm-options}). We optimize hyper-parameters as before, using the EB bound for policy selection. In this setup, we also consider hyper-parameters $K \in \{10,20,50,100\}$ and $H \in \{0.01, 0.02, 0.05, 0.1\}$. Due to the large number of hyper-parameters and the computational overhead of EB, we only run EB for the small dataset sizes ($0.01$ and $0.1$ fraction of the original dataset).

\xhdr{Results overview.}
\asedit{Figure~\ref{fig:main_continuous} (left) visualizes the relative improvement (\RelImp) over vanilla policy optimization with no pessimism. As with discrete actions, PL-based pessimism exhibits consistent significant improvement across environmental and algorithmic configurations. The median \RelImp is $12\%$ and we see improvement in $97.5\%$ of the experimental conditions. In comparison, on configurations when EB is applicable, EB offers a median \RelImp of $25.2\%$, while PL offers a median \RelImp of $15.8\%$.}


\asedit{We present detailed visualizations in Appendix~\ref{sec:additional-exp}, isolating each algorithm and environment like we did for discrete actions.} Representative visualizations are in Figure~\ref{fig:main_continuous} (right)
and Table~\ref{tb:exp_c_eps0.1_size0.1} (just for 2 datasets). We also report computation times.

\begin{table}[ht]
\caption{Same semantics as in Table~\ref{tb:exp-d-real-eps0.1-size0.1}.
This experiment: logging policy $\mu_{ \epsilon=0.1}$, data size $\times 0.1$, 10 runs.}
\centering
\begin{tabular}{llllll}
\toprule
Risk * 100	& Wisconsin	& AutoPrice		\\
\midrule
PG+IPW+PL	& 22.7±0.8	& 16.9±1.4		\\
PG+IPW+EB	& \fbox{\textbf{21.5±0.1}}	& \textbf{14.7±0.3}	\\
PG+IPW	& 26.6±1.1	& 20.2±0.9	\\
\midrule
PG+DR+PL	& 21.8±0.1	& 15.3±0.2		\\
PG+DR+EB	& \fbox{\textbf{21.5±0.3}}	& \fbox{\textbf{14.4±0.1}}		\\
PG+DR	& 24.0±0.2	& 18.0±0.9		\\
\midrule
LR+IPW+PL	& \textbf{24.1±1.4}	& \textbf{18.5±0.8}		\\
LR+IPW	& 27.7±0.6	& 21.1±0.8	\\
\midrule
LR+DR+PL	& \textbf{23.1±0.8}	& \textbf{17.7±0.9}		\\
LR+DR	& 26.4±0.4	& 18.7±0.5		\\
\bottomrule
\end{tabular}
\label{tb:exp_c_eps0.1_size0.1}
\end{table}


\ifthenelse{\boolean{arxiv-version}}{
\section{Discussion}
}
{
\section{DISCUSSION}
}
\label{sec:discussion}
We develop a new pessimistic approach for offline policy optimization in contextual bandits based on the pseudo-loss (PL) regularizer. The approach offers a favorable balance between computational complexity and statistical performance. It is oracle efficient and thus supports a wide range of policy classes and underlying optimization methods while offering a best-effort guarantee analogous to, but slightly worse than, prior computationally inefficient approaches. We observe this balance in our experiments, offering the guidance that pessimism should always be used and that PL should be used when computation is a concern or when sharper approaches are not applicable. 

\xhdr{Limitations.}
Our experimental study is semi-synthetic, transforming fully-labeled classification and regression datasets to contextual bandit instances. A standard practice in most prior work on contextual bandits, it gives us access to ground truth, but may not accurately reflect the performance in production.

Our experiments demonstrate that OPO methods are rather sensitive to a variety of factors, paralleling similar observations in Offline RL~\citep{swaminathan2015batch,joachims2018deep,su2019cab,wang2021instabilities}. Some factors, notably the choice of optimizer (PG vs LR), choice of estimator (IPW vs DR), and dataset quality, appear in our experimental results. In our preliminary experiments, we found that other factors may also be relevant, e.g., optimizer hyper-parameters. We did not rigorously evaluate these factors to keep the experiments at a manageable level of complexity. However,
understanding whether/how our qualitative findings carry over to production environments
is an important next step.




{
\bibliographystyle{abbrvnat}
\bibliography{refs,bib-abbrv,bib-slivkins,bib-bandits,bib-AGT}
}

\newpage
\appendix
\onecolumn
\section*{APPENDIX}

The appendix consists the proofs in Appendix~\ref{sec:proof} and detailed empirical evaluation in Appendix~\ref{sec:additional-exp}. The evaluation for discrete actions is in Appendix~\ref{sec:additional-exp-discr} on p. \pageref{sec:additional-exp-discr}, and for continuous actions in Appendix~\ref{sec:additional-exp-cont} on p. \pageref{sec:additional-exp-cont}.


\section{Theoretical analysis}
\label{sec:proof}
We invoke Bennett's inequality in our analysis.
\begin{lemma}[\citet{bennett1962probability}]
\label{lemma:bennett}
Let $z, z_1, \dots, z_N$ be \iid\, random variables with values in $[0, 1]$, For any $\alpha\in(0,1)$, with probability at least $1-\alpha$, we have
\begin{equation*}
\EE[z] - \frac{1}{N}\sum_{i=1}^Nz_i \leq \sqrt{ \frac{2\mathrm{Var}(z)\ln(1/\alpha) }{N}} + \frac{\ln(1/\alpha)}{3N}.
\end{equation*}
\end{lemma}

\subsection{Confidence intervals}

We upper-bound the IPW-variance of a given policy $\pi$ in terms of $\PL(\pi)$, and characterize the difference between the pseudo-loss and its expectation. Then, we characterize confidence intervals for the policy risk via pseudo-loss.


\begin{proposition}
\label{prop:PL-upper-bound-of-V}
$V(\pi) \leq \supdensity(\pi)\cdot \PL(\pi)$
for any policy $\pi$.
\end{proposition}

\begin{proof}
 \begin{equation*}
 \begin{split}
 V(\pi) &= \EE_{(x, \ell) \sim \Dscr, a\sim\mu(\cdot\given x)}\sbr{\rbr{\frac{\pi(a \given x)}{\mu(a \given x)}\ell(a)}^2} - \EE^2_{(x, \ell) \sim \Dscr, a\sim\mu(\cdot\given x)}\sbr{\frac{\pi(a \given x)}{\mu(a \given x)}\ell(a)}\\
 &\leq \EE_{(x, \ell) \sim \Dscr, a\sim\mu(\cdot\given x)}\sbr{\rbr{\frac{\pi(a \given x)}{\mu(a \given x)}\ell(a)}^2} \\
 &\leq \EE_{x \sim \Dscr}\sbr{\sum_{a \in \Acal}{\frac{\pi^2(a \given x)}{\mu(a \given x)}} } \\
 &\leq \supdensity(\pi) \EE_{x \sim \Dscr}\sbr{\sum_{a \in \Acal}\frac{\pi(a \given x)}{\mu(a \given x)}} \\
 &= \supdensity(\pi) \textnormal{PL}(\pi).
 \end{split}
 \end{equation*}
The second inequality holds since $0 \leq \ell^2(a)\leq 1$. The third inequality holds because $\pi(a\given x) \leq \supdensity(\pi)$.
\end{proof}


\begin{proposition}
\label{prop:PL-and-expected-PL}
For any policy $\pi$ and any $\alpha \in (0,1)$, with probability at least $1 - \alpha$,
\begin{align*}
\tfrac{1}{2}\; \PL(\pi) - O\rbr{\frac{\ln(1/\alpha)}{N\cdot \infdensity(\mu)}} \leq \widehat{\PL}(\pi)  \leq \tfrac{3}{2}\;\PL(\pi) + O\rbr{\frac{\ln(1/\alpha)}{N\cdot \infdensity(\mu)}}.
\end{align*}
\end{proposition}

\begin{proof}
We prove Proposition~\ref{prop:PL-and-expected-PL} with exact constants, which states that for any policy $\pi$ and any $\alpha \in (0,1)$, with probability at least $1 - \alpha$,
\begin{equation*}
    \frac{1}{2}\; \PL(\pi) - \frac{4\ln(2/\alpha)}{3N \infdensity(\mu)} \leq \widehat{\PL}(\pi) \leq \frac{3}{2}\;\PL(\pi) + \frac{4\ln(2/\alpha)}{3N \infdensity(\mu)}.
\end{equation*}
First, we bound the variance of $\sum_{a\in\Acal}\pi(a\given x)/\mu(a\given x)$. For any $x \in \Xcal$,
\begin{equation*}
\sum_{a\in\Acal}\frac{\pi(a\given x)}{\mu(a\given x)} \leq \frac{1}{\infdensity(\mu)}\sum_{a\in\Acal} \pi(a\given x) = \frac{1}{\infdensity(\mu)}.
\end{equation*}
Therefore,
\begin{equation}
\label{eq:variance-upper-bound-PL}
\Var\rbr{\sum_{a\in\Acal}\frac{\pi(a\given x)}{\mu(a\given x)}}\leq \EE\sbr{\rbr{\sum_{a\in\Acal}\frac{\pi(a\given x)}{\mu(a\given x)}}^2} \leq \frac{1}{\infdensity(\mu)} \EE\sbr{\sum_{a\in\Acal}\frac{\pi(a\given x)}{\mu(a\given x)}}  = \frac{\textnormal{PL}(\pi)}{ \infdensity(\mu)}.
\end{equation}
Applying Lemma~\ref{lemma:bennett} to \iid\, random variables $\cbr{\sum_{a\in\Acal}\frac{\pi(a\given x_i)}{\mu(a\given x_i)}}_{i\in[N]}$ (precisely by multiplying the random variables by $\infdensity(\mu)$ so their range is $[0,1]$, and considering both tails via a union bound),
 we have that for any $\alpha \in (0,1)$, with probability $1-\alpha$, it holds that
\begin{equation*}
\begin{split}
    \left|\textnormal{PL}(\pi) - \widehat{\textnormal{PL}}(\pi)\right| &\leq \sqrt{\Var\rbr{\sum_{a\in\Acal}\frac{\pi(a\given x)}{\mu(a\given x)}}\ln(2/\alpha)  \frac{2}{N}} + \frac{\ln(2/\alpha)}{3N\infdensity(\mu)}\\
&\leq \sqrt{\frac{2\textnormal{PL}(\pi)\ln(2/\alpha)}{\infdensity(\mu)N}} + \frac{\ln(2/\alpha)}{3N\infdensity(\mu)}\\
&\leq \frac{\textnormal{PL}(\pi)}{2} + \frac{\ln(2/\alpha)}{N\infdensity(\mu)} + \frac{\ln(2/\alpha)}{3N\infdensity(\mu)}\\
&=\frac{\textnormal{PL}(\pi)}{2} + \frac{4\ln(2/\alpha)}{3N\infdensity(\mu)}.
\end{split}
\end{equation*}
The second inequality is Eq.~\eqref{eq:variance-upper-bound-PL}. The last inequality follows from the AM-GM inequality. 
\end{proof}


\begin{lemma}
\label{lemma:IPW+PL-confidence-discrete}
Fix any policy $\pi$ and $\alpha \in (0,1)$. With probability at least $ 1- \alpha$,
\begin{align}
\label{eq:lemma:IPW+PL-confidence-discrete}
|R(\pi) - \widehat{R}_{\textnormal{IPW}}(\pi)|
    \lesssim  \sqrt{\tfrac{\ln(1/\alpha)}{N}\;\supdensity(\pi)\cdot \widehat{\PL}(\pi)}
    +\tfrac{\ln(1/\alpha)}{N}\;\Delta(\pi,\mu).
\end{align}
\end{lemma}

\begin{proof}
We prove a version of the lemma with exact constants: namely,
\refeq{eq:lemma:IPW+PL-confidence-discrete} is spelled out as
\begin{equation*}
|R(\pi) - R(\widehat{\pi}_{\textnormal{IPW}})|
    \leq  \sqrt{\tfrac{3\ln(4/\alpha)}{N}\;\supdensity(\pi)\cdot \widehat{\PL}(\pi)}
    +\tfrac{\ln(4/\alpha)}{N} \max\rbr{2\sqrt{\frac{8\supdensity(\pi)} { 3\infdensity(\mu)}}, \frac{2}{3}\supdensity(\pi,\mu)}.
\end{equation*}

Applying Bennett's inequality (Lemma~\ref{lemma:bennett}) to \iid\, random variables $\cbr{\widehat{\ell}_i(\pi)}$ (precisely by dividing by $\supdensity(\pi, \mu)$, and also applying a union bound to account for both tails), we have that with probability at least $1-\alpha/2$,
\begin{equation*}
\begin{split}
\left|R(\pi) - \widehat{R}_{\textnormal{IPW}}(\pi)\right| &\leq \sqrt{\frac{2V(\pi)\ln(4/\alpha)}{N}} + \frac{\ln(4/\alpha)\supdensity(\pi,\mu)}{3N}\\
&\leq \sqrt{\frac{2\supdensity(\pi)\textnormal{PL}(\pi)\ln(4/\alpha)}{N}} + \frac{\ln(4/\alpha)\supdensity(\pi,\mu)}{3N},
\end{split}
\end{equation*}
where the last inequality is from Proposition~\ref{prop:PL-upper-bound-of-V}.

From Proposition~\ref{prop:PL-and-expected-PL}, we know that with probability at least $1-\alpha/2$,
\begin{equation*}
    \left|\textnormal{PL}(\pi) - \widehat{\textnormal{PL}}(\pi)\right| \leq \frac{\textnormal{PL}(\pi)}{2} + \frac{4\ln(4/\alpha)}{3N\infdensity(\mu)}.
\end{equation*}
Applying union bound to both, we have that with probability at least $1 - \alpha$,
\ifthenelse{\boolean{arxiv-version}}
{
\begin{align*}
\left|R(\pi) - \widehat{R}_{\textnormal{IPW}}(\pi)\right|  &\leq \sqrt{\frac{2\supdensity(\pi)\textnormal{PL}(\pi)\ln(4/\alpha)}{N}} + \frac{\ln(4/\alpha)\supdensity(\pi,\mu)}{3N}\\
 &\leq \sqrt{\frac{2\supdensity(\pi)\rbr{3/2\widehat{\textnormal{PL}}(\pi) + 4\ln(4/\alpha)/(3N\infdensity(\mu)) }\ln(4/\alpha)}{N}} + \frac{\ln(4/\alpha)\supdensity(\pi,\mu)}{3N}\\
&\leq \sqrt{\frac{3\supdensity(\pi)\widehat{\textnormal{PL}}(\pi)\ln(4/\alpha)}{N}} + \sqrt{\frac{8\supdensity(\pi)}{3\infdensity(\mu)}}\frac{\ln(4/\alpha)}{N} + \frac{\ln(4/\alpha)\supdensity(\pi,\mu)}{3N},
\end{align*}
}
{
\begin{multline*}
\left|R(\pi) - \widehat{R}_{\textnormal{IPW}}(\pi)\right|  \leq \sqrt{\frac{2\supdensity(\pi)\textnormal{PL}(\pi)\ln(4/\alpha)}{N}} + \frac{\ln(4/\alpha)\supdensity(\pi,\mu)}{3N}\\
 \leq \sqrt{\frac{2\supdensity(\pi)\rbr{3/2\widehat{\textnormal{PL}}(\pi) + 4\ln(4/\alpha)/(3N\infdensity(\mu)) }\ln(4/\alpha)}{N}} + \frac{\ln(4/\alpha)\supdensity(\pi,\mu)}{3N}\\
\leq \sqrt{\frac{3\supdensity(\pi)\widehat{\textnormal{PL}}(\pi)\ln(4/\alpha)}{N}} + \sqrt{\frac{8\supdensity(\pi)}{3\infdensity(\mu)}}\frac{\ln(4/\alpha)}{N} + \frac{\ln(4/\alpha)\supdensity(\pi,\mu)}{3N},
\end{multline*}
}
where the last inequality holds since $\sqrt{B_1 + B_2} \leq \sqrt{B_1} + \sqrt{B_2}$ for any $B_1 \geq 0$ and $B_2\geq 0$. 
\end{proof}

\subsection{Proof of Lemma~\ref{lemma:IPW+PL-best-effort}}

We prove a version of the lemma with exact constants: namely, \refeq{eq:lemma:IPW+PL-best-effort} is spelled out as
\begin{equation*}
R(\pi) \leq \widehat{R}_{\textnormal{IPW}}(\pi) + \beta \widehat{\textnormal{PL}}(\pi) + \frac{3\supdensity(\Pi)\ln(4|\Pi|/\alpha)}{4\beta N} + \sqrt{\frac{8\supdensity(\Pi)}{3\infdensity(\mu)}}\frac{\ln(4|\Pi|/\alpha)}{N} + \frac{\ln(4|\Pi|/\alpha)\supdensity(\Pi,\mu)}{3N}. 
\end{equation*}

Applying the union bound to the inequality in Lemma~\ref{lemma:IPW+PL-confidence-discrete} for all policies $\pi \in \Pi$, we have that with probability at least $1 - \alpha$, for any $\pi \in \Pi$, $\beta > 0$, 

\begin{align*}
    \left|R(\pi) - \widehat{R}_{\textnormal{IPW}}(\pi)\right| &\leq \sqrt{\frac{3\supdensity(\pi)\widehat{\textnormal{PL}}(\pi)\ln(4|\Pi|/\alpha)}{N}} + \sqrt{\frac{8\supdensity(\pi)}{3\infdensity(\mu)}}\frac{\ln(4|\Pi|/\alpha)}{N} + \frac{\ln(4|\Pi|/\alpha)\supdensity(\pi,\mu)}{3N} \\
    &\leq \beta \widehat{\textnormal{PL}}(\pi) + \frac{3\supdensity(\pi)\ln(4|\Pi|/\alpha)}{4\beta N} + \sqrt{\frac{8\supdensity(\pi)}{3\infdensity(\mu)}}\frac{\ln(4|\Pi|/\alpha)}{N} + \frac{\ln(4|\Pi|/\alpha)\supdensity(\pi,\mu)}{3N}\\
    &\leq \beta \widehat{\textnormal{PL}}(\pi) + \frac{3\supdensity(\Pi)\ln(4|\Pi|/\alpha)}{4\beta N} + \sqrt{\frac{8\supdensity(\Pi)}{3\infdensity(\mu)}}\frac{\ln(4|\Pi|/\alpha)}{N} + \frac{\ln(4|\Pi|/\alpha)\supdensity(\Pi,\mu)}{3N},
\end{align*}

where the second inequality is derived by applying the AM-GM inequality, and the last inequality is by the definition of $\supdensity(\pi,\mu)$, $\supdensity(\Pi,\mu)$, $\supdensity(\pi)$, and $\supdensity(\Pi)$. This concludes the proof.

\subsection{Proof of Proposition~\ref{prop:IPW+PL-compatible-classification-oracle-discrete}}

Recall that the proposition states that the optimization in \refeq{eq:objective-pl-discrete}
can be solved by calling any CSC oracle for policy class $\Pi$ once,
with modified loss vectors $a \mapsto \ell_i(a_i)/\mu(a_i \mid x_i) \cdot \one\{a=a_i\} + \beta/\mu(a\given x_i) $.

The objective in Eq.~\eqref{eq:objective-pl-discrete} can be re-written as
\begin{align*}
    \widehat{R}_{\textnormal{IPW}}(\pi) + \beta\widehat{\textnormal{PL}}(\pi) &= \frac{1}{N}\sum_{i=1}^N \frac{\pi(a_i\given x_i)}{\mu(a_i\given x_i)}\ell_i(a_i) + \beta \frac{1}{N}\sum_{i=1}^N\sum_{a\in\Acal}\frac{\pi(a\given x_i)}{\mu(a\given x_i)}\\
    &= \frac{1}{N}\sum_{i=1}^N\sum_{a\in\Acal}\pi(a\given x_i)\rbr{\one\cbr{a=a_i}\ell_i(a_i)/\mu(a_i\given x_i) + \beta/\mu(a\given x_i)},
\end{align*}
which concludes the proof.

\subsection{Proof of Theorem~\ref{theo:IPW+PL-oracle-inequality-discrete}}

We prove Theorem~\ref{theo:IPW+PL-oracle-inequality-discrete} with exact constants: \asedit{the guarantees}
in the theorem statement are spelled out as, resp.,
\begin{align}
R(\widehat{\pi}_{\textnormal{IPW+PL},\beta})
   &\leq \min_{\pi \in \Pi}\cbr{R(\pi) + 2\beta \widehat{\PL}(\pi)
   + \frac{
       \rbr{\frac{3}{2}\supdensity\rbr{\Pi}/\beta +8\Delta(\Pi,\mu)}
       \cdot\ln\rbr{4|\Pi| / \alpha}}{N}}.\\
R(\widehat{\pi}_{\textnormal{IPW+PL}, \beta^\star})
   &\leq \min_{\pi \in \Pi}\cbr{R(\pi)
       + \sqrt{\frac{18\supdensity(\Pi)\cdot \PL(\pi)\cdot\ln(4|\Pi|/\alpha)}{N}} +
       \frac{ 12\Delta(\Pi,\mu)\cdot \ln(4|\Pi|/\alpha)}{N} }.
   \label{eq:IPW+PL-oracle-inequality-discrete-beta-star}
\end{align}

From the proof of Lemma ~\ref{lemma:IPW+PL-best-effort}, we know that with probability at least $1 - \alpha$, for any $\beta > 0$ and $\pi \in \Pi$,

\ifthenelse{\boolean{arxiv-version}}
{
\begin{align*}
&R(\widehat{\pi}_{\textnormal{IPW+PL},\beta})  \\
&\leq \widehat{R}_{\textnormal{IPW}}(\widehat{\pi}_{\textnormal{IPW+PL},\beta})  +  \beta \widehat{\textnormal{PL}}(\widehat{\pi}_{\textnormal{IPW+PL},\beta}) + \frac{3\supdensity(\Pi)\ln(4|\Pi|/\alpha)}{4\beta N} + \sqrt{\frac{8\supdensity(\Pi)}{3\infdensity(\mu)}}\frac{\ln(4|\Pi|/\alpha)}{N} + \frac{\ln(4|\Pi|/\alpha)\supdensity(\Pi,\mu)}{3N}\\
&\leq \widehat{R}_{\textnormal{IPW}}(\pi) +  \beta \widehat{\textnormal{PL}}(\pi) + \frac{3\supdensity(\Pi)\ln(4|\Pi|/\alpha)}{4\beta N} + \sqrt{\frac{8\supdensity(\Pi)}{3\infdensity(\mu)}}\frac{\ln(4|\Pi|/\alpha)}{N} + \frac{\ln(4|\Pi|/\alpha)\supdensity(\Pi,\mu)}{3N}\\
&\leq R(\pi) + 2\beta\widehat{\textnormal{PL}}(\pi) + \frac{3\supdensity(\Pi)\ln(4|\Pi|/\alpha)}{2\beta N} + \sqrt{\frac{32\supdensity(\Pi)}{3\infdensity(\mu)}}\frac{\ln(4|\Pi|/\alpha)}{N} + \frac{2\ln(4|\Pi|/\alpha)\supdensity(\Pi,\mu)}{3N},
\end{align*}
}
{
\begin{multline*}
R(\widehat{\pi}_{\textnormal{IPW+PL},\beta})  \leq \widehat{R}_{\textnormal{IPW}}(\widehat{\pi}_{\textnormal{IPW+PL},\beta})  +  \beta \widehat{\textnormal{PL}}(\widehat{\pi}_{\textnormal{IPW+PL},\beta}) \\+ \frac{3\supdensity(\Pi)\ln(4|\Pi|/\alpha)}{4\beta N} + \sqrt{\frac{8\supdensity(\Pi)}{3\infdensity(\mu)}}\frac{\ln(4|\Pi|/\alpha)}{N} + \frac{\ln(4|\Pi|/\alpha)\supdensity(\Pi,\mu)}{3N}\\
\leq \widehat{R}_{\textnormal{IPW}}(\pi) +  \beta \widehat{\textnormal{PL}}(\pi) + \frac{3\supdensity(\Pi)\ln(4|\Pi|/\alpha)}{4\beta N} + \sqrt{\frac{8\supdensity(\Pi)}{3\infdensity(\mu)}}\frac{\ln(4|\Pi|/\alpha)}{N} + \frac{\ln(4|\Pi|/\alpha)\supdensity(\Pi,\mu)}{3N}\\
\leq R(\pi) + 2\beta\widehat{\textnormal{PL}}(\pi) + \frac{3\supdensity(\Pi)\ln(4|\Pi|/\alpha)}{2\beta N} + \sqrt{\frac{32\supdensity(\Pi)}{3\infdensity(\mu)}}\frac{\ln(4|\Pi|/\alpha)}{N} + \frac{2\ln(4|\Pi|/\alpha)\supdensity(\Pi,\mu)}{3N},
\end{multline*}
}
where the first and third inequalities come from the proof of Lemma~\ref{lemma:IPW+PL-best-effort}, and the second inequality is by the definition of $\widehat{\pi}_{\textnormal{IPW+PL}, \beta}$.
This concludes the proof for the first part of the theorem.

Since the above inequality holds for every policy $\pi\in\Pi$ and $\beta>0$, let $\beta_{\pi} \coloneqq \sqrt{\frac{3\supdensity(\Pi)\ln(4|\Pi|/\alpha)}{4N\widehat{\textnormal{PL}}(\pi)}}$, we know that with probability at least $1-\alpha$, for any policy $\pi \in \Pi$,
\begin{multline*}
R(\widehat{\pi}_{\textnormal{IPW+PL},\beta_\pi})\leq R(\pi) + 2\sqrt{\frac{3\supdensity(\Pi)\ln(4|\Pi|/\alpha)\widehat{\textnormal{PL}}(\pi)}{N}} + \sqrt{\frac{32\supdensity(\Pi)}{3\infdensity(\mu)}}\frac{\ln(4|\Pi|/\alpha)}{N} + \frac{2\ln(4|\Pi|/\alpha)\supdensity(\Pi,\mu)}{3N}.
\end{multline*}

So with probability at least $1-\alpha$, there exists $\beta^\star \in \argmin_{\beta_\pi: \pi \in \Pi}R(\widehat{\pi}_{\textnormal{IPW+PL},\beta_\pi})$ such that
\begin{multline*}
R(\widehat{\pi}_{\textnormal{IPW+PL},\beta^\star})\leq \min_{\pi\in\Pi} R\rbr{\widehat{\pi}_{\textnormal{IPW+PL},\beta_\pi}} \\
\leq \min_{\pi\in\Pi}\cbr{R(\pi) + 2\sqrt{\frac{3\supdensity(\Pi)\ln(4|\Pi|/\alpha)\widehat{\textnormal{PL}}(\pi)}{N}} + \sqrt{\frac{32\supdensity(\Pi)}{3\infdensity(\mu)}}\frac{\ln(4|\Pi|/\alpha)}{N} + \frac{2\ln(4|\Pi|/\alpha)\supdensity(\Pi,\mu)}{3N}}.
\end{multline*}
And we know from Proposition~\ref{prop:PL-and-expected-PL} that
\ifthenelse{\boolean{arxiv-version}}
{
\begin{align*}
\sqrt{\frac{3\supdensity(\Pi)\ln(4|\Pi|/\alpha)\widehat{\textnormal{PL}}(\pi)}{N}}
&\leq\sqrt{\frac{3\supdensity(\Pi)\ln(4|\Pi|/\alpha)\rbr{3/2\textnormal{PL}(\pi) + 4\ln(4|\Pi|/\alpha)/(3N\infdensity(\mu))}}{N}} \\
&\leq \sqrt{\frac{9/2\supdensity(\Pi)\ln(4|\Pi|/\alpha)\textnormal{PL}(\pi)}{N}} + \sqrt{\frac{4\supdensity(\Pi)\ln^2(4|\Pi|/\alpha)}{N^2\infdensity(\mu)}},
\end{align*}
}
{
\begin{multline*}
\sqrt{\frac{3\supdensity(\Pi)\ln(4|\Pi|/\alpha)\widehat{\textnormal{PL}}(\pi)}{N}} 
\leq\sqrt{\frac{3\supdensity(\Pi)\ln(4|\Pi|/\alpha)\rbr{3/2\textnormal{PL}(\pi) + 4\ln(4|\Pi|/\alpha)/(3N\infdensity(\mu))}}{N}} \\
\leq \sqrt{\frac{9/2\supdensity(\Pi)\ln(4|\Pi|/\alpha)\textnormal{PL}(\pi)}{N}} + \sqrt{\frac{4\supdensity(\Pi)\ln^2(4|\Pi|/\alpha)}{N^2\infdensity(\mu)}},
\end{multline*}
}
where the second inequality again uses the fact that $\sqrt{B_1 + B_2} \leq \sqrt{B_1} + \sqrt{B_2}$ for any $B_1\geq 0$ and $B_2\geq 0$.
So we can get
\ifthenelse{\boolean{arxiv-version}}
{
\begin{multline*}
R(\widehat{\pi}_{\textnormal{IPW+PL},\beta^\star})\\
\leq\min_{\pi\in\Pi}\cbr{R(\pi) + \sqrt{\frac{18\supdensity(\Pi)\ln(4|\Pi|/\alpha)\textnormal{PL}(\pi)}{N}} + 6\sqrt{\frac{\supdensity(\Pi)}{\infdensity(\mu)}}\frac{\ln(4|\Pi|/\alpha)}{N} + \frac{2\ln(4|\Pi|/\alpha)\supdensity(\Pi,\mu)}{3N}},
\end{multline*}
}
{
\begin{multline*}
R(\widehat{\pi}_{\textnormal{IPW+PL},\beta^\star})\\
\leq\min_{\pi\in\Pi}\cbr{R(\pi) + \sqrt{\frac{18\supdensity(\Pi)\ln(4|\Pi|/\alpha)\textnormal{PL}(\pi)}{N}} + 6\sqrt{\frac{\supdensity(\Pi)}{\infdensity(\mu)}}\frac{\ln(4|\Pi|/\alpha)}{N} + \frac{2\ln(4|\Pi|/\alpha)\supdensity(\Pi,\mu)}{3N}},
\end{multline*}
}
which concludes the proof.


\subsection{Proof of Proposition~\ref{prop:IPW+PL-compatible-classification-oracle-continuous}}

We prove the formal version of Proposition~\ref{prop:IPW+PL-compatible-classification-oracle-continuous}, stated as follows.

\begin{proposition}
Fix $K,H\in\N$. Consider the density-based policy class $\Pi = \Pi_{K, H}$ as constructed above. Then the objective in \refeq{eq:objective-pl-discrete} can be optimized via a single call to a CSC oracle for the mass-based policy class $\widetilde{\Pi}_{K}$, with
loss function:
\begin{align}
    \widetilde{a} \mapsto \frac{\ell_i(a_i)}{\EH\rbr{\widetilde{a}} \mu (a_i \given x_i)} \one\cbr{\widetilde{a} \in A_{K,H}(a_i)} + \frac{\beta}{\EH\rbr{\widetilde{a}}}\int_{\max\rbr{0, \widetilde{a} + H/2}}^{\min\rbr{1, \widetilde{a} - H / 2}}\frac{1}{ \mu(a \given x_i)} \dif{a}, \label{eq:continuous_modified_losses}
\end{align}
where $\tilde{a} \in \widetilde{A}_K$, the latter being the surrogate-action-set, $\EH(\widetilde{a}) \coloneqq \min\rbr{1, \widetilde{a} + H / 2} - \max\rbr{0, \widetilde{a} - H / 2}$ is the effective bandwidth, and $A_{K,H}(a) \coloneqq \cbr{\widetilde{a} \in \widetilde{\Acal}_K : \widetilde{a}\in \sbr{a - h / 2, a + h / 2}}$ is the surrogate-action-set identity function.
\end{proposition}

With the above definitions, the objective of in Eq.~\eqref{eq:objective-pl-discrete} for a policy $\pi \in \Pi_{K,H}$ is
\begin{equation*}
\begin{split}
    &\widehat{R}_{\textnormal{IPW}}(\pi) + \beta \textnormal{PL}(\pi) \\
    &= \frac{1}{N} \sum_{i=1}^N \sbr{\frac{\pi(a_i \given x_i)}{\mu(a_i \given x_i)}\ell_i(a_i) + \beta\int_{0}^{1} \frac{\pi(a \given x_i)}{\mu(a\given x_i)} \dif{a}} \\
    &= \frac{1}{N} \sum_{i=1}^N \sbr{\frac{\ell_i(a_i)}{\mu(a_i \given x_i)}\sum_{\widetilde{a} \in A_{K, H}(a_i)}\frac{\widetilde{\pi}\rbr{\widetilde{a} \given x_i}}{\EH\rbr{\widetilde{a}}}  + \beta \int_{0}^{1} \frac{\sum_{\widetilde{a} \in A_{K, H}(a_i)}\widetilde{\pi}\rbr{\widetilde{a} \given x_i}}{\EH(\widetilde{a}) \mu\rbr{a \given x_i}}\dif{a}} \\
    &= \frac{1}{N} \sum_{i=1}^N \sbr{\sum_{\widetilde{a} \in A_{K, H}(a_i)} \widetilde{\pi}(\widetilde{a} \given x_i)\frac{ \ell_i(a_i)}{\EH\rbr{\widetilde{a}} \mu\rbr{a_i \given x_i}} + \beta \sum_{\widetilde{a} \in \widetilde{\Acal}_K} \widetilde{\pi}\rbr{\widetilde{a} \given x_i} \int_{\max\rbr{0, \widetilde{a} + H/2}}^{\min\rbr{1, \widetilde{a} - H / 2}} \frac{1}{\EH\rbr{\widetilde{a}} \mu(a \given x_i)}\dif{a}} \\
    &= \frac{1}{N} \sum_{i=1}^N \sbr{\sum_{\widetilde{a} \in \widetilde{\Acal}_K} \widetilde{\pi}\rbr{\widetilde{a} \given x_i} \rbr{\frac{\ell_i(a_i)}{\EH\rbr{\widetilde{a}} \mu (a_i \given x_i)} \one\cbr{\widetilde{a} \in A_{K,H}(a_i)} + \frac{\beta}{\EH\rbr{\widetilde{a}}}\int_{\max\rbr{0, \widetilde{a} + H/2}}^{\min\rbr{1, \widetilde{a} - H / 2}}\frac{1}{ \mu(a \given x_i)} \dif{a}}}.
\end{split}
\end{equation*}
So minimizing the objective in Eq~\eqref{eq:objective-pl-discrete} over $\Pi$ is equivalent to minimizing the CSC objective over $\widetilde{\Pi}_K$ over the modified loss vectors defined in Eq.~\eqref{eq:continuous_modified_losses}.

\subsection{Proof of Corollary~\ref{coro:IPW+PL-oracle-inequality-for-smoothed-policies}}

We prove Corollary~\ref{coro:IPW+PL-oracle-inequality-for-smoothed-policies} with exact constants: \asedit{the guarantees}
are spelled out, resp., as
\begin{align*}
R(\widehat{\pi}_{\textnormal{IPW+PL},\beta})
   &\leq \min_{\pi \in \Pi}\cbr{R(\pi) + 2\beta \widehat{\PL}(\pi)
   + \frac{
       \rbr{3/\beta +16/\infdensity(\mu)}
       \cdot\ln\rbr{|4\Pi| / \alpha}}{NH}}. \\
R(\widehat{\pi}_{\textnormal{IPW+PL}, \beta^\star})
   &\leq \min_{\pi \in \Pi}\cbr{R(\pi)
       + 6\sqrt{\frac{\PL(\pi)\cdot\ln(4|\Pi|/\alpha)}{NH}} +
       \frac{ 24\ln(4|\Pi|/\alpha)}{NH\infdensity(\mu)} }.
\end{align*}

This follows by setting $\supdensity(\Pi) = 2/H$ and $\Delta(\Pi,\mu) = 2/(H\infdensity(\mu))$ in Theorem~\ref{theo:IPW+PL-oracle-inequality-discrete}.

\subsection{How to set hyper-parameters in the continuous-action setting}
\label{sec:hyper-parameter-continuous-action}

While our theory in Section~\ref{sec:cont-PL} is presented for fixed hyper-parameters $K,H$, this appendix provides suggestions for how to choose them in practice. To this end,
we analyze how $K$ and $H$ affect the risk of the policies in $\Pi_{K,H}$ and the generalization performance of $\widehat{\pi}_{\textnormal{IPW+PL},\beta^\star}$,  similar to ~\citep{krishnamurthy2020contextual}.

\xhdr{How to set $K$ for a fixed $H$?}
Let us start with some set of density-based policies, denoted $\Pi_{\infty}$. Note that $\supdensity(\Pi_{\infty})$ can be huge and the generalization guarantee in  Theorem~\ref{theo:IPW+PL-oracle-inequality-discrete} becomes vacuous for general density-based policies. So we are not providing excess risk guarantees for this policy class. Instead, we consider a policy class smoothed from some mass-based policy class. For an integer $K > 0$, let $\widetilde{\Pi}_{K} = \cbr{\textnormal{Discretize}_K(\pi): \pi \in \Pi_{\infty}}$ be the set of mass-based policies discretized from $\Pi_{\infty}$, where $\widetilde{\pi} = \textnormal{Discretize}_K(\pi)$ is a mass-based policy such that $\widetilde{\pi}(\widetilde{a} \given x) = \int_{\widetilde{a} - 1/(2K)}^{\widetilde{a} + 1/ (2K)}\pi(a \given x)\dif{a}$ for any $x \in \Xcal$ and $\widetilde{a} \in \widetilde{\Acal}_K$. Let $\Pi_{K, H} = \cbr{\textnormal{Smooth}_H \rbr{\widetilde{\pi}} : \widetilde{\pi} \in \widetilde{\Pi}_K}$. We have analyzed the generalization performance of $\widehat{\pi}_{\textnormal{IPW+PL},\beta^\star}$ for $\Pi_{K,H}$ for a particular $K$. We now want to see how different $K$ might affect the policy risks in $\Pi_{K,H}$. In particular, we consider $\Pi_{\infty, H} = \cbr{\textnormal{Smooth}_H(\pi): \pi \in \Pi_{\infty}}$ be the set of density-based policies smoothed from $\Pi_{\infty}$, which represents the policy class when $K$ approaches infinity.
We know that, for any $\pi \in \Pi_{\infty}$,

\begin{equation*}
\left|R\rbr{\textnormal{Smooth}_H\rbr{\textnormal{Discretize}_K\rbr{\pi}}} - R\rbr{\textnormal{Smooth}_H\rbr{\pi}}\right| \leq \min\rbr{1, \frac{1}{HK}}.
\end{equation*}

To analyze how to set $K$, we consider the excess risk of $\widehat{\pi}_{\textnormal{IPW+PL}, \beta^\star}$ in Eq~\eqref{eq:IPW+PL-oracle-inequality-discrete-beta-star}
\begin{equation*}
\begin{split}
R(\widehat{\pi}_{\textnormal{IPW+PL}, \beta^\star})
    &\lesssim \min_{\pi \in \Pi_{K,H}}\cbr{R(\pi)
        + \sqrt{\frac{\PL(\pi)\cdot\ln(|\Pi_{K,H}|/\alpha)}{NH}} +
        \frac{ \ln(|\Pi_{K,H}|/\alpha)}{NH\infdensity(\mu)} }\\
        &\leq \min_{\pi \in \Pi_{K,H}}\cbr{R(\pi)
        + \sqrt{\frac{\ln(|\Pi_{K,H}|/\alpha)}{NH\infdensity(\mu)}} +
        \frac{ \ln(|\Pi_{K,H}|/\alpha)}{NH\infdensity(\mu)} }\\
        &\leq \min_{\pi \in \Pi_{\infty,H}}\cbr{R(\pi)
        + \sqrt{\frac{\ln(|\Pi_{K,H}|/\alpha)}{NH\infdensity(\mu)}} +
        \frac{ \ln(|\Pi_{K,H}|/\alpha)}{NH\infdensity(\mu)} + \frac{1}{HK} },
\end{split}
\end{equation*}
where the second inequality holds since $\PL(\pi) \leq 1/\infdensity(\mu)$. Now, if $|\Pi_{K,H}|$ scales exponentially with $K$, then we should set $K$ on the order of $\rbr{\frac{N\infdensity(\mu)}{H\ln(1/\alpha)}}^{1/3}$ to optimize the second and fourth terms. If we assume $|\Pi_{K,H}|$ does not depend on $K$, then we should set $K$ to be sufficiently large so that the fourth term is lower order.

\xhdr{How to choose H?
For the sake of intuition, consider a fixed $K$, and}
let $\Pi_{K, H}$ and $\Pi_{K, H+ \gamma}$ be density-based policy classes smoothed from the same mass-based policy class $\widetilde{\Pi}_{K}$ with bandwidth $H$ and $H+\gamma$ respectively. For any mass-based policy $\widetilde{\pi} \in \Pi_{K}$, we have
\begin{equation*}
    \left|R(\textnormal{Smooth}_H\rbr{\widetilde{\pi}}) - R(\textnormal{Smooth}_{H + \gamma}\rbr{\widetilde{\pi}})\right| \leq \min\rbr{1, \frac{2\gamma}{H}}.
\end{equation*}

This suggests that we might want to search over a space of $H$ such that $1/H$ is equally spaced.

\newpage
\section{Detailed empirical evaluation}
\label{sec:additional-exp}

\subsection{Experiments with discrete actions}
\label{sec:additional-exp-discr}
\xhdr{Experimental setup. }
Following prior works~\citep{beygelzimer2009offset,dudik2014doubly,wang2017optimal,su2019cab,su2020doubly}, we empirically examine the performance of policy optimization with the PL regularizer on simulated bandit instances from full-information classification datasets. This allows us to evaluate the performance of different policy optimization methods with ground-truth policy values and to control the setting of the problem so that we can test the robustness of PL in a variety of experimental scenarios.

\xhdr{Datasets. } We conduct experiments on 4 multi-class classification datasets with real-valued features and 1 million examples from OpenML~\citep{OpenML2013}, see Table~\ref{tb:discrete-data-detail} for detailed statistics.

\begin{table}[h]
\centering
\caption{Discrete action datasets}
\label{tb:discrete-data-detail}
\begin{tabular}{lllll}
\toprule
Dataset    & Letter & PenDigits & SatImage &  JPVowel    \\
\midrule
OpenML ID   &247        & 261       & 1183      &       1214 \\
\# Data     & 1,000,000  & 1,000,000      & 1,000,000     & 1,000,000   \\
\# Features & 16     & 16        & 36       & 14      \\
\# Classes  & 26     & 10        & 6        & 9  \\
    \bottomrule
\end{tabular}
\end{table}

For each dataset, we hold out $1\%$ of the data for training a logging policy and $30\%$ of the data for testing. The rest of the data are used for simulating the bandit feedback.

We test the performance of different policy optimization algorithms under various settings, which are summarized in Table~\ref{tb:discrete-setting-options}. First, we vary the data size by randomly selecting $1\%$, $10\%$, or $100\%$ of the $69\%$ of the data to simulate bandit feedback data. To vary the number of actions and the type of cost, we follow~\citet{foster2018practical} to transform the original multi-class classification dataset to a CSC dataset that can have either real-valued or binary-valued cost and arbitrary number of classes.
For each dataset with $K$ classes, we construct a cost matrix $C \in \RR^{K_{\textnormal{cs}} \times K}$ where $K_{\textnormal{cs}} \in \{K, 5K\}$. Each entry $C(a, a^\star)$ is the cost of classifying an example with true class label $a^\star$ as class $a$ in the set of $K_{\textnormal{cs}}$ actions. The entries $C(iK + a, a)$ for any integer $i \geq 0$ such that $iK + a \leq K_{\textnormal{cs}}$ and $a \in \{1,2, \dots, K\}$ are set to be $0$. For the binary-valued-cost experiments, the rest of the entries are set to be $1$. For the real-valued-cost experiments, the rest of the entries are generated uniformly at random from the interval $[0, 1]$.
To simulate bandit feedback, for each example $(x, c)$ in the CSC dataset, we take an action $a$ following a logging policy $\mu$ and observe the binary loss $\ell(a) \sim \textnormal{Bernoulli}(c(a))$ where $c(a)$ is the cost of action $a$ for $x$.

\xhdr{Logging policies.} We consider three different types of logging policies. For each dataset, we learn one ``good'' and one ``bad'' deterministic multi-class classification models with the $1\%$ held out data using the linear-regression CSC oracle described below to predict the class with the smallest and largest cost, respectively. Then we construct three stochastic logging policies $\mu_{\textnormal{good}, \epsilon=0.1}$, $\mu_{\textnormal{good}, \epsilon=0.01}$, and $\mu_{\textnormal{bad}, \epsilon=0.1}$ by combining the deterministic policies with the uniform-random policy where $\epsilon = 0.1$ and $\epsilon = 0.01$ are the probabilities of using the uniform-random policy.

\xhdr{OPO methods.}
    We compare the performance of using PL regularizer with that of no pessimism (None) and EB under different estimators and oracles, which are summarized in Table~\ref{tb:discrete-algorithm-options}. For all the regularizers, we consider two types of estimators: IPW and the doubly robust estimator (DR)~\citep{dudik2014doubly}. For PL and None, we run experiments on two types of CSC oracles. The first one is policy gradient (PG) with a softmax-linear parametrization, which selects actions proportional to $\exp(\inner{\theta}{\phi(x,a)})$ where $\theta$ is the policy parameter and $\phi(x,a)$ are the features.
    The policy parameters are fit by directly optimizing the CSC objective with $\ell_2$ regularization. The second CSC oracle is based on linear regression with $\ell_2$ regularization and we denote it as LR. The policy is derived by regressing the costs onto the features using (regularized) least squares regression and then taking the action with the minimum predicted cost.
    As we have discussed, EB is not compatible with CSC oracles in general. We follow prior works and parameterize the policy identically to the PG-based CSC oracle and directly optimize the EB objective.

\xhdr{Policy selection.} We split the bandit feedback data and use $50\%$ of the data for policy optimization and $50\%$ of the data for policy selection. For policy selection, we adopt the strategy using the EB bound in Eq.~\eqref{eq:oracle-inequality-EB} with $\alpha=0.1$, since it is tighter than the PL bound as shown in Eq.~\eqref{eq:EB-objective}.  We run each experiment $50$ times and report the mean and standard error of the results.

\xhdr{Hyper-parameter details. } We shift the loss to the range $[-1,0]$, since this improves the performance on all the methods in our exploratory experiments, which is also consistent with findings in many prior works~\citep{swaminathan2015batch,joachims2018deep,bietti2021contextual}.\footnote{We note that our theoretical results continue to apply with the loss range. In particular, the proofs of Proposition~\ref{prop:PL-upper-bound-of-V},~\ref{prop:PL-and-expected-PL}, Lemma~\ref{lemma:IPW+PL-confidence-discrete} and Theorem~\ref{theo:IPW+PL-oracle-inequality-discrete} still hold when $-1 \leq \ell(a)\leq 0$.}
For policy optimization with the DR estimator, we further split the data for policy optimization into $10\%$ for training the cost regression model and $90\%$ for policy optimization. The cost model of DR is trained using linear regression with $\ell_2$ regularization on the $10\%$ of bandit feedback data for policy optimization. For both PL and EB, we grid search $\beta$ in $\{0, 1e-3, 3e-3, 1e-2, 3e-2, 1e-1, 3e-1, 1\}$. For CSC oracles, we set the weight decay to be $1e-6$ and grid search the learning rate in $\{1e-3, 1e-2, 1e-1, 1, 10\}$. And we use stochastic gradient descent with batch size $100$ and epoch $1$ to optimize each model.
For the PG oracle for EB, we optimize the model with LBFGS for 10 steps since \citet{swaminathan2015batch} found that LBFGS performs better than gradient descent in their empirical evaluation. We note that this optimizer is around $20$ times slower than the CSC oracles optimized via batch stochastic gradient descent. We use the same weight decay and grid search the same learning rates as in CSC oracles. We implement all the oracles in PyTorch~\citep{paszke2019pytorch}. And the experiments are run on a shared cluster with different types of CPUs and thousands of CPU cores.

\begin{figure}
\centering
\begin{minipage}{.48\textwidth}
  \centering
  \includegraphics[width=.98\linewidth]{plots/d_eps0.1_improvement.pdf}
  \captionof{figure}{
  Relative improvement (\RelImp, see \refeq{eq:RelImp}) for PL against the baseline with no pessimism, averaged over all 50 runs (mean $\pm$ $2$ standard errors). 
   \vspace{2mm}\newline
  Shown for a particular (dataset, environment) pair and the best-performing (CSC oracle, risk estimator) pair. Each bar corresponds to a (dataset, data-size) pair.
  \vspace{2mm}\newline
24 environments, see Table~\ref{tb:discrete-setting-options}; 
4 datasets, see Table~\ref{tb:discrete-data-detail}.
\vspace{2mm}\newline
Environment: real-valued cost, $\mu_{\textnormal{good},\epsilon=0.1}$, and \#actions = \#classes.}
  \label{fig:d_eps0.1_improvement_appendix}
\end{minipage}
\hfill
\begin{minipage}{.48\textwidth}
  \centering
  \includegraphics[width=.98\linewidth]{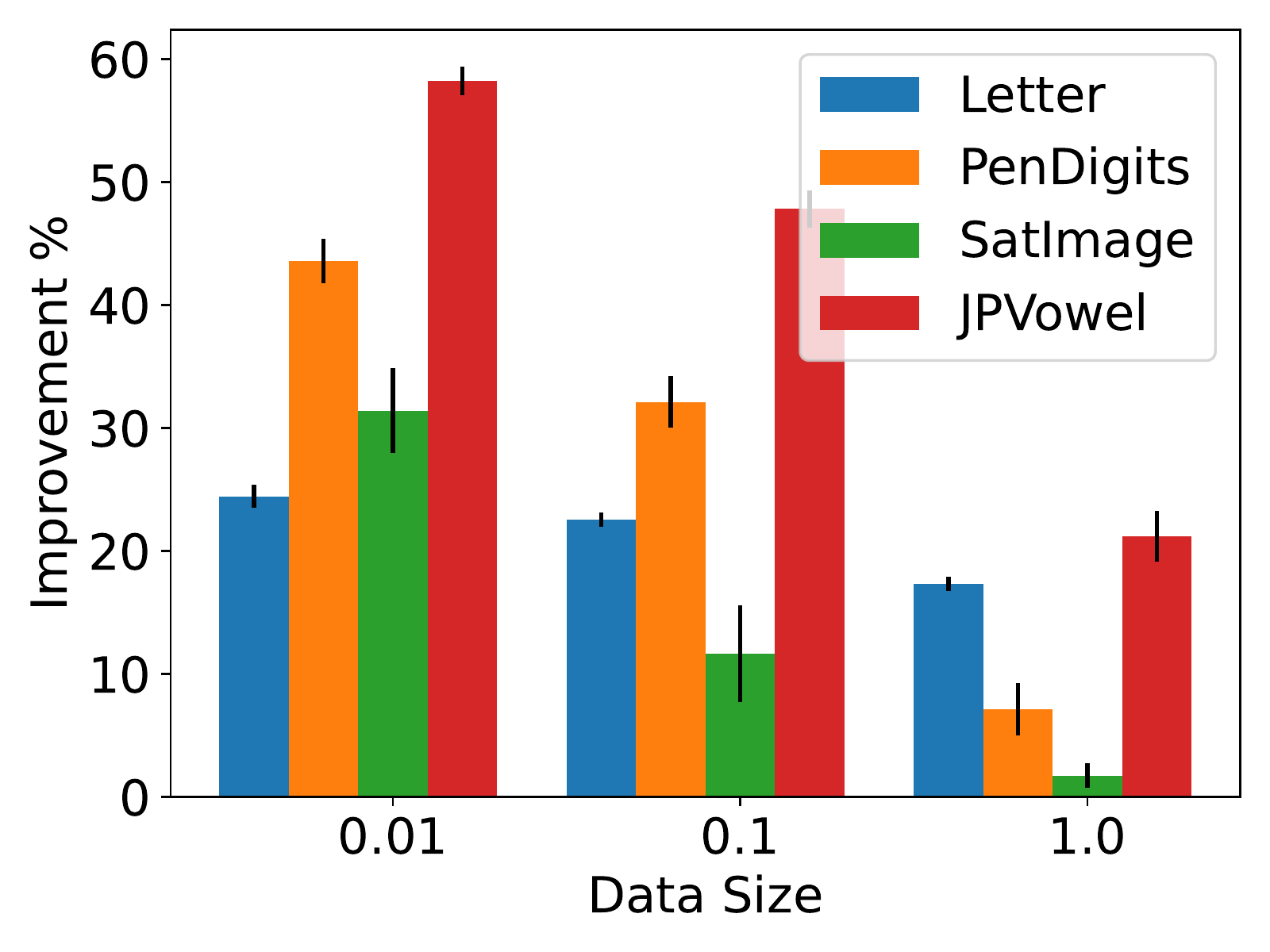}
\captionof{figure}{Same semantics as in Figure~\ref{fig:d_eps0.1_improvement_appendix}. \\
Environment: real-valued cost, $\mu_{\textnormal{good},\epsilon=0.01}$, and \# actions = \# classes. }
  \label{fig:d_eps0.01_improvement}
\end{minipage}
\end{figure}

\begin{figure}
\begin{minipage}{.48\textwidth}
  \centering
  \includegraphics[width=.98\linewidth]{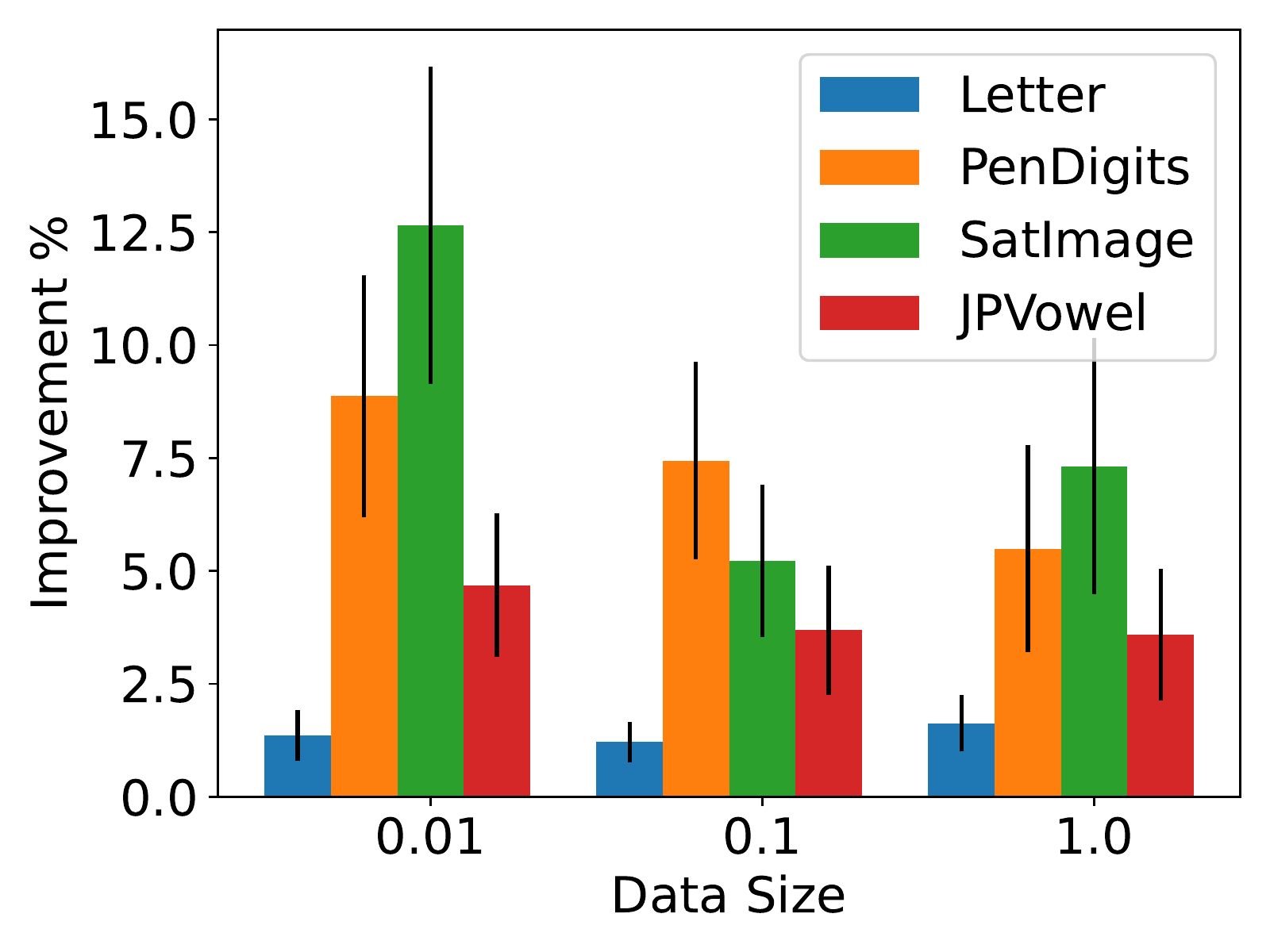}
  \captionof{figure}{Same semantics as in Figure~\ref{fig:d_eps0.1_improvement_appendix}. \\
Environment: real-valued cost, $\mu_{\textnormal{bad},\epsilon=0.1}$, and \# actions = \# classes. }
  \label{fig:d_bad_improvement}
\end{minipage}
\hfill
\begin{minipage}{.48\textwidth}
  \centering
  \includegraphics[width=.98\linewidth]{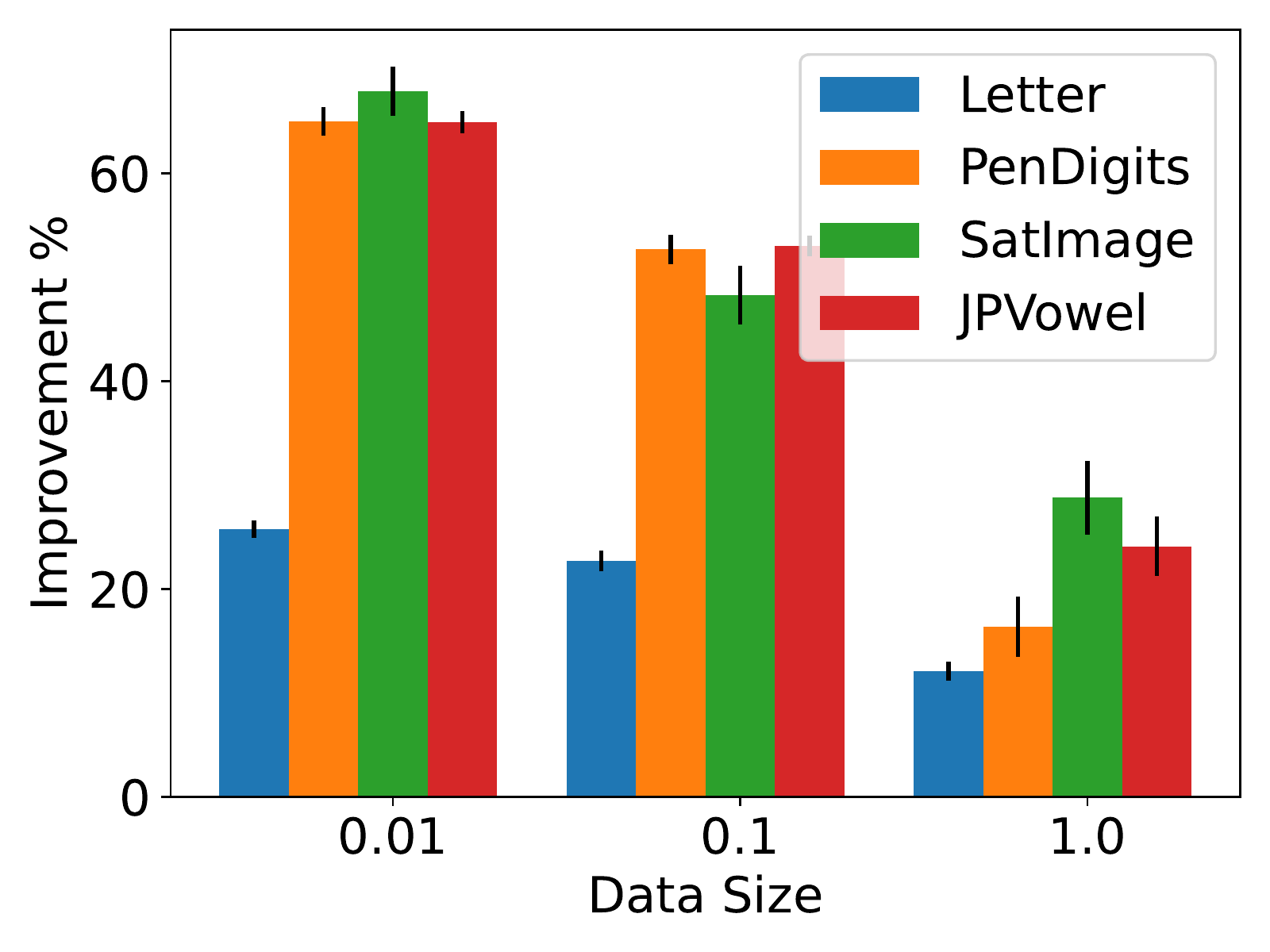}
  \captionof{figure}{Same semantics as in Figure~\ref{fig:d_eps0.1_improvement_appendix}. \\
Environment: real-valued cost, $\mu_{\textnormal{good},\epsilon=0.1}$, and \# actions = 5 $\times$ \# classes. }
  \label{fig:d_large_action_improvement}
\end{minipage}
\end{figure}

\begin{figure}
\begin{minipage}{.48\textwidth}
  \centering
  \includegraphics[width=.98\linewidth]{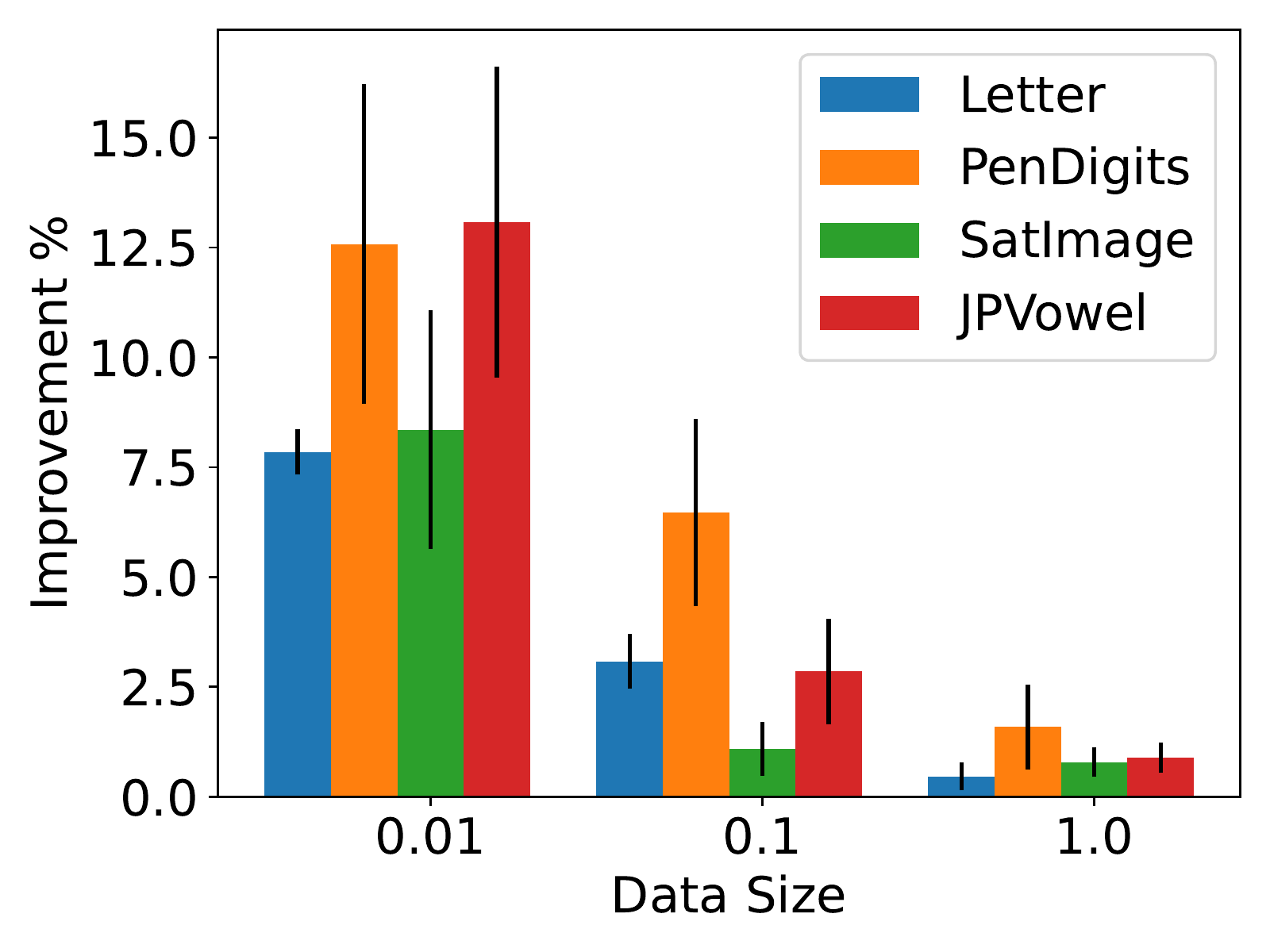}
    \captionof{figure}{Same semantics as in Figure~\ref{fig:d_eps0.1_improvement_appendix}. \\
Environment: real-valued cost, $\mu_{\textnormal{good},\epsilon=0.1}$, and \# actions = \# classes. }
  \label{fig:d_no_noise_eps0.1_improvement}
\end{minipage}
\hfill
\begin{minipage}{.48\textwidth}
  \centering
  \includegraphics[width=.98\linewidth]{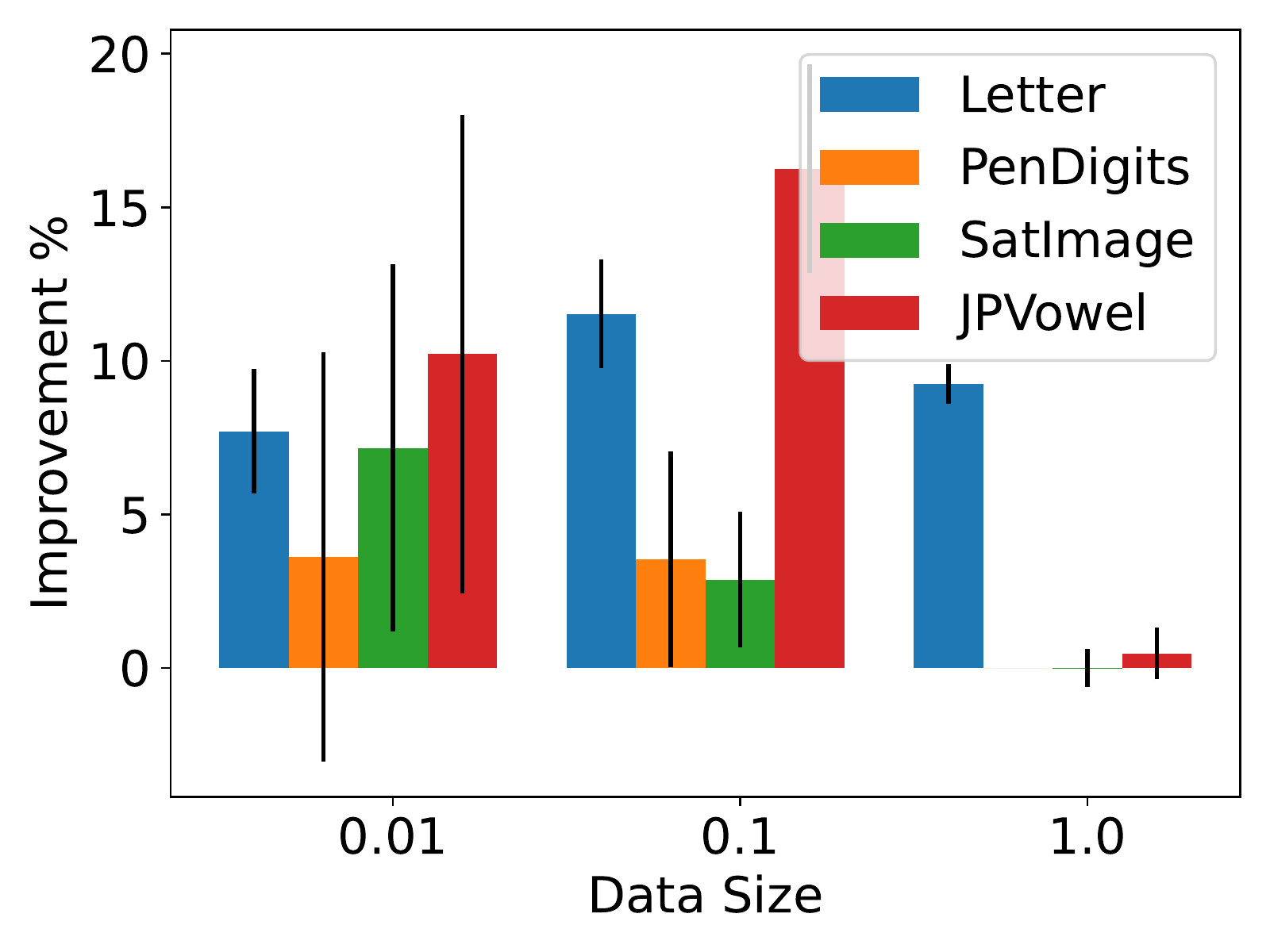}
\captionof{figure}{Same semantics as in Figure~\ref{fig:d_eps0.1_improvement_appendix}. \\
Environment: binary-valued cost, $\mu_{\textnormal{good},\epsilon=0.01}$, and \# actions = \# classes. }
  \label{fig:d_no_noise_eps0.01_improvement}
\end{minipage}
\end{figure}

\begin{figure}
\centering
\begin{minipage}{.48\textwidth}
  \centering
  \includegraphics[width=.98\linewidth]{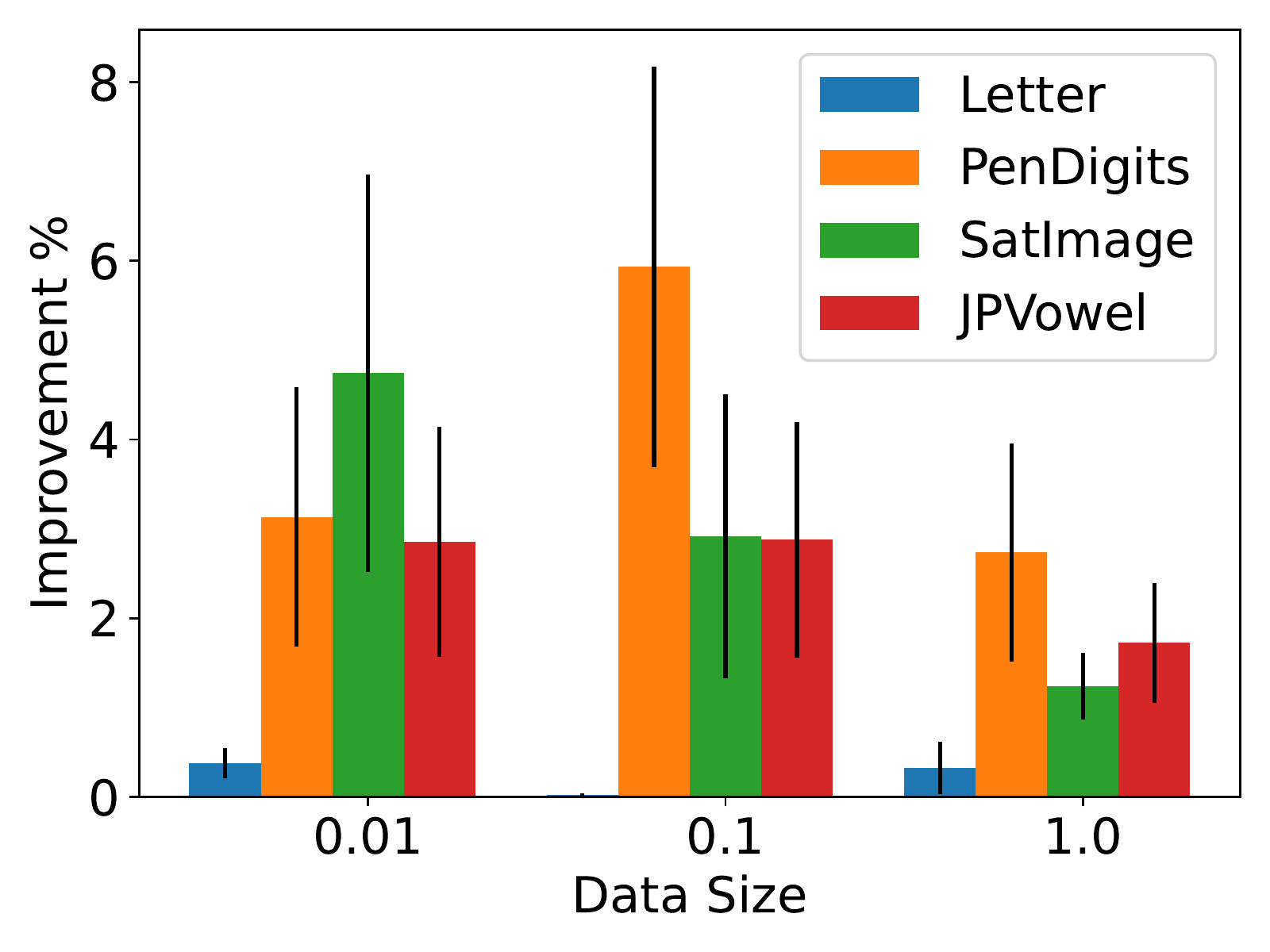}
\captionof{figure}{Same semantics as in Figure~\ref{fig:d_eps0.1_improvement_appendix}. \\
Environment: binary-valued cost, $\mu_{\textnormal{bad},\epsilon=0.1}$, and \# actions = \# classes. }
  \label{fig:d_no_noise_bad_improvement}
\end{minipage}
\hfill
\begin{minipage}{.48\textwidth}
  \centering
  \includegraphics[width=.98\linewidth]{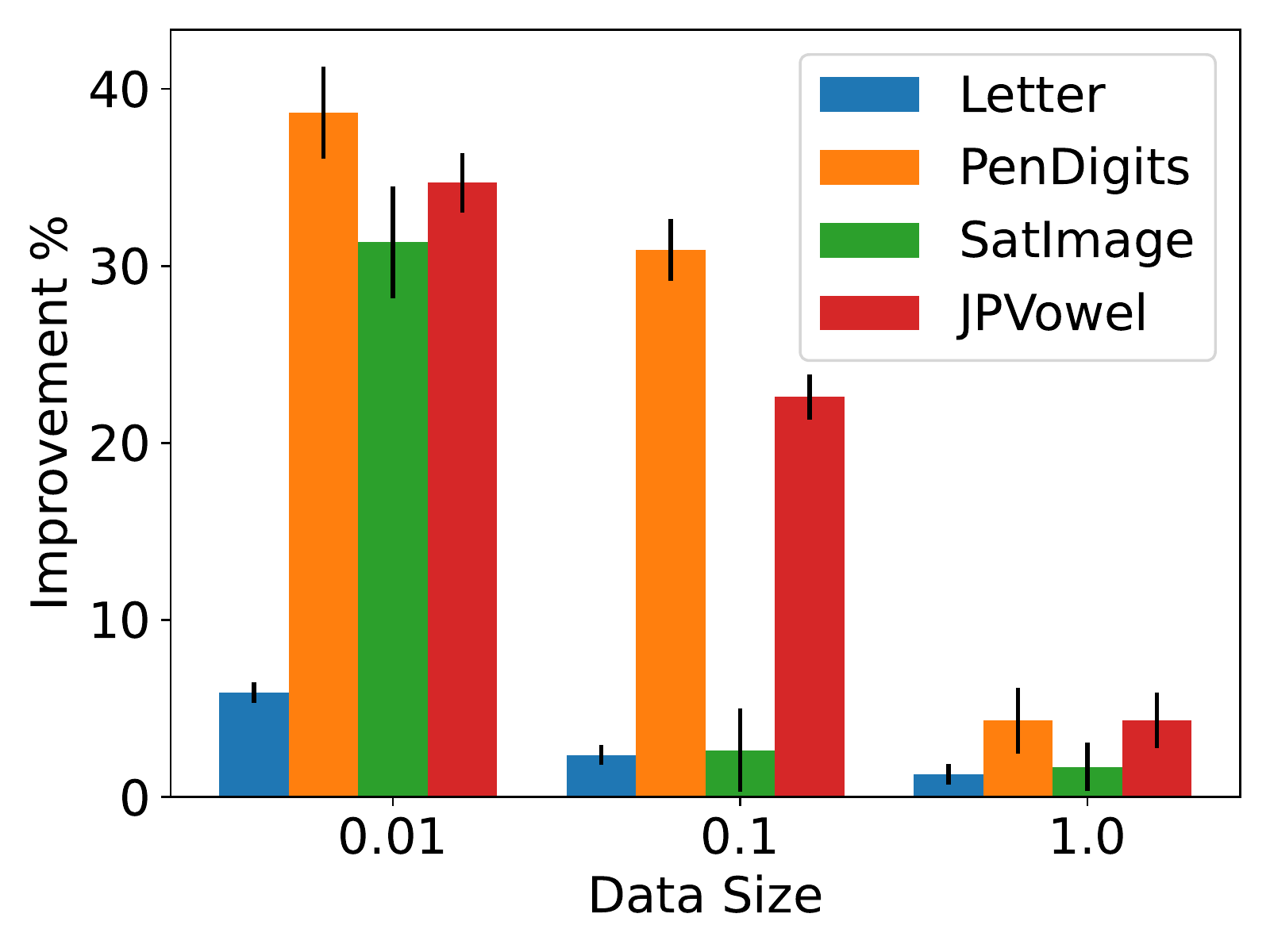}
\captionof{figure}{Same semantics as in Figure~\ref{fig:d_eps0.1_improvement_appendix}. \\
Environment: binary-valued cost, $\mu_{\textnormal{good},\epsilon=0.1}$, and \# actions = 5$\times$\# classes. }
  \label{fig:d_no_noise_large_action_improvement}
\end{minipage}
\end{figure}

\xhdr{Results. } We conduct experiments on all combinations of data sizes, cost types and logging policies with the number of actions being the number of classes in Table~\ref{tb:discrete-setting-options}. For the setting where the number of actions is 5 times the number of classes, we still do experiments on all combinations of data sizes, cost types, but only with logging policy $\pi_{\textnormal{good}, \epsilon = 0.1}$. 
The results are shown in 
\asedit{Figures \ref{fig:d_eps0.1_improvement_appendix}-\ref{fig:d_no_noise_large_action_improvement} and Tables~\ref{tb:exp-d-real-eps0.1-size0.01}-\ref{tb:exp-d-binary-large-action-size1}.
Their semantics mirror those of Figure~\ref{fig:main_discrete} (right) and Table~\ref{tb:exp-d-real-eps0.1-size0.1}, respectively. Relative improvement of a method against a baseline is defined as 
\begin{align}\label{eq:RelImp}
\RelImp := 
\frac{R\rbr{\widehat{\pi}_{\textnormal{baseline}}}-
            R\rbr{\widehat{\pi}_{\textnormal{method}}}}
{R\rbr{\widehat{\pi}_{\textnormal{baseline}}}}.
\end{align}
}
The total training time on each dataset is summarized in Table~\ref{tb:computation-discrete}.

\begin{table}[h]
\centering
\caption{Computation time for the discrete-action experiments}
\label{tb:computation-discrete}
\begin{tabular}{lllll}
\toprule
Dataset   &   Letter & PenDigits & SatImage & JPVowel  \\
\midrule
total time (CPU core $\cdot$ hours) & 730.8 & 424.2 & 417.9 & 397.0 \\
\bottomrule
\end{tabular}
\end{table}

\begin{table}[t]
\centering
\caption{
Performance of different OPO methods: mean $\pm$ two standard errors over 50 runs. Bold numbers represent the best performance within each (CSC oracle, estimator) pair. Boxed numbers represent the best across all algorithmic configurations. \\ This experiment:  
real-valued cost, logging policy $\mu_{\textnormal{good}, \epsilon=0.1}$, data size $\times0.01$, and \# actions = \# classes.}

\begin{tabular}{lllll}\toprule
Risk * 100	& Letter	& PenDigits	& SatImage	& JPVowel	\\
\midrule
PG+IPW+PL	& \fbox{\textbf{31.6±0.1}}	& 23.8±0.2	& \textbf{17.6±0.9}	& \textbf{12.0±0.1}	\\
PG+IPW+EB	& 34.0±0.9	& \textbf{23.4±0.2}	& 18.1±0.8	& 13.4±0.3	\\
PG+IPW	& 47.1±0.6	& 31.3±1.0	& 22.4±1.6	& 27.3±1.4	\\
\midrule
PG+DR+PL	& \fbox{\textbf{31.6±0.1}}	& 23.6±0.3	& 13.4±0.5	& \textbf{12.1±0.1}	\\
PG+DR+EB	& 40.1±0.4	& \fbox{\textbf{23.0±0.6}}	& \fbox{\textbf{13.1±0.4}}	& 20.1±0.8	\\
PG+DR	& 47.6±0.6	& 31.2±1.1	& 16.5±1.0	& 29.0±1.7	\\
\midrule
LR+IPW+PL	& \textbf{32.0±0.0}	& \textbf{23.6±0.2}	& \textbf{21.3±0.8}	& \fbox{\textbf{11.2±0.0}}	\\
LR+IPW	& 46.2±0.5	& 32.5±0.8	& 26.7±1.2	& 30.4±1.0	\\
\midrule
LR+DR+PL	& \textbf{32.0±0.1}	& \textbf{23.6±0.2}	& \textbf{21.3±0.8}	& \fbox{\textbf{11.2±0.0}}	\\
LR+DR	& 46.4±0.4	& 34.1±0.6	& 25.7±1.1	& 32.8±1.0	\\
\bottomrule
\end{tabular}
\label{tb:exp-d-real-eps0.1-size0.01}
\end{table}

\begin{table}[t]
\centering
\caption{Same semantics as in Table~\ref{tb:exp-d-real-eps0.1-size0.01}. \\
This experiment: real-valued cost, logging policy $\mu_{\textnormal{good}, \epsilon=0.1}$, data size $\times0.1$, and \# actions = \# classes. }

\begin{tabular}{llllll}\toprule
Risk $\times$ 100	& Letter	& PenDigits	& SatImage	& JPVowel	\\
\midrule
PG+IPW+PL	& \textbf{31.6±0.1}	& 22.0±0.3	& \textbf{10.2±0.4}	& \textbf{11.9±0.1}	\\
PG+IPW+EB	& 32.4±0.1	& \textbf{20.4±0.4}	& 11.0±0.3	& 13.1±0.1	\\
PG+IPW	& 45.5±0.8	& 26.4±0.8	& 11.9±0.7	& 19.0±0.8	\\
\midrule
PG+DR+PL	& \fbox{\textbf{31.5±0.1}}	& 18.5±0.3	& \fbox{\textbf{8.8±0.3}}	& \textbf{11.3±0.1}	\\
PG+DR+EB	& 37.0±0.4	& \fbox{\textbf{15.7±0.3}}	& \fbox{\textbf{8.8±0.2}}	& 11.7±0.2	\\
PG+DR	& 43.1±0.6	& 21.3±0.6	& 10.1±0.4	& 14.2±0.4	\\
\midrule
LR+IPW+PL	& \textbf{31.8±0.0}	& \textbf{23.1±0.3}	& \textbf{18.1±0.4}	& \fbox{\textbf{11.1±0.0}}	\\
LR+IPW	& 42.5±0.4	& 28.4±0.7	& 21.0±0.8	& 19.8±0.5	\\
\midrule
LR+DR+PL	& \textbf{31.8±0.1}	& \textbf{22.9±0.2}	& \textbf{17.8±0.5}	& \fbox{\textbf{11.1±0.0}}	\\
LR+DR	& 42.0±0.3	& 27.2±0.6	& 19.5±0.7	& 18.5±0.5	\\
\bottomrule
\end{tabular}
\label{tb:exp-d-real-eps0.1-size0.1-appendix}
\end{table}

\begin{table}[t]
\centering
\caption{Same semantics as in Table~\ref{tb:exp-d-real-eps0.1-size0.01}. \\
This experiment: real-valued cost, logging policy $\mu_{\textnormal{good}, \epsilon=0.1}$, data size $\times1$, and \# actions = \# classes. }

\begin{tabular}{llllll}\toprule
Risk * 100	& Letter	& PenDigits	& SatImage	& JPVowel	\\
\midrule
PG+IPW+PL	& \textbf{31.0±0.1}	& 15.8±0.3	& \textbf{7.4±0.1}	& 10.5±0.2	\\
PG+IPW+EB	& 31.7±0.2	& \textbf{13.2±0.2}	& 7.7±0.1	& \textbf{9.9±0.2}	\\
PG+IPW	& 35.1±0.4	& 16.9±0.5	& 8.3±0.4	& 11.7±0.3	\\
\midrule
PG+DR+PL	& 30.7±0.2	& 14.2±0.3	& 6.9±0.1	& 9.3±0.2	\\
PG+DR+EB	& \fbox{\textbf{29.9±0.1}}	& \fbox{\textbf{10.2±0.1}}	& \fbox{\textbf{6.3±0.0}}	& \fbox{\textbf{8.0±0.1}}	\\
PG+DR	& 33.0±0.4	& 15.1±0.4	& 7.2±0.1	& 9.8±0.3	\\
\midrule
LR+IPW+PL	& \textbf{31.5±0.1}	& \textbf{20.8±0.2}	& \textbf{12.8±0.3}	& \textbf{11.1±0.0}	\\
LR+IPW	& 36.2±0.3	& 21.9±0.4	& 15.0±0.7	& 12.5±0.2	\\
\midrule
LR+DR+PL	& \textbf{31.5±0.1}	& \textbf{20.4±0.2}	& \textbf{11.1±0.2}	& \textbf{11.0±0.0}	\\
LR+DR	& 35.7±0.2	& 21.2±0.3	& 12.1±0.5	& 11.8±0.1	\\
\bottomrule
\end{tabular}
\label{tb:exp-d-real-eps0.1-size1}
\end{table}

\begin{table}[t]
\centering
\caption{Same semantics as in Table~\ref{tb:exp-d-real-eps0.1-size0.01}. \\
This experiment: real-valued cost, logging policy $\mu_{\textnormal{good}, \epsilon=0.01}$, data size $\times0.01$, and \# actions = \# classes. }

\begin{tabular}{lllll}\toprule
Risk * 100	& Letter	& PenDigits	& SatImage	& JPVowel	\\
\midrule
PG+IPW+PL	& 34.2±0.1	& \textbf{21.5±0.2}	& \textbf{24.3±0.5}	& 16.1±0.1	\\
PG+IPW+EB	& \fbox{\textbf{33.7±0.3}}	& 22.0±0.2	& 25.6±0.4	& \fbox{\textbf{16.0±0.0}}	\\
PG+IPW	& 46.8±0.9	& 39.8±1.7	& 39.7±2.7	& 41.6±1.5	\\
\midrule
PG+DR+PL	& \textbf{34.3±0.1}	& \textbf{21.6±0.1}	& 24.2±0.5	& \textbf{16.1±0.1}	\\
PG+DR+EB	& 36.8±0.4	& 24.8±0.8	& \fbox{\textbf{18.8±0.4}}	& 22.0±0.9	\\
PG+DR	& 47.3±0.8	& 42.9±1.4	& 42.9±2.4	& 43.5±1.4	\\
\midrule
LR+IPW+PL	& \textbf{33.8±0.1}	& \fbox{\textbf{21.0±0.1}}	& \textbf{25.6±0.6}	& \textbf{16.6±0.1}	\\
LR+IPW	& 46.1±0.6	& 40.7±1.3	& 44.4±1.7	& 41.3±1.2	\\
\midrule
LR+DR+PL	& \textbf{33.9±0.1}	& \fbox{\textbf{21.0±0.1}}	& \textbf{25.8±0.4}	& \textbf{16.7±0.1}	\\
LR+DR	& 46.8±0.5	& 43.5±1.1	& 48.0±1.7	& 44.1±0.9	\\
\bottomrule
\end{tabular}
\label{tb:exp-d-real-eps0.01-size0.01}
\end{table}

\begin{table}[t]
\centering
\caption{Same semantics as in Table~\ref{tb:exp-d-real-eps0.1-size0.01}. \\
This experiment: real-valued cost, logging policy $\mu_{\textnormal{good}, \epsilon=0.01}$, data size $\times0.1$, and \# actions = \# classes. }

\begin{tabular}{llllll}\toprule
Risk * 100	& Letter	& PenDigits	& SatImage	& JPVowel	\\
\midrule
PG+IPW+PL	& \textbf{33.7±0.1}	& \textbf{21.5±0.2}	& \textbf{23.0±0.6}	& \fbox{\textbf{15.9±0.0}}	\\
PG+IPW+EB	& 36.0±0.8	& 22.5±0.4	& 24.4±0.6	& 16.2±0.1	\\
PG+IPW	& 45.1±0.7	& 33.8±1.2	& 31.0±2.0	& 33.1±1.2	\\
\midrule
PG+DR+PL	& \fbox{\textbf{33.6±0.1}}	& \textbf{21.3±0.2}	& 21.6±0.6	& \fbox{\textbf{15.9±0.0}}	\\
PG+DR+EB	& 36.6±0.6	& 22.8±0.7	& \fbox{\textbf{18.1±0.4}}	& 21.1±1.0	\\
PG+DR	& 45.1±0.5	& 33.0±1.1	& 26.2±1.7	& 33.9±1.5	\\
\midrule
LR+IPW+PL	& \textbf{34.0±0.1}	& \textbf{21.0±0.1}	& \textbf{25.1±0.4}	& \textbf{16.4±0.0}	\\
LR+IPW	& 45.0±0.6	& 35.2±0.9	& 36.8±1.5	& 35.6±1.0	\\
\midrule
LR+DR+PL	& \textbf{34.0±0.1}	& \fbox{\textbf{20.9±0.1}}	& \textbf{25.3±0.3}	& \textbf{16.4±0.1}	\\
LR+DR	& 45.2±0.5	& 38.0±0.8	& 37.2±1.2	& 37.3±1.0	\\
\bottomrule
\end{tabular}
\label{tb:exp-d-real-eps0.01-size0.1}
\end{table}

\begin{table}[t]
\centering
\caption{Same semantics as in Table~\ref{tb:exp-d-real-eps0.1-size0.01}. \\
This experiment: real-valued cost, logging policy $\mu_{\textnormal{good}, \epsilon=0.01}$, data size $\times1$, and \# actions = \# classes. }

\begin{tabular}{llllll}\toprule
Risk * 100	& Letter	& PenDigits	& SatImage	& JPVowel	\\
\midrule
PG+IPW+PL	& \fbox{\textbf{33.6±0.1}}	& 20.8±0.3	& 19.3±0.4	& \textbf{15.9±0.1}	\\
PG+IPW+EB	& 36.0±0.6	& \textbf{20.3±0.4}	& \textbf{19.1±0.4}	& \textbf{15.9±0.1}	\\
PG+IPW	& 43.7±0.7	& 25.5±0.9	& 20.4±0.7	& 24.8±0.8	\\
\midrule
PG+DR+PL	& \fbox{\textbf{33.6±0.1}}	& 19.8±0.3	& 18.2±0.4	& \fbox{\textbf{15.3±0.1}}	\\
PG+DR+EB	& 35.8±0.4	& \fbox{\textbf{16.7±0.3}}	& \fbox{\textbf{15.9±0.3}}	& \fbox{\textbf{15.3±0.3}}	\\
PG+DR	& 42.9±0.6	& 21.8±0.8	& 18.8±0.6	& 19.6±0.6	\\
\midrule
LR+IPW+PL	& \textbf{33.7±0.1}	& \textbf{20.5±0.1}	& \textbf{24.9±0.4}	& \textbf{16.3±0.0}	\\
LR+IPW	& 41.6±0.4	& 28.5±0.7	& 34.6±1.6	& 25.8±0.7	\\
\midrule
LR+DR+PL	& \textbf{33.9±0.1}	& \textbf{20.7±0.1}	& \textbf{24.6±0.4}	& \textbf{16.2±0.1}	\\
LR+DR	& 41.1±0.3	& 28.1±0.7	& 34.3±1.7	& 25.1±0.6	\\
\bottomrule
\end{tabular}
\label{tb:exp-d-real-eps0.01-size1}
\end{table}

\begin{table}[t]
\centering
\caption{Same semantics as in Table~\ref{tb:exp-d-real-eps0.1-size0.01}. \\
This experiment: real-valued cost, logging policy $\mu_{\textnormal{bad}, \epsilon=0.1}$, data size $\times0.01$, and \# actions = \# classes. }

\begin{tabular}{lllll}\toprule
Risk * 100	& Letter	& PenDigits	& SatImage	& JPVowel	\\
\midrule
PG+IPW+PL	& 44.1±0.5	& 31.1±1.1	& \textbf{16.7±1.2}	& \textbf{27.4±0.7}	\\
PG+IPW+EB	& \textbf{43.8±0.4}	& \textbf{31.0±0.8}	& 17.9±1.0	& 28.6±0.9	\\
PG+IPW	& 45.3±0.6	& 34.4±1.5	& 20.5±1.7	& 29.2±0.9	\\
\midrule
PG+DR+PL	& 43.9±0.6	& 29.8±1.1	& \fbox{\textbf{13.0±0.7}}	& 27.1±0.8	\\
PG+DR+EB	& \fbox{\textbf{43.0±0.4}}	& \fbox{\textbf{28.0±0.8}}	& 15.1±0.6	& \fbox{\textbf{25.0±0.8}}	\\
PG+DR	& 45.6±0.9	& 33.1±1.1	& 15.4±0.9	& 29.5±1.0	\\
\midrule
LR+IPW+PL	& \textbf{43.9±0.4}	& \textbf{34.2±0.8}	& \textbf{21.9±0.9}	& \textbf{29.5±0.7}	\\
LR+IPW	& 44.6±0.5	& 35.9±0.9	& 25.4±1.5	& 31.0±0.9	\\
\midrule
LR+DR+PL	& \textbf{43.7±0.5}	& \textbf{34.7±0.7}	& \textbf{21.1±0.7}	& \textbf{29.8±0.7}	\\
LR+DR	& 44.2±0.5	& 35.7±0.8	& 23.5±1.1	& 32.0±1.0	\\
\bottomrule
\end{tabular}
\label{tb:exp-d-real-bad-size0.01}
\end{table}

\begin{table}[t]
\centering
\caption{Same semantics as in Table~\ref{tb:exp-d-real-eps0.1-size0.01}. \\
This experiment: real-valued cost, logging policy $\mu_{\textnormal{bad}, \epsilon=0.1}$, data size $\times0.1$, and \# actions = \# classes. }

\begin{tabular}{llllll}\toprule
Risk * 100	& Letter	& PenDigits	& SatImage	& JPVowel	\\
\midrule
PG+IPW+PL	& 42.2±0.5	& 23.6±0.6	& \textbf{10.1±0.3}	& 21.7±0.6	\\
PG+IPW+EB	& \textbf{40.2±0.3}	& \textbf{20.0±0.5}	& 11.8±0.3	& \textbf{19.1±0.4}	\\
PG+IPW	& 43.7±0.6	& 26.3±1.0	& 11.4±0.5	& 24.0±0.9	\\
\midrule
PG+DR+PL	& 41.1±0.5	& 21.3±0.8	& \fbox{\textbf{9.2±0.2}}	& 18.1±0.4	\\
PG+DR+EB	& \fbox{\textbf{38.9±0.3}}	& \fbox{\textbf{17.8±0.5}}	& 10.1±0.3	& \fbox{\textbf{17.2±0.4}}	\\
PG+DR	& 43.0±0.8	& 23.8±0.8	& 10.0±0.4	& 19.2±0.6	\\
\midrule
LR+IPW+PL	& \textbf{40.2±0.4}	& \textbf{27.1±0.6}	& \textbf{17.0±0.4}	& \textbf{22.5±0.5}	\\
LR+IPW	& 40.9±0.4	& 28.3±0.8	& 18.8±0.7	& 24.0±0.7	\\
\midrule
LR+DR+PL	& \textbf{39.6±0.3}	& \textbf{25.7±0.7}	& \textbf{16.6±0.6}	& \textbf{20.3±0.4}	\\
LR+DR	& 40.1±0.3	& 26.8±0.8	& 18.5±0.8	& 21.1±0.5	\\
\bottomrule
\end{tabular}
\label{tb:exp-d-real-bad-size0.1}
\end{table}

\begin{table}[t]
\centering
\caption{Same semantics as in Table~\ref{tb:exp-d-real-eps0.1-size0.01}. \\
This experiment: real-valued cost, logging policy $\mu_{\textnormal{bad}, \epsilon=0.1}$, data size $\times1$, and \# actions = \# classes. }

\begin{tabular}{llllll}\toprule
Risk * 100	& Letter	& PenDigits	& SatImage	& JPVowel	\\
\midrule
PG+IPW+PL	& 35.3±0.4	& 12.8±0.4	& 7.9±0.2	& 13.6±0.2	\\
PG+IPW+EB	& \textbf{32.6±0.2}	& \textbf{12.0±0.2}	& \textbf{7.6±0.1}	& \textbf{13.0±0.2}	\\
PG+IPW	& 36.1±0.5	& 14.2±0.7	& 8.4±0.2	& 14.3±0.3	\\
\midrule
PG+DR+PL	& 33.0±0.4	& 11.9±0.3	& 6.8±0.2	& 12.9±0.2	\\
PG+DR+EB	& \fbox{\textbf{30.5±0.1}}	& \fbox{\textbf{10.9±0.1}}	& \fbox{\textbf{6.5±0.1}}	& \fbox{\textbf{11.0±0.1}}	\\
PG+DR	& 33.6±0.4	& 12.9±0.4	& 7.4±0.3	& 13.5±0.3	\\
\midrule
LR+IPW+PL	& \textbf{35.4±0.2}	& \textbf{19.6±0.4}	& \textbf{11.7±0.4}	& \textbf{15.9±0.2}	\\
LR+IPW	& 35.5±0.2	& 20.9±0.6	& 13.4±0.7	& 16.3±0.2	\\
\midrule
LR+DR+PL	& \textbf{35.0±0.2}	& \textbf{19.0±0.3}	& \textbf{10.5±0.2}	& \textbf{15.0±0.2}	\\
LR+DR	& 35.2±0.2	& 20.3±0.4	& 11.6±0.3	& 15.3±0.2	\\
\bottomrule
\end{tabular}
\label{tb:exp-d-real-bad-size1}
\end{table}

\begin{table}[t]
\centering
\caption{Same semantics as in Table~\ref{tb:exp-d-real-eps0.1-size0.01}. \\
This experiment: real-valued cost, logging policy $\mu_{\textnormal{good}, \epsilon=0.1}$, data size $\times0.01$, and \# actions = $5\times$ \# classes. }

\begin{tabular}{lllll}\toprule
Risk * 100	& Letter	& PenDigits	& SatImage	& JPVowel	\\
\midrule
PG+IPW+PL	& \fbox{\textbf{33.3±0.2}}	& \fbox{\textbf{11.1±0.2}}	& \textbf{8.1±0.3}	& \textbf{11.8±0.2}	\\
PG+IPW+EB	& 40.4±1.0	& 16.9±1.6	& 9.8±0.9	& 17.1±1.9	\\
PG+IPW	& 47.2±0.7	& 35.2±1.6	& 28.2±2.1	& 37.1±1.3	\\
\midrule
PG+DR+PL	& \textbf{33.4±0.2}	& \fbox{\textbf{11.1±0.1}}	& \textbf{8.1±0.3}	& \textbf{11.9±0.2}	\\
PG+DR+EB	& 36.0±0.4	& 15.8±0.6	& 9.9±0.5	& 15.0±0.5	\\
PG+DR	& 47.4±0.9	& 38.2±1.6	& 31.3±2.3	& 39.6±1.1	\\
\midrule
LR+IPW+PL	& \textbf{35.0±0.2}	& \textbf{11.9±0.1}	& \fbox{\textbf{7.4±0.1}}	& \textbf{11.7±0.1}	\\
LR+IPW	& 45.4±0.5	& 32.7±1.1	& 28.2±1.4	& 33.0±1.0	\\
\midrule
LR+DR+PL	& \textbf{33.5±0.1}	& \textbf{12.2±0.2}	& \textbf{7.6±0.1}	& \fbox{\textbf{11.5±0.1}}	\\
LR+DR	& 47.2±0.5	& 39.6±1.1	& 35.5±1.3	& 39.7±0.8	\\
\bottomrule
\end{tabular}
\label{tb:exp-d-real-large-action-size0.01}
\end{table}

\begin{table}[t]
\centering
\caption{Same semantics as in Table~\ref{tb:exp-d-real-eps0.1-size0.01}. \\
This experiment: real-valued cost, logging policy $\mu_{\textnormal{good}, \epsilon=0.1}$, data size $\times0.1$, and \# actions = $5\times$ \# classes. }

\begin{tabular}{llllll}\toprule
Risk * 100	& Letter	& PenDigits	& SatImage	& JPVowel	\\
\midrule
PG+IPW+PL	& \textbf{33.5±0.1}	& \textbf{11.4±0.1}	& \textbf{7.3±0.1}	& \textbf{11.1±0.1}	\\
PG+IPW+EB	& 41.7±0.6	& 17.8±1.4	& 8.4±0.2	& 15.0±1.6	\\
PG+IPW	& 45.5±1.0	& 30.4±1.5	& 18.5±1.3	& 32.0±1.5	\\
\midrule
PG+DR+PL	& \textbf{33.3±0.1}	& \textbf{11.3±0.1}	& \fbox{\textbf{7.0±0.2}}	& \fbox{\textbf{10.9±0.1}}	\\
PG+DR+EB	& 40.1±0.3	& 14.6±0.6	& 8.4±0.2	& 14.2±0.6	\\
PG+DR	& 45.6±0.8	& 27.0±1.4	& 15.2±1.2	& 29.2±1.6	\\
\midrule
LR+IPW+PL	& \textbf{33.1±0.1}	& \fbox{\textbf{11.1±0.0}}	& \textbf{7.6±0.1}	& \textbf{11.6±0.0}	\\
LR+IPW	& 43.7±0.3	& 25.2±0.7	& 20.3±0.7	& 24.2±0.6	\\
\midrule
LR+DR+PL	& \fbox{\textbf{33.0±0.0}}	& \fbox{\textbf{11.1±0.0}}	& \textbf{7.7±0.1}	& \textbf{11.6±0.0}	\\
LR+DR	& 45.0±0.3	& 26.7±0.6	& 23.3±1.1	& 26.3±0.5	\\
\bottomrule
\end{tabular}
\label{tb:exp-d-real-large-action-size0.1}
\end{table}

\begin{table}[t]
\centering
\caption{Same semantics as in Table~\ref{tb:exp-d-real-eps0.1-size0.01}. \\
This experiment: real-valued cost, logging policy $\mu_{\textnormal{good}, \epsilon=0.1}$, data size $\times1$, and \# actions = $5\times$ \# classes. }

\begin{tabular}{llllll}\toprule
Risk * 100	& Letter	& PenDigits	& SatImage	& JPVowel	\\
\midrule
PG+IPW+PL	& \textbf{33.3±0.1}	& \textbf{10.9±0.2}	& \textbf{6.8±0.2}	& \textbf{10.9±0.1}	\\
PG+IPW+EB	& 34.9±0.2	& 12.6±0.1	& 7.9±0.1	& 11.0±0.0	\\
PG+IPW	& 40.6±0.7	& 16.6±0.8	& 12.3±0.8	& 18.0±0.7	\\
\midrule
PG+DR+PL	& 32.7±0.2	& 10.2±0.2	& 5.8±0.2	& 10.7±0.1	\\
PG+DR+EB	& \textbf{32.4±0.3}	& \fbox{\textbf{9.6±0.1}}	& \fbox{\textbf{5.6±0.1}}	& \fbox{\textbf{9.9±0.1}}	\\
PG+DR	& 37.5±0.7	& 12.3±0.4	& 8.4±0.4	& 14.3±0.6	\\
\midrule
LR+IPW+PL	& \textbf{32.7±0.1}	& \textbf{11.0±0.1}	& \textbf{7.5±0.1}	& \textbf{11.4±0.1}	\\
LR+IPW	& 38.6±0.3	& 18.9±0.3	& 13.4±0.3	& 18.1±0.3	\\
\midrule
LR+DR+PL	& \fbox{\textbf{32.2±0.1}}	& \textbf{10.9±0.1}	& \textbf{7.9±0.1}	& \textbf{11.4±0.0}	\\
LR+DR	& 37.5±0.3	& 20.9±0.6	& 13.6±0.3	& 18.9±0.4	\\
\bottomrule
\end{tabular}
\label{tb:exp-d-real-large-action-size1}
\end{table}

\begin{table}[t]
\centering
\caption{Same semantics as in Table~\ref{tb:exp-d-real-eps0.1-size0.01}. \\
This experiment:  binary-valued cost, logging policy $\mu_{\textnormal{good}, \epsilon=0.1}$, data size $\times0.01$, and \# actions = \# classes. }

\begin{tabular}{lllll}\toprule
Risk * 100	& Letter	& PenDigits	& SatImage	& JPVowel	\\
\midrule
PG+IPW+PL	& 83.5±0.1	& 43.0±1.0	& 25.2±0.8	& 36.8±0.5	\\
PG+IPW+EB	& \fbox{\textbf{76.9±0.6}}	& \textbf{31.8±0.8}	& \textbf{22.9±0.4}	& \textbf{30.7±0.4}	\\
PG+IPW	& 91.2±0.6	& 49.1±2.1	& 27.0±0.8	& 43.0±2.0	\\
\midrule
PG+DR+PL	& 84.2±0.3	& 44.5±0.5	& 26.4±1.0	& 38.0±0.7	\\
PG+DR+EB	& \textbf{83.6±0.8}	& \fbox{\textbf{31.2±1.2}}	& \fbox{\textbf{21.4±0.5}}	& \fbox{\textbf{29.8±0.7}}	\\
PG+DR	& 94.0±0.4	& 66.2±2.3	& 33.2±2.3	& 71.1±1.5	\\
\midrule
LR+IPW+PL	& \textbf{84.6±0.2}	& \textbf{44.5±0.7}	& \textbf{34.8±1.1}	& \textbf{38.7±0.3}	\\
LR+IPW	& 91.8±0.4	& 59.5±1.8	& 40.4±1.4	& 57.3±1.7	\\
\midrule
LR+DR+PL	& \textbf{85.1±0.5}	& \textbf{46.3±0.3}	& \textbf{37.0±1.1}	& \textbf{40.2±0.3}	\\
LR+DR	& 92.9±0.4	& 77.7±1.2	& 52.3±2.8	& 77.0±1.2	\\
\bottomrule
\end{tabular}
\label{tb:exp-d-binary-eps0.1-size0.01}
\end{table}

\begin{table}[t]
\centering
\caption{Same semantics as in Table~\ref{tb:exp-d-real-eps0.1-size0.01}. \\
This experiment:  binary-valued cost, logging policy $\mu_{\textnormal{good}, \epsilon=0.1}$, data size $\times0.1$, and \# actions = \# classes. }

\begin{tabular}{llllll}\toprule
Risk * 100	& Letter	& PenDigits	& SatImage	& JPVowel	\\
\midrule
PG+IPW+PL	& 82.7±0.1	& 26.4±0.4	& 19.8±0.1	& 29.6±0.5	\\
PG+IPW+EB	& \textbf{74.7±1.4}	& \textbf{24.8±0.7}	& \textbf{19.5±0.1}	& \fbox{\textbf{26.2±0.5}}	\\
PG+IPW	& 85.5±0.6	& 28.7±0.9	& 20.1±0.2	& 30.5±0.5	\\
\midrule
PG+DR+PL	& 82.7±0.1	& 29.1±0.9	& 19.7±0.1	& 31.3±0.5	\\
PG+DR+EB	& \fbox{\textbf{69.5±0.9}}	& \fbox{\textbf{23.8±0.5}}	& \fbox{\textbf{18.8±0.2}}	& \textbf{26.7±0.4}	\\
PG+DR	& 87.8±0.5	& 34.7±1.6	& 20.9±0.6	& 33.4±0.8	\\
\midrule
LR+IPW+PL	& \textbf{82.1±0.5}	& \textbf{32.0±0.5}	& \textbf{26.5±0.4}	& \textbf{31.8±0.4}	\\
LR+IPW	& 85.2±0.4	& 34.4±0.9	& 30.0±1.1	& 32.9±0.5	\\
\midrule
LR+DR+PL	& \textbf{83.6±0.3}	& \textbf{41.9±0.6}	& \textbf{31.8±0.6}	& \textbf{37.4±0.3}	\\
LR+DR	& 87.1±0.4	& 53.1±1.6	& 38.1±1.9	& 49.3±1.0	\\
\bottomrule
\end{tabular}
\label{tb:exp-d-binary-eps0.1-size0.1}
\end{table}

\begin{table}[t]
\centering
\caption{Same semantics as in Table~\ref{tb:exp-d-real-eps0.1-size0.01}. \\
This experiment:  binary-valued cost, logging policy $\mu_{\textnormal{good}, \epsilon=0.1}$, data size $\times1$, and \# actions = \# classes. }

\begin{tabular}{llllll}\toprule
Risk * 100	& Letter	& PenDigits	& SatImage	& JPVowel	\\
\midrule
PG+IPW+PL	& 76.5±0.5	& 22.3±0.2	& 18.7±0.1	& 23.5±0.1	\\
PG+IPW+EB	& \textbf{66.4±2.5}	& \fbox{\textbf{20.3±0.1}}	& \textbf{16.8±0.1}	& \fbox{\textbf{21.9±0.1}}	\\
PG+IPW	& 76.9±0.6	& 23.0±0.5	& 18.9±0.1	& 23.8±0.2	\\
\midrule
PG+DR+PL	& 77.1±0.5	& 22.3±0.1	& 18.5±0.1	& 23.3±0.1	\\
PG+DR+EB	& \fbox{\textbf{64.4±1.6}}	& \textbf{20.4±0.1}	& \fbox{\textbf{16.6±0.0}}	& \fbox{\textbf{21.9±0.0}}	\\
PG+DR	& 77.5±0.5	& 23.8±0.6	& 18.7±0.1	& 23.8±0.3	\\
\midrule
LR+IPW+PL	& \textbf{78.2±0.4}	& \textbf{27.0±0.2}	& \textbf{24.2±0.3}	& \textbf{28.9±0.2}	\\
LR+IPW	& 81.0±0.5	& 27.2±0.2	& 25.3±0.3	& 29.0±0.3	\\
\midrule
LR+DR+PL	& \textbf{80.0±0.4}	& \textbf{29.1±0.2}	& \textbf{23.9±0.4}	& \textbf{31.0±0.2}	\\
LR+DR	& 83.2±0.5	& 29.7±0.3	& 25.4±0.4	& 31.4±0.2	\\
\bottomrule
\end{tabular}
\label{tb:exp-d-binary-eps0.1-size1}
\end{table}

\begin{table}[t]
\centering
\caption{Same semantics as in Table~\ref{tb:exp-d-real-eps0.1-size0.01}. \\
This experiment:  binary-valued cost, logging policy $\mu_{\textnormal{good}, \epsilon=0.01}$, data size $\times0.01$, and \# actions = \# classes. }

\begin{tabular}{lllll}\toprule
Risk * 100	& Letter	& PenDigits	& SatImage	& JPVowel	\\
\midrule
PG+IPW+PL	& 78.1±0.5	& 48.6±1.1	& 31.9±1.0	& 38.8±0.5	\\
PG+IPW+EB	& \fbox{\textbf{69.3±1.1}}	& \fbox{\textbf{38.4±0.6}}	& \textbf{28.3±0.4}	& \textbf{33.9±1.3}	\\
PG+IPW	& 85.4±2.2	& 53.4±4.2	& 35.8±2.8	& 48.1±4.8	\\
\midrule
PG+DR+PL	& 83.6±1.7	& 53.7±1.8	& 33.6±0.6	& 45.3±3.2	\\
PG+DR+EB	& \textbf{77.7±1.1}	& \textbf{38.5±2.9}	& \fbox{\textbf{25.5±1.2}}	& \fbox{\textbf{31.7±1.5}}	\\
PG+DR	& 95.3±0.3	& 84.0±1.7	& 61.2±3.9	& 85.0±1.1	\\
\midrule
LR+IPW+PL	& \textbf{81.3±0.1}	& \textbf{53.2±1.0}	& \textbf{41.0±1.2}	& \textbf{46.5±1.4}	\\
LR+IPW	& 91.7±0.7	& 69.5±3.5	& 55.8±3.1	& 66.5±4.0	\\
\midrule
LR+DR+PL	& \textbf{86.4±1.5}	& \textbf{57.2±2.3}	& \textbf{40.7±1.4}	& \textbf{52.6±2.6}	\\
LR+DR	& 95.3±0.3	& 83.8±1.4	& 72.4±2.5	& 84.2±1.1	\\
\bottomrule
\end{tabular}
\label{tb:exp-d-binary-eps0.01-size0.01}
\end{table}

\begin{table}[t]
\centering
\caption{Same semantics as in Table~\ref{tb:exp-d-real-eps0.1-size0.01}. \\
This experiment:  binary-valued cost, logging policy $\mu_{\textnormal{good}, \epsilon=0.01}$, data size $\times0.1$, and \# actions = \# classes. }

\begin{tabular}{llllll}\toprule
Risk * 100	& Letter	& PenDigits	& SatImage	& JPVowel	\\
\midrule
PG+IPW+PL	& 76.4±0.3	& 49.3±0.3	& 27.4±1.1	& 38.9±0.2	\\
PG+IPW+EB	& \fbox{\textbf{69.9±0.5}}	& \textbf{32.4±1.2}	& \textbf{24.5±0.5}	& \textbf{31.3±0.8}	\\
PG+IPW	& 87.3±1.7	& 51.0±1.8	& 27.6±0.9	& 47.5±1.9	\\
\midrule
PG+DR+PL	& \textbf{77.0±0.3}	& 49.4±0.5	& 27.8±1.0	& 39.1±0.2	\\
PG+DR+EB	& 77.7±1.6	& \fbox{\textbf{26.9±1.6}}	& \fbox{\textbf{19.2±0.5}}	& \fbox{\textbf{26.1±0.5}}	\\
PG+DR	& 94.2±0.4	& 64.3±1.8	& 31.3±1.9	& 67.3±1.9	\\
\midrule
LR+IPW+PL	& \textbf{80.9±0.1}	& \textbf{51.2±0.8}	& \textbf{33.2±1.1}	& \textbf{44.4±0.6}	\\
LR+IPW	& 91.1±1.0	& 61.8±1.9	& 39.1±1.4	& 56.0±2.2	\\
\midrule
LR+DR+PL	& \textbf{81.9±0.5}	& \textbf{52.6±0.4}	& \textbf{36.2±0.9}	& \textbf{47.3±0.6}	\\
LR+DR	& 93.5±0.3	& 74.2±1.4	& 47.1±1.7	& 76.1±1.4	\\
\bottomrule
\end{tabular}
\label{tb:exp-d-binary-eps0.01-size0.1}
\end{table}

\begin{table}[t]
\centering
\caption{Same semantics as in Table~\ref{tb:exp-d-real-eps0.1-size0.01}. \\
This experiment:  binary-valued cost, logging policy $\mu_{\textnormal{good}, \epsilon=0.01}$, data size $\times1$, and \# actions = \# classes. }

\begin{tabular}{llllll}\toprule
Risk * 100	& Letter	& PenDigits	& SatImage	& JPVowel	\\
\midrule
PG+IPW+PL	& 75.7±0.2	& 28.6±0.8	& 20.4±0.3	& 29.4±0.4	\\
PG+IPW+EB	& \textbf{66.4±1.3}	& \textbf{24.6±0.7}	& \textbf{19.7±0.1}	& \fbox{\textbf{25.6±0.3}}	\\
PG+IPW	& 84.6±0.7	& 28.6±0.7	& 20.6±0.4	& 29.4±0.4	\\
\midrule
PG+DR+PL	& 76.0±0.2	& 35.4±1.2	& 20.5±0.6	& 31.6±0.6	\\
PG+DR+EB	& \fbox{\textbf{66.0±0.9}}	& \fbox{\textbf{24.1±0.5}}	& \fbox{\textbf{18.4±0.2}}	& \textbf{26.2±0.4}	\\
PG+DR	& 90.0±0.7	& 37.2±1.7	& 21.0±0.7	& 35.5±1.1	\\
\midrule
LR+IPW+PL	& \textbf{80.5±0.3}	& \textbf{34.9±1.0}	& \textbf{27.9±0.6}	& 34.0±0.6	\\
LR+IPW	& 84.7±0.6	& 35.0±1.0	& 29.4±0.8	& \textbf{33.9±0.6}	\\
\midrule
LR+DR+PL	& \textbf{80.9±0.2}	& \textbf{45.7±1.0}	& \textbf{31.2±0.8}	& \textbf{42.1±0.5}	\\
LR+DR	& 89.0±0.4	& 52.3±1.5	& 41.5±2.1	& 51.4±1.2	\\
\bottomrule
\end{tabular}
\label{tb:exp-d-binary-eps0.01-size1}
\end{table}

\begin{table}[t]
\centering
\caption{Same semantics as in Table~\ref{tb:exp-d-real-eps0.1-size0.01}. \\
This experiment:  binary-valued cost, logging policy $\mu_{\textnormal{bad}, \epsilon=0.1}$, data size $\times0.01$, and \# actions = \# classes. }

\begin{tabular}{lllll}\toprule
Risk * 100	& Letter	& PenDigits	& SatImage	& JPVowel	\\
\midrule
PG+IPW+PL	& 92.1±0.3	& 60.7±1.6	& 31.7±1.0	& 62.2±1.1	\\
PG+IPW+EB	& \fbox{\textbf{90.1±0.5}}	& \fbox{\textbf{46.3±1.4}}	& \textbf{27.6±0.9}	& \textbf{49.5±1.3}	\\
PG+IPW	& 92.5±0.4	& 63.8±1.7	& 35.8±2.0	& 64.4±1.3	\\
\midrule
PG+DR+PL	& 92.4±0.3	& 61.9±1.5	& 31.6±1.1	& 63.1±1.5	\\
PG+DR+EB	& \textbf{90.6±0.5}	& \textbf{46.4±1.6}	& \fbox{\textbf{26.6±0.9}}	& \fbox{\textbf{49.4±1.1}}	\\
PG+DR	& 93.0±0.4	& 64.4±1.7	& 34.4±1.9	& 66.0±1.6	\\
\midrule
LR+IPW+PL	& \textbf{90.3±0.5}	& \textbf{67.2±1.3}	& \textbf{40.9±1.2}	& \textbf{67.2±1.2}	\\
LR+IPW	& 90.7±0.5	& 69.8±1.6	& 45.6±2.0	& 69.6±1.6	\\
\midrule
LR+DR+PL	& \textbf{90.3±0.5}	& \textbf{66.5±1.5}	& \textbf{43.1±1.5}	& \textbf{66.2±1.4}	\\
LR+DR	& 90.8±0.5	& 69.9±1.9	& 48.7±1.9	& 70.6±1.6	\\
\bottomrule
\end{tabular}
\label{tb:exp-d-binary-bad-size0.01}
\end{table}

\begin{table}[t]
\centering
\caption{Same semantics as in Table~\ref{tb:exp-d-real-eps0.1-size0.01}. \\
This experiment:  binary-valued cost, logging policy $\mu_{\textnormal{bad}, \epsilon=0.1}$, data size $\times0.1$, and \# actions = \# classes. }

\begin{tabular}{llllll}\toprule
Risk * 100	& Letter	& PenDigits	& SatImage	& JPVowel	\\
\midrule
PG+IPW+PL	& 88.1±0.4	& 34.9±1.1	& 21.4±0.6	& 34.4±0.5	\\
PG+IPW+EB	& \textbf{83.8±1.7}	& \textbf{30.3±1.0}	& \textbf{21.0±0.2}	& \textbf{31.9±0.9}	\\
PG+IPW	& 88.3±0.5	& 37.8±1.4	& 22.3±0.7	& 36.3±0.7	\\
\midrule
PG+DR+PL	& 88.2±0.4	& 34.3±0.9	& 21.3±0.5	& 35.2±0.7	\\
PG+DR+EB	& \fbox{\textbf{83.0±1.5}}	& \fbox{\textbf{28.4±0.7}}	& \fbox{\textbf{20.3±0.2}}	& \fbox{\textbf{31.6±1.1}}	\\
PG+DR	& 88.4±0.5	& 38.8±1.5	& 22.4±0.7	& 36.1±0.9	\\
\midrule
LR+IPW+PL	& \textbf{85.7±0.4}	& \textbf{41.1±1.0}	& \textbf{31.4±0.7}	& \textbf{40.6±0.7}	\\
LR+IPW	& 85.9±0.4	& 42.5±1.2	& 34.5±1.6	& 41.9±0.8	\\
\midrule
LR+DR+PL	& \textbf{86.0±0.4}	& \textbf{41.1±1.0}	& \textbf{30.5±0.7}	& \textbf{41.4±0.7}	\\
LR+DR	& 86.1±0.4	& 42.5±1.3	& 35.0±1.6	& 42.6±0.9	\\
\bottomrule
\end{tabular}
\label{tb:exp-d-binary-bad-size0.1}
\end{table}

\begin{table}[t]
\centering
\caption{Same semantics as in Table~\ref{tb:exp-d-real-eps0.1-size0.01}. \\
This experiment:  binary-valued cost, logging policy $\mu_{\textnormal{bad}, \epsilon=0.1}$, data size $\times1$, and \# actions = \# classes. }

\begin{tabular}{llllll}\toprule
Risk * 100	& Letter	& PenDigits	& SatImage	& JPVowel	\\
\midrule
PG+IPW+PL	& 77.7±0.6	& 23.0±0.2	& 18.6±0.1	& 25.0±0.2	\\
PG+IPW+EB	& \fbox{\textbf{63.5±1.9}}	& \fbox{\textbf{21.1±0.1}}	& \textbf{17.0±0.0}	& \fbox{\textbf{22.5±0.1}}	\\
PG+IPW	& 77.9±0.6	& 24.5±0.6	& 19.0±0.1	& 25.6±0.4	\\
\midrule
PG+DR+PL	& 78.8±0.5	& 23.2±0.3	& 18.7±0.1	& 25.1±0.2	\\
PG+DR+EB	& \textbf{66.7±2.5}	& \textbf{21.2±0.1}	& \fbox{\textbf{16.9±0.1}}	& \fbox{\textbf{22.5±0.1}}	\\
PG+DR	& 79.3±0.6	& 24.6±0.5	& 19.0±0.1	& 26.1±0.4	\\
\midrule
LR+IPW+PL	& \textbf{81.5±0.5}	& \textbf{28.3±0.3}	& \textbf{25.2±0.3}	& \textbf{30.1±0.2}	\\
LR+IPW	& 81.6±0.5	& 28.9±0.4	& 25.5±0.3	& 30.3±0.3	\\
\midrule
LR+DR+PL	& \textbf{82.4±0.5}	& \textbf{28.7±0.3}	& \textbf{25.1±0.3}	& \textbf{29.8±0.3}	\\
LR+DR	& 82.6±0.5	& 29.0±0.4	& 25.5±0.3	& 30.1±0.4	\\
\bottomrule
\end{tabular}
\label{tb:exp-d-binary-bad-size1}
\end{table}

\begin{table}[t]
\centering
\caption{Same semantics as in Table~\ref{tb:exp-d-real-eps0.1-size0.01}. \\
This experiment:  binary-valued cost, logging policy $\mu_{\textnormal{good}, \epsilon=0.1}$, data size $\times0.01$, and \# actions = $5\times$ \# classes. }

\begin{tabular}{lllll}\toprule
Risk * 100	& Letter	& PenDigits	& SatImage	& JPVowel	\\
\midrule
PG+IPW+PL	& 86.3±0.4	& 39.3±2.4	& 29.9±1.1	& 40.8±1.9	\\
PG+IPW+EB	& \fbox{\textbf{80.2±2.4}}	& \fbox{\textbf{31.5±1.7}}	& \fbox{\textbf{24.5±0.8}}	& \fbox{\textbf{32.0±1.1}}	\\
PG+IPW	& 92.4±0.4	& 68.5±1.9	& 43.2±2.3	& 69.9±1.9	\\
\midrule
PG+DR+PL	& \textbf{89.3±0.9}	& \textbf{43.4±3.9}	& 31.0±1.8	& \textbf{42.6±3.3}	\\
PG+DR+EB	& 90.2±1.4	& 44.3±3.2	& \textbf{26.8±1.3}	& 49.9±2.8	\\
PG+DR	& 93.8±0.4	& 81.5±2.0	& 57.0±3.4	& 82.7±1.5	\\
\midrule
LR+IPW+PL	& \textbf{87.9±0.5}	& \textbf{39.1±1.1}	& \textbf{29.1±0.9}	& \textbf{43.2±1.1}	\\
LR+IPW	& 91.5±0.5	& 57.2±1.7	& 47.6±2.6	& 58.6±1.2	\\
\midrule
LR+DR+PL	& \textbf{88.0±0.7}	& \textbf{45.1±3.6}	& \textbf{29.4±0.9}	& \textbf{44.9±2.0}	\\
LR+DR	& 92.4±0.4	& 79.0±1.4	& 68.0±2.6	& 78.9±1.1	\\
\bottomrule
\end{tabular}
\label{tb:exp-d-binary-large-action-size0.01}
\end{table}

\begin{table}[t]
\centering
\caption{Same semantics as in Table~\ref{tb:exp-d-real-eps0.1-size0.01}. \\
This experiment:  binary-valued cost, logging policy $\mu_{\textnormal{good}, \epsilon=0.1}$, data size $\times0.1$, and \# actions = $5\times$ \# classes. }

\begin{tabular}{llllll}\toprule
Risk * 100	& Letter	& PenDigits	& SatImage	& JPVowel	\\
\midrule
PG+IPW+PL	& 87.0±0.4	& 30.4±0.6	& 26.8±0.5	& 34.0±0.4	\\
PG+IPW+EB	& \fbox{\textbf{73.1±1.2}}	& \textbf{28.5±0.8}	& \textbf{21.9±0.5}	& \fbox{\textbf{30.9±0.5}}	\\
PG+IPW	& 90.5±0.6	& 56.0±1.5	& 28.4±1.1	& 55.2±1.3	\\
\midrule
PG+DR+PL	& 86.9±0.4	& 30.4±0.6	& 26.5±0.5	& 34.2±0.5	\\
PG+DR+EB	& \textbf{84.5±2.2}	& \fbox{\textbf{27.7±0.7}}	& \fbox{\textbf{21.3±0.5}}	& \textbf{31.7±0.7}	\\
PG+DR	& 92.2±0.4	& 69.6±2.2	& 29.4±1.1	& 66.9±2.1	\\
\midrule
LR+IPW+PL	& \textbf{85.0±0.3}	& \textbf{34.4±0.5}	& \textbf{28.5±0.7}	& \textbf{39.3±0.4}	\\
LR+IPW	& 87.6±0.5	& 42.6±0.9	& 28.9±0.8	& 43.1±0.7	\\
\midrule
LR+DR+PL	& \textbf{85.4±0.3}	& \textbf{34.8±0.3}	& \textbf{29.2±0.5}	& \textbf{40.2±0.6}	\\
LR+DR	& 88.0±0.5	& 61.3±1.3	& 41.2±1.7	& 57.5±1.0	\\
\bottomrule
\end{tabular}
\label{tb:exp-d-binary-large-action-size0.1}
\end{table}

\begin{table}[ht]
\centering
\caption{Same semantics as in Table~\ref{tb:exp-d-real-eps0.1-size0.01}. \\
This experiment:  binary-valued cost, logging policy $\mu_{\textnormal{good}, \epsilon=0.1}$, data size $\times1$, and \# actions = $5\times$ \# classes. }

\begin{tabular}{llllll}
\toprule
Risk * 100	& Letter	& PenDigits	& SatImage	& JPVowel	\\
\midrule
PG+IPW+PL	& 85.3±0.4	& \textbf{25.9±0.5}	& \textbf{19.7±0.2}	& 31.3±0.4	\\
PG+IPW+EB	& \textbf{70.1±3.0}	& 26.9±0.7	& 20.6±0.4	& \textbf{30.3±0.5}	\\
PG+IPW	& 87.9±0.6	& 27.2±0.7	& 20.2±0.5	& 33.2±0.7	\\
\midrule
PG+DR+PL	& 85.5±0.4	& 25.8±0.4	& \fbox{\textbf{19.2±0.1}}	& 31.3±0.4	\\
PG+DR+EB	& \fbox{\textbf{67.0±2.5}}	& \fbox{\textbf{24.6±0.5}}	& 19.3±0.2	& \fbox{\textbf{27.5±0.6}}	\\
PG+DR	& 88.8±0.5	& 29.7±1.0	& 20.4±0.7	& 33.2±0.5	\\
\midrule
LR+IPW+PL	& \textbf{75.7±0.6}	& \textbf{29.8±0.4}	& \textbf{25.9±0.2}	& \textbf{32.9±0.5}	\\
LR+IPW	& 76.6±0.5	& 31.5±0.7	& 26.0±0.2	& 33.8±0.5	\\
\midrule
LR+DR+PL	& \textbf{76.4±0.5}	& \textbf{33.0±0.2}	& \textbf{26.5±0.2}	& \textbf{37.2±0.3}	\\
LR+DR	& 77.5±0.5	& 47.0±1.1	& 26.9±0.2	& 44.2±0.6	\\
\bottomrule
\end{tabular}
\label{tb:exp-d-binary-large-action-size1}
\end{table}

\clearpage

\subsection{Experiments with continuous actions}
\label{sec:additional-exp-cont}
\xhdr{Datasets.} For the continuous-action setting, we follow prior works~\citep{bietti2021contextual,majzoubi2020efficient,zhu2022contextual} to simulate bandit instances using $5$ regression datasets from OpenML~\citep{OpenML2013}, see Table~\ref{tb:coninuous-data-detail} for details.

\begin{table}[h]
\centering
\caption{Continuous action datasets}
\label{tb:coninuous-data-detail}
\begin{tabular}{llllll}
\toprule
Dataset    & Wisconsin & AutoPrice   & CpuAct & Zurich  & BlackFriday \\
\midrule
OpenML ID   & 1187 & 1189 & 1190 & 40753    & 44057 \\
\# Data     & 1,000,000  & 1,000,000  & 1,000,000      & 5,465,575     & 166,821      \\
\# Features & 32         & 15        & 21           & 14            & 9            \\
\bottomrule
\end{tabular}
\end{table}

We use one-hot representations for categorical features and map the regression targets to $[0,1]$. We adopt the same split of the datasets as in the discrete-action experiments for training logging policies, simulating bandit feedback, and testing performance. And we still consider three data sizes the same as the discrete-action experiments.

To simulate bandit feedback, for each example $(x,y)$ from the regression dataset, where $y$ is the regression target, we take an action $a$ following a logging policy $\mu$, and observe the loss $\ell(a) = |a - y|$.

\xhdr{Logging policies.} We consider logging policies that are combinations of a policy smoothed from a deterministic policy and a policy that selects actions uniformly at random. To learn a deterministic policy, we train a linear regression model with $\ell_2$ regularization to predict the regression target on the $1\%$ for held out data, and regard the regression estimate clipped into $[0,1]$ as the taken action. Then we construct stochastic logging policies $\mu_{\epsilon=0.1}$ and $\mu_{\epsilon=0.01}$ by smoothing the deterministic policy with bandwidth $0.1$, and combing it with a uniformly at random policy where $\epsilon =0.1$ and $\epsilon=0.01$ represent the probabilities of using the uniformly-at-random policy.
To summarize, for environment setting, we have data size in $\{X 0.01, X 0.1, X 1\}$ logging policy in $\{\mu_{\epsilon=0.1}, \mu_{\epsilon = 0.01}\}$.

\xhdr{Methods.} All the regularizers, oracles, and estimators are the same as those of the discrete-action experiments as shown in Table~\ref{tb:discrete-algorithm-options}. Since EB takes much longer time to run due to its reliance on going through the whole dataset multiple times, we only run experiments for EB on data sizes $\times 0.01$ and $\times 0.1$.

\xhdr{Hyper-parameter details.} For the continuous-action experiments, we run each experiment for $10$ times. We grid search $K$ in $[10, 20, 50, 100]$, and $H$ in $[1e-2, 2e-2, 5e-2, 1e-1]$. For the continuous-action experiments, we grid search in a smaller set of learning rates $[1e-4, 1e-3, 1e-2, 1e-1]$. All the other hyper-parameters are the same as those of the discrete-action experiments.

\begin{figure}
\begin{minipage}{.48\textwidth}
  \centering
  \includegraphics[width=.98\linewidth]{plots/c_eps0.1_improvement.pdf}
\captionof{figure}{
Relative improvement (\RelImp, see \eqref{eq:RelImp}) for PL against the baseline with no pessimism, averaged over all 10 runs (mean $\pm$ $2$ standard errors).
\vspace{2mm}\newline
Shown for a particular (dataset, environment) pair and the best-performing (CSC oracle, risk estimator) pair. Each bar corresponds to a (dataset, data-size) pair.
\vspace{2mm}\newline
6 environments, see Table~\ref{tb:continuous-setting-options}; 
5 datasets, see Table~\ref{tb:coninuous-data-detail}.
\vspace{2mm}\newline
Environment: logging policy $\mu_{\epsilon=0.1}$. }
  \label{fig:c_eps0.1_improvement_appendix}
\end{minipage}
\hfill
\begin{minipage}{.48\textwidth}
  \centering
  \includegraphics[width=.98\linewidth]{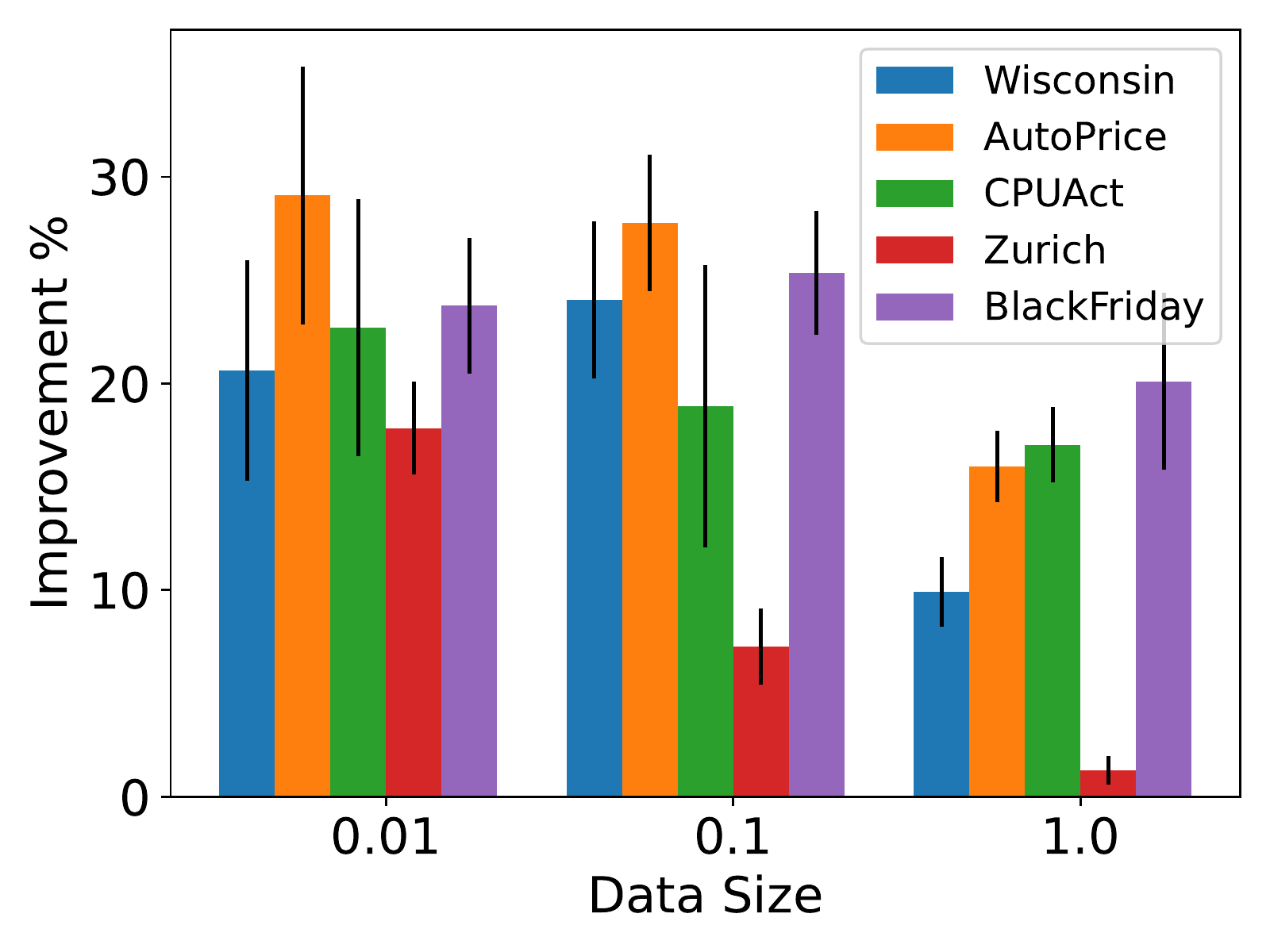}
    \captionof{figure}{Same semantics as in Figure~\ref{fig:c_eps0.1_improvement_appendix}. \\
    Environment: logging policy $\mu_{\epsilon=0.01}$. }
  \label{fig:c_eps0.01_improvement}
\end{minipage}
\end{figure}

\xhdr{Results.} We conduct experiments on all combinations of data sizes and logging policies.
The experiment results are illustrated in
\asedit{Figures~\ref{fig:c_eps0.1_improvement_appendix}-\ref{fig:c_eps0.01_improvement} and Tables~\ref{tb:exp_c_eps0.1_size0.01}-\ref{tb:exp_c_eps0.01_size1}, with semantics mirroring those of of Figure~\ref{fig:main_discrete} (right) and Table~\ref{tb:exp-d-real-eps0.1-size0.1}, respectively.}


\begin{table}[h]
\centering
\caption{Computation time for the continuous-action experiments }
\label{tb:computation-continuous}
\begin{tabular}{lllllll}
\toprule
Dataset    & Wisconsin & AutoPrice   & CpuAct & Zurich  & BlackFriday   \\
\midrule
total time (core $\cdot$ hours) & 93.7 & 88.8 & 92.5 & 499.9 & 19.3 \\
\bottomrule
\end{tabular}
\end{table}

\begin{table}[t]
\centering
\caption{
Performance of different OPO methods: mean $\pm$ two standard errors over 10 runs. Bold numbers represent the best performance within each (CSC oracle, estimator) pair. Boxed numbers represent the best across all algorithmic configurations. \\ This experiment:  logging policy $\mu_{ \epsilon=0.1}$, and data size $\times 0.01$. }

\begin{tabular}{llllll}
\toprule
Risk * 100	& Wisconsin	& AutoPrice	& CPUAct	& Zurich	& BlackFriday	\\
\midrule
PG+IPW+PL	& 26.2±2.1	& 18.8±1.6	& 20.4±1.8	& \textbf{24.6±0.3}	& 29.8±1.5	\\
PG+IPW+EB	& \textbf{22.4±1.4}	& \textbf{15.1±1.2}	& \textbf{16.4±1.7}	& 25.1±1.1	& \fbox{\textbf{22.7±0.6}}	\\
PG+IPW	& 30.3±1.0	& 23.5±2.1	& 21.4±1.1	& 27.7±1.1	& 33.0±1.4	\\
\midrule
PG+DR+PL	& 30.5±3.6	& 21.5±2.3	& 19.6±1.1	& 24.4±0.1	& 31.2±2.4	\\
PG+DR+EB	& \fbox{\textbf{21.4±0.1}}	& \fbox{\textbf{14.5±0.2}}	& \fbox{\textbf{14.8±0.3}}	& \fbox{\textbf{24.3±0.0}}	& \textbf{24.9±2.0}	\\
PG+DR	& 33.0±1.9	& 24.6±1.4	& 23.0±1.7	& 26.0±0.4	& 33.6±0.7	\\
\midrule
LR+IPW+PL	& \textbf{25.5±2.4}	& \textbf{19.4±1.5}	& \textbf{20.1±1.2}	& \textbf{25.8±1.9}	& \textbf{30.2±2.6}	\\
LR+IPW	& 30.6±1.0	& 25.4±2.0	& 24.0±1.5	& 30.6±1.3	& 32.4±1.0	\\
\midrule
LR+DR+PL	& \textbf{32.6±2.9}	& \textbf{22.5±3.6}	& \textbf{25.8±2.7}	& \fbox{\textbf{24.3±0.0}}	& \textbf{33.0±2.7}	\\
LR+DR	& 34.1±1.2	& 24.6±1.8	& 26.3±1.4	& 29.1±1.0	& 33.5±1.5	\\
\bottomrule
\end{tabular}
\label{tb:exp_c_eps0.1_size0.01}
\end{table}

\begin{table}[t]
\centering
\caption{Same semantics as in Table~\ref{tb:exp_c_eps0.1_size0.01}. \\
This experiment: logging policy $\mu_{ \epsilon=0.1}$, and data size $\times 0.1$. }

\begin{tabular}{llllll}
\toprule
Risk * 100	& Wisconsin	& AutoPrice	& CPUAct	& Zurich	& BlackFriday	\\
\midrule
PG+IPW+PL	& 22.7±0.8	& 16.9±1.4	& 17.8±1.6	& \textbf{24.4±0.1}	& 25.0±1.8	\\
PG+IPW+EB	& \fbox{\textbf{21.5±0.1}}	& \textbf{14.7±0.3}	& \textbf{14.6±0.2}	& \textbf{24.4±0.1}	& \textbf{22.9±1.1}	\\
PG+IPW	& 26.6±1.1	& 20.2±0.9	& 19.9±1.1	& 26.3±0.7	& 30.0±1.2	\\
\midrule
PG+DR+PL	& 21.8±0.1	& 15.3±0.2	& 15.8±0.6	& \fbox{\textbf{24.3±0.0}}	& 24.8±2.2	\\
PG+DR+EB	& \fbox{\textbf{21.5±0.3}}	& \fbox{\textbf{14.4±0.1}}	& \fbox{\textbf{14.5±0.1}}	& 24.4±0.1	& \fbox{\textbf{22.5±0.4}}	\\
PG+DR	& 24.0±0.2	& 18.0±0.9	& 17.3±0.3	& 24.6±0.1	& 30.9±2.2	\\
\midrule
LR+IPW+PL	& \textbf{24.1±1.4}	& \textbf{18.5±0.8}	& \textbf{18.9±1.2}	& \textbf{24.4±0.0}	& \textbf{24.7±0.8}	\\
LR+IPW	& 27.7±0.6	& 21.1±0.8	& 20.2±1.1	& 26.9±0.4	& 31.8±1.3	\\
\midrule
LR+DR+PL	& \textbf{23.1±0.8}	& \textbf{17.7±0.9}	& 19.5±1.6	& \textbf{24.4±0.0}	& \textbf{27.1±3.1}	\\
LR+DR	& 26.4±0.4	& 18.7±0.5	& \textbf{19.3±0.7}	& 25.2±0.2	& 33.0±2.5	\\
\bottomrule
\end{tabular}
\label{tb:exp_c_eps0.1_size0.1_appendix}
\end{table}

\begin{table}[t]
\centering
\caption{Same semantics as in Table~\ref{tb:exp_c_eps0.1_size0.01}. \\
This experiment: logging policy $\mu_{ \epsilon=0.1}$, and data size $\times 1$. }

\begin{tabular}{llllll}
\toprule
Risk * 100	& Wisconsin	& AutoPrice	& CPUAct	& Zurich	& BlackFriday	\\
\midrule
PG+IPW+PL	& \textbf{21.9±0.2}	& \textbf{15.3±0.5}	& \textbf{15.5±0.3}	& \textbf{24.4±0.1}	& \textbf{22.8±0.3}	\\
PG+IPW	& 24.3±0.3	& 18.2±0.4	& 17.7±0.4	& 24.6±0.1	& 26.5±0.8	\\
\midrule
PG+DR+PL	& \fbox{\textbf{21.5±0.0}}	& \fbox{\textbf{14.7±0.1}}	& \fbox{\textbf{14.8±0.1}}	& \fbox{\textbf{24.3±0.1}}	& \fbox{\textbf{22.6±0.3}}	\\
PG+DR	& 22.5±0.1	& 15.6±0.2	& 15.2±0.3	& \fbox{\textbf{24.3±0.1}}	& 23.2±0.5	\\
\midrule
LR+IPW+PL	& \textbf{22.8±0.2}	& \textbf{16.9±0.4}	& 17.3±0.5	& \textbf{24.4±0.0}	& \textbf{24.7±0.8}	\\
LR+IPW	& 24.3±0.2	& 17.8±0.6	& \textbf{17.2±0.6}	& 25.1±0.4	& 26.3±0.6	\\
\midrule
LR+DR+PL	& \textbf{22.2±0.1}	& \textbf{15.3±0.2}	& \textbf{15.4±0.3}	& \fbox{\textbf{24.3±0.0}}	& \textbf{23.0±0.4}	\\
LR+DR	& 22.7±0.3	& 15.6±0.2	& 15.7±0.3	& 24.5±0.0	& 23.9±0.3	\\
\bottomrule
\end{tabular}
\label{tb:exp_c_eps0.1_size1}
\end{table}

\begin{table}[t]
\centering
\caption{Same semantics as in Table~\ref{tb:exp_c_eps0.1_size0.01}. \\
This experiment: logging policy $\mu_{ \epsilon=0.01}$, and data size $\times 0.01$. }

\begin{tabular}{llllll}
\toprule
Risk * 100	& Wisconsin	& AutoPrice	& CPUAct	& Zurich	& BlackFriday	\\
\midrule
PG+IPW+PL	& 22.6±0.3	& 16.1±0.2	& 18.8±2.3	& \fbox{\textbf{24.4±0.0}}	& 24.2±0.2	\\
PG+IPW+EB	& \fbox{\textbf{21.3±0.0}}	& \fbox{\textbf{14.6±0.2}}	& \fbox{\textbf{14.5±0.1}}	& \fbox{\textbf{24.4±0.0}}	& \fbox{\textbf{22.6±0.1}}	\\
PG+IPW	& 29.9±2.7	& 25.8±3.5	& 25.7±1.5	& 30.8±1.8	& 33.6±1.2	\\
\midrule
PG+DR+PL	& 30.9±2.6	& 20.9±3.0	& 23.9±4.9	& 28.8±4.1	& 26.0±1.6	\\
PG+DR+EB	& \textbf{25.5±2.2}	& \textbf{18.1±1.6}	& \textbf{18.5±1.4}	& \textbf{25.5±2.1}	& \textbf{25.9±1.5}	\\
PG+DR	& 32.3±1.7	& 27.2±3.0	& 29.6±5.7	& 34.3±2.2	& 34.7±1.4	\\
\midrule
LR+IPW+PL	& \textbf{23.1±0.3}	& \textbf{17.2±0.5}	& \textbf{19.0±1.5}	& \fbox{\textbf{24.4±0.1}}	& \textbf{24.1±0.2}	\\
LR+IPW	& 31.7±2.4	& 27.6±3.3	& 26.5±1.7	& 33.1±1.3	& 31.8±1.3	\\
\midrule
LR+DR+PL	& \textbf{30.8±3.2}	& \textbf{24.6±3.1}	& \textbf{26.0±3.7}	& \textbf{31.4±3.2}	& \textbf{27.4±2.5}	\\
LR+DR	& 33.5±2.2	& 26.8±1.7	& 33.5±3.1	& 34.8±1.3	& 34.8±1.4	\\
\bottomrule
\end{tabular}
\label{tb:exp_c_eps0.01_size0.01}
\end{table}

\begin{table}[t]
\centering
\caption{Same semantics as in Table~\ref{tb:exp_c_eps0.1_size0.01}. \\
This experiment: logging policy $\mu_{ \epsilon=0.01}$, and data size $\times 0.1$. }

\begin{tabular}{llllll}
\toprule
Risk * 100	& Wisconsin	& AutoPrice	& CPUAct	& Zurich	& BlackFriday	\\
\midrule
PG+IPW+PL	& 22.4±0.9	& 16.6±1.6	& 18.1±1.8	& 25.1±0.9	& 23.0±0.1	\\
PG+IPW+EB	& \fbox{\textbf{21.3±0.0}}	& \fbox{\textbf{14.4±0.0}}	& \fbox{\textbf{14.3±0.1}}	& \fbox{\textbf{24.4±0.0}}	& \fbox{\textbf{22.5±0.1}}	\\
PG+IPW	& 29.6±1.9	& 24.4±1.4	& 22.8±1.7	& 28.4±1.1	& 31.6±1.7	\\
\midrule
PG+DR+PL	& 27.4±2.9	& 23.5±4.0	& 19.8±1.8	& \textbf{24.5±0.4}	& 27.6±4.2	\\
PG+DR+EB	& \textbf{22.7±1.8}	& \textbf{16.8±2.2}	& \textbf{18.3±2.4}	& 24.9±0.7	& \textbf{25.8±2.8}	\\
PG+DR	& 34.5±2.2	& 25.3±2.3	& 24.2±2.8	& 26.4±0.6	& 35.7±1.8	\\
\midrule
LR+IPW+PL	& \textbf{23.5±1.4}	& \textbf{20.5±1.7}	& \textbf{20.3±1.6}	& \textbf{24.8±0.4}	& \textbf{24.8±0.8}	\\
LR+IPW	& 31.0±0.9	& 23.9±1.3	& 23.3±2.3	& 30.3±1.2	& 32.6±1.1	\\
\midrule
LR+DR+PL	& \textbf{33.2±2.2}	& \textbf{25.6±3.3}	& \textbf{25.2±4.6}	& \textbf{24.5±0.1}	& \textbf{31.6±4.0}	\\
LR+DR	& 33.6±1.4	& 25.7±1.5	& 26.6±2.3	& 29.3±1.1	& 36.5±2.5	\\
\bottomrule
\end{tabular}
\label{tb:exp_c_eps0.01_size0.1}
\end{table}

\begin{table}[t]
\centering
\caption{Same semantics as in Table~\ref{tb:exp_c_eps0.1_size0.01}. \\
This experiment: logging policy $\mu_{ \epsilon=0.01}$, and data size $\times 1$. }

\begin{tabular}{llllll}
\toprule
Risk * 100	& Wisconsin	& AutoPrice	& CPUAct	& Zurich	& BlackFriday	\\
\midrule
PG+IPW+PL	& \textbf{23.1±1.6}	& \textbf{16.3±1.5}	& \textbf{15.4±0.7}	& \fbox{\textbf{24.4±0.1}}	& \textbf{24.8±2.4}	\\
PG+IPW	& 27.5±1.3	& 21.1±1.6	& 20.0±1.3	& 25.9±0.5	& 29.7±2.1	\\
\midrule
PG+DR+PL	& \fbox{\textbf{21.7±0.1}}	& \fbox{\textbf{14.8±0.2}}	& \fbox{\textbf{14.6±0.1}}	& \fbox{\textbf{24.4±0.0}}	& \fbox{\textbf{22.9±0.3}}	\\
PG+DR	& 24.0±0.4	& 17.8±0.6	& 17.7±0.5	& 24.7±0.3	& 33.2±2.2	\\
\midrule
LR+IPW+PL	& \textbf{22.9±0.4}	& \textbf{18.8±0.9}	& \textbf{18.7±0.3}	& \textbf{24.9±0.8}	& \textbf{26.0±1.7}	\\
LR+IPW	& 27.5±0.8	& 21.2±0.8	& 20.2±1.1	& 26.9±0.6	& 29.9±0.8	\\
\midrule
LR+DR+PL	& \textbf{22.7±0.4}	& \textbf{18.0±1.0}	& \textbf{18.9±0.8}	& \textbf{24.5±0.0}	& \textbf{27.7±2.5}	\\
LR+DR	& 27.0±1.1	& 18.6±0.5	& 19.8±1.0	& 25.5±0.2	& 33.0±2.5	\\
\bottomrule
\end{tabular}
\label{tb:exp_c_eps0.01_size1}
\end{table} 

\end{document}